\documentclass[journal]{IEEEtran}
\usepackage[utf8]{inputenc}
\usepackage[T1]{fontenc}
\usepackage{lmodern}
\usepackage[english]{babel}

\usepackage{graphicx}
\usepackage{tabularx}

\usepackage{amsmath, amssymb,amsfonts}
\usepackage{hyperref}
\usepackage{csquotes}
\usepackage{authblk} 
\usepackage[style=ieee]{biblatex}
\AtEveryBibitem{\clearfield{language,url,eprint}}
\AtEveryBibitem{\clearfield{url}}
\AtEveryBibitem{\clearfield{eprint}}

\AtEveryBibitem{\clearlist{language}} 
\AtEveryBibitem{\clearlist{url}}
\AtEveryBibitem{\clearlist{eprint}} 

\usepackage{etoolbox} 

\usepackage{booktabs}
\usepackage{rotating}

\title{Beyond Silicon: Materials, Mechanisms, and Methods for Physical Neural Computing}

\author[1]{Stefan Fischer\thanks{Also available at \url{https://ieeexplore.ieee.org/document/11513224}.}}
\author[2]{Nihat Ay}
\author[3,4,5]{Olaf Landsiedel}
\author[6]{Esfandiar Mohammadi}
\author[7]{Sebastian Otte}
\author[8]{Bernd-Christian Renner}
\author[9]{Nele Rußwinkel}

\affil[1]{Institute for Telematics, University of Lübeck, Germany}
\affil[2]{Institute for Data Science Foundations, TU Hamburg, Germany}

\affil[3]{Institute for Networked Cyber-Physical Systems, TU Hamburg, Germany}
\affil[4]{United Nations University Hub on Engineering to Face Climate Change at TU Hamburg, United Nations University Institute for Water, Environment and Health (UNU-INWEH), Hamburg, Germany}
\affil[5]{Kiel University, Kiel, Germany}

\affil[6]{Institute for IT Security, University of Lübeck, Germany}
\affil[7]{Institute for Robotics and Cognitive Systems, University of Lübeck, Germany}
\affil[8]{Institute for Autonomous Cyber-Physical Systems, TU Hamburg, Germany}
\affil[9]{Institute for Information Systems, University of Lübeck, Germany}

\begin{document}

\maketitle

\begin{abstract}
Physical implementations of neural computation now extend far beyond silicon hardware, encompassing substrates such as memristive devices, photonic circuits, mechanical metamaterials, microfluidic networks, chemical reaction systems, and living neural tissue. By exploiting intrinsic physical processes, such as charge transport, wave interference, elastic deformation, mass transport, and biochemical regulation, these substrates can realize neural inference and adaptation directly in matter.
As silicon GPU-centered AI faces growing energy and data-movement constraints, physical neural computation becomes increasingly relevant as a complementary path beyond conventional digital accelerators. This trend is driven in particular by pervasive intelligence, i.e., the deployment of on-device and edge AI across large numbers of resource-constrained systems. In such settings, co-locating computation with sensing and memory can reduce data shuttling and improve efficiency. Meanwhile, physical neural approaches have emerged across disparate disciplines, yet progress remains fragmented, with limited shared terminology and few principled ways to compare platforms.
This survey unifies the field by mapping neural primitives to substrate-specific mechanisms, analyzing architectural and training paradigms, and identifying key engineering constraints including scalability, precision, programmability, and I/O interfacing overhead. 
To enable cross-domain comparison, we introduce a first-order benchmarking scheme based on standardized static and dynamic tasks and physically interpretable performance dimensions. 
We show that no single substrate dominates across the considered dimensions; instead, physical neural systems occupy complementary operating regimes, enabling applications ranging from ultrafast signal processing and in-memory inference to embodied control and in-sample biochemical decision making.
\end{abstract}

% PART I ---------------------------------------------------------

\section{Introduction}
\label{sec:introduction}

In the domain of artificial intelligence, in-silico implementations of Artificial Neural Networks (ANN) are currently dominating. From the earliest perceptrons to modern Large Language Models (LLMs) and Foundation Models (FMs), the substrate of choice has been the transistor-based integrated circuit. Digital CMOS has dominated AI hardware mainly due to manufacturability, programmability, and ecosystem maturity—but not because alternative physical substrates are fundamentally incapable. As we approach the asymptotic limits of Moore's Law and Dennard Scaling \cite{Shalf2020}, the research community is increasingly forced to look ``beyond silicon.'' 
However, the motivation to explore alternative substrates extends far beyond the mere stagnation of lithographic improvements. 
Beyond data-center training, the rise of pervasive intelligence in the form of on-device and edge AI deployed across large numbers of resource-constrained systems makes energy and internal data-transport costs first-order design constraints.
In this setting, physical neural computing becomes newly relevant as a complementary path to conventional digital accelerators by co-locating computation with sensing and memory and exploiting substrate dynamics to reduce data shuttling.
This survey explores the emerging field of Physical Neural Networks (PNNs), systems that do not merely simulate neural dynamics digitally, but physically embody them in ``wet'' (biological/chemical) and ``solid/fluid'' (mechanical/microfluidic) media.
These developments motivate treating a coherent research direction centered on physical neural networks, where substrate physics is not an implementation detail but a determinant of how learning itself can be realized \cite{Momeni2025}.

\subsection{Beyond the Turing and Von Neumann Paradigms}

Contemporary Artificial Neural Networks are abstractions. They are mathematical constructs composed of weights, biases, and activation functions, executed on hardware that is not originally tailored for their mode of operation. The mainstream computing architecture, the Von Neumann machine, relies on a strict separation of processing units (CPU/GPU) and memory.

This separation gives rise to the ``Von Neumann bottleneck,'' where a significant fraction of both energy and execution time is primarily consumed by data movement rather than by the computation itself \cite{10.1145/359576.359579}. As ANN models continue to scale in size and complexity, this mismatch between algorithmic structure and hardware organization becomes increasingly salient. PNNs instead couple algorithmic operations to substrate dynamics, embedding computation directly in physical state evolution \cite{Momeni2025}.

While neuromorphic engineering has attempted to bridge this gap using silicon (e.g.\ through digital layouts optimized for sparse, event-driven computation, analog sub-threshold circuits, or memristors) \cite{58356, Strukov2008}, these approaches still largely rely on electron flow in solid semiconductors. They remain constrained by the physics of the electron in a rigid lattice. By contrast, biological brains (the inspiration for ANNs) utilize ions, chemical gradients, and fluid dynamics in a wet, three-dimensional medium. 

Replicating and potentially even surpassing biological efficiency demands a fundamental rethinking of the physical substrate of neural computation and learning, way beyond evidently suboptimal electron transport in silicon-based hardware.

\subsection{The Need for Physical Neural Networks (PNNs)}

The motivation for non-silicon neural networks is often reduced to energy efficiency. While this is valid, it is incomplete. We outline core arguments for PNNs that enable capabilities not achievable with conventional electronic computing alone.

\subsubsection{Eliminating the ``Transduction Tax'' (In-Situ Computation)}
Current cyber-physical pipelines rely on costly transduction chains. For example, biomarker detection with deep learning requires converting a chemical signal into an electrical current, amplifying and digitizing it (ADC), processing it on digital hardware, and often converting results back into analog actions. Each step adds latency, noise, and energy loss. Chemical PNNs can instead process molecules directly via reaction-diffusion dynamics \cite{adamatzky2005reaction}, computing in the data’s native domain. This transduction overhead is quantifiable: ADC and digital signal processing can dominate the energy budget of sensing pipelines, in some systems accounting for most of total consumption \cite{Chen2025}. Nucleic-acid-based molecular logic toolkits demonstrate diagnostic decisions directly in biological fluids without electronic readout \cite{D5SC06176H}, supporting in-situ computation as a practical way to bypass the transduction chain.

\subsubsection{Thermodynamic Computing: Physics as Logic}
Digital logic expends energy to maintain stable ``$0$'' and ``$1$'' states against thermal noise. PNNs can instead exploit thermodynamic relaxation. Mechanical or chemical systems naturally converge to energy minima, which can encode solutions to optimization problems \cite{Feynman1982}. Spring glasses and elastic networks relax into states that map to Ising-type energy landscapes \cite{Hopfield1985,rolandi2026}. In such systems, relaxation, diffusion, or wave propagation directly perform the computation.

\subsubsection{Volumetric Density and Connectivity}
Conventional circuits are essentially planar, with 3D stacking constrained by heat and interconnect limits. Wet and molecular computing substrates are inherently volumetric. Even small droplets can host extremely large numbers of interacting molecules, with 3D diffusion providing connectivity densities that can exceed solid-state wiring schemes \cite{QianWinfree2011}.

\subsubsection{The ``Clockless'' Continuum}
Digital simulation of physics requires time discretization and numerical integration, introducing approximation error and clock overhead. Physical substrates operate in continuous time. A fluidic system implements Navier-Stokes \cite{sohr2001navierstokes} dynamics natively; the computational question is the I/O mapping and controllability of that dynamics. While temporal resolution is fundamentally bounded only by physical limits (Planck time), practical analog systems are constrained by thermal and material noise, yielding an effective signal-to-noise–limited bit depth often below 32-bit floating point precision \cite{Wang2025}.

\subsubsection{Resilience and Self-Healing}
Semiconductor hardware is structurally brittle, whereas biological and fluidic substrates can remain functional under deformation or partial damage. Reaction-diffusion media such as Belousov–Zhabotinsky systems can continue operating despite container deformation or division \cite{Adamatzky_2011}. Synthetic biological implementations additionally offer self-repair and regeneration capabilities absent in silicon systems.

\subsubsection{Environmental Integration and Biocompatibility}
Silicon devices are poorly suited for saline or biological environments. PNNs based on DNA, proteins, or biocompatible fluids can operate inside living tissue, in the bloodstream, or in sensitive ecosystems \cite{D5SC06176H}. This enables embedded intelligence for medical and ecological applications.

\subsection{Scope}
\label{sec:scope}

This review focuses on \emph{physical neural networks} (PNNs), defined as systems in which training and inference exploit the intrinsic dynamics of a material substrate rather than being realized solely through numerical operations on conventional digital processors \cite{Wright2022, Momeni2025}. The emphasis is on computation implemented through physical state evolution, including charge transport, wave propagation, mechanical deformation, mass transport, chemical kinetics, and biological plasticity.

To maintain a clear technical focus, the following topics are explicitly excluded:

\begin{itemize}
    \item \textbf{In-silico neural networks:} conventional artificial neural networks executed on CPUs, GPUs, or TPUs without essential reliance on substrate physics.
    \item \textbf{Standard CMOS neuromorphic hardware:} digital or mixed-signal silicon platforms such as Loihi \cite{8259423} or TrueNorth \cite{akopyan2015truenorth}, unless their operation fundamentally depends on non-silicon physical effects or materials.
    \item \textbf{Quantum neural computing:} neural models based on quantum superposition or entanglement, which constitute a distinct computational paradigm with separate theoretical and engineering challenges.
\end{itemize}

Physical neural computing, as used in this review, overlaps with but is not identical to several neighboring research areas. First, it is narrower than analog computing: while analog systems in general process continuous variables through physical dynamics, they are only considered physical neural systems here if these dynamics realize neural primitives such as weighted summation, nonlinearity, memory, or adaptation in a functionally neural architecture. Second, it overlaps only partially with neuromorphic hardware. Neuromorphic systems are inspired by neural organization, often through spiking, event-driven, or in-memory architectures, but many such systems remain conventional CMOS implementations and therefore fall outside our scope unless their computational advantage fundamentally relies on non-silicon substrate physics. Third, reservoir computing is treated here not as a separate substrate class, but as a computational regime in which a physical system with rich internal dynamics serves as a fixed reservoir and only the readout is trained. Reservoir computing can therefore be realized within several of the substrate classes reviewed below.

Within these boundaries, the paper is organized into four conceptual parts that span the design space of physical neural computing.

\paragraph*{Part 1: Foundations and conceptual framework (Sections \ref{sec:introduction}--\ref{sec:theory})}
The first part motivates physical neural computing, defines core terminology, and maps neural primitives (weighted summation, nonlinearity, memory, adaptation) to physical mechanisms. It positions PNNs relative to neuromorphic engineering and digital ML accelerators, and highlights system-level challenges when computation is embedded in physical substrates.

\paragraph*{Part 2: Wetware and molecular intelligence (Sections \ref{sec:dna}--\ref{sec:sbi})}
The second part surveys chemical and biological substrates, from DNA strand displacement and molecular reaction circuits to reaction-diffusion systems and synthetic biological platforms. It emphasizes how biochemical kinetics, diffusion, and cellular regulation realize neural primitives and enable partially autonomous computation in chemical or living media.

\paragraph*{Part 3: Physical neural hardware substrates (Sections \ref{sec:memristive_pcm}--\ref{sec:iontronics})}
The third part reviews engineered non-biological platforms, including memristive, phase-change, and ferroelectric in-memory devices, spintronic and superconducting systems, photonic, mechanical, metamaterial, microfluidic, and iontronic networks. The focus is on shared architectures, training and calibration strategies, and trade-offs in bandwidth, precision, scalability, robustness, and I/O overhead.

\paragraph*{Part 4: Cross-substrate benchmarking and outlook (Sections \ref{sec:benchmarking}--\ref{sec:outlook})}
The final part introduces a comparative benchmarking framework based on standardized static and dynamic tasks and physically meaningful metrics. It then discusses design principles, application domains, and the key challenges in translating laboratory PNN concepts into scalable engineering technologies.

In our view, the recent proliferation of physical neural substrates represents not merely a diversification of hardware platforms, but a shift toward treating physical dynamics as a first-class computational resource. This perspective motivates the unified treatment adopted throughout the review.

%------------------------------------------------

\section{The Physics of Neural Computation}
\label{sec:theory}

%To engineer intelligence into physical substrates, we must establish a rigorous mapping between the mathematical abstractions of neural modeling and computation and the governing laws of physics. While conventional hardware implements these abstractions logically via Boolean gates, Physical Neural Networks (PNNs) implement them isomorphically: the physical interaction \textit{is} the computation. 
%To engineer intelligence in physical substrates, we must relate neural computation to the governing laws of physics. Conventional hardware implements neural models indirectly through Boolean logic. In contrast, Physical Neural Networks (PNNs) implement them directly in the substrate: the physical interaction itself performs the computation.
%This section generalizes the neural primitives connectivity (Weights), non-linearity, and optimization (learning) across both ``wet'' (diffusion/reaction) and ``solid/fluid'' (transport/interaction) domains, and establishes a taxonomy for their training and classification.

To engineer intelligence in physical substrates, we must relate neural computation to the governing laws of physics.
Conventional hardware implements neural models indirectly through Boolean logic. In contrast, Physical Neural Networks (PNNs) implement them directly in the substrate: the physical interaction itself performs the computation. This section therefore does not only generalize the neural primitives of connectivity (weights), nonlinearity, and optimization (learning) across both ``wet'' (diffusion/reaction) and ``solid/fluid'' (transport/interaction) domains; it also makes explicit how material properties determine which neural primitives are natively available, and how device and network architecture determine how these primitives are organized into a computational system. In other words, the material sets the native computational vocabulary, whereas the architecture governs signal propagation, coupling structure, and the overall organization of computation.

%------------------------------

\subsection{Physics as Computation: Transport, Interaction, and Relaxation}
\label{subsec:physics}

A generic neural network relies on three core operations: \emph{weighted signal transmission}, \emph{nonlinear activation}, and \emph{state evolution toward a solution}. In Physical Neural Networks (PNNs), these primitives are not executed symbolically by Boolean logic; instead, they arise directly from substrate physics through (i) linear transport and superposition, (ii) intrinsic nonlinear interaction and thresholds, and (iii) relaxation dynamics that implement optimization.

\subsubsection{Signal Transport and Weighting (The Linear Regime)}
In ANNs, the weighted sum $\sum_j w_{ij}x_j$ is the fundamental linear operation. In wetware, such linear accumulation is naturally mediated by diffusion: molecular connectivity is effectively ``wireless'' and governed by Fick's laws, so concentration fields spread and mix without explicit wiring. At cellular scales, flows typically occur at very low Reynolds numbers where viscosity dominates inertia; motion is laminar and effectively time-reversible, making diffusion---rather than turbulent mixing---the dominant mechanism for distributing signals \cite{Purcell1977}. In engineered hardware, the same linear primitive is realized by conservation laws: Kirchhoff-type constraints make nodal summation of currents or flows thermodynamically ``free'' \cite{Mead1989}. Here, $w_{ij}$ becomes a physical parameter such as electrical conductance, hydraulic resistance, or (in mechanical networks) an entry of the stiffness matrix that maps forces and displacements. Across both wetware and hardware, the shared computational essence is linear superposition: physics performs accumulation by default, with weights encoded in geometry or material parameters.

\subsubsection{Physical Interaction (The Non-Linear Regime)}
A purely linear network collapses to a single matrix multiplication and cannot represent functions requiring decision boundaries (e.g., XOR). Nonlinearity in PNNs therefore comes from exploiting material interactions rather than explicitly programming an activation function. In chemical systems, reaction kinetics provide nonlinear transforms directly: the law of mass action yields native multiplicative terms (e.g., rates proportional to $[A][B]$), enabling polynomial computation \cite{Soloveichik2010}, while enzymatic or catalytic saturation produces transfer curves that can closely match sigmoidal neuron models \cite{Hjelmfelt1991}. In solid/fluidic hardware, nonlinearity is often engineered via instabilities and thresholds. Mechanical elements can exhibit bifurcations (e.g., buckling) that act as sharp threshold functions and thus enable discrete decisions from continuous inputs \cite{Florijn2014}; similarly, microfluidic valves can switch state above a pressure threshold, implementing physical gating. In all cases, the computational role of nonlinearity is the same: critical thresholds break superposition ($f(a+b)\neq f(a)+f(b)$) and partition state space, enabling expressive function approximation.

\subsubsection{Relaxation as Optimization (The Solver)}
In conventional computing, optimization is implemented as an explicit iterative algorithm that updates parameters step-by-step. In many physical substrates, by contrast, ``optimization'' is an intrinsic tendency to relax toward lower free energy or lower internal stress. In wetware, this is exemplified by molecular constraint satisfaction: DNA strands spontaneously bind to Watson--Crick complements to minimize free energy, and massive parallelism arises because many candidate assemblies explore the solution space simultaneously. Adleman's seminal demonstration showed how such thermodynamic relaxation can be harnessed to solve combinatorial problems by encoding constraints into molecular binding and then letting the system settle \cite{Adleman1994}. In engineered hardware, analogous solver behavior appears in systems whose dynamics minimize an implicit cost function, such as spin glasses or elastic networks that relax to reduce frustration or stress. The connection between annealing in statistical physics and combinatorial optimization was formalized by Kirkpatrick et al.\ \cite{Kirkpatrick1983}. Conceptually, these systems admit a scalar ``energy'' (or Hamiltonian) that acts as a Lyapunov-like function with a non-increasing trajectory, $\frac{dH}{dt}\le 0$, so that mapping problem constraints onto the substrate's energy landscape turns computation into controlled relaxation rather than explicit iteration.

\subsection{Training Strategies for Physical Systems}
\label{subsec:training}

In deep learning, training is typically cast as minimizing a loss function via gradient descent, with backpropagation providing exact gradients $\nabla_{\theta}\mathcal{L}$ for each parameter $\theta$. Physical substrates do not natively expose a ``backward pass'': atoms neither store gradients nor provide explicit credit assignment. The resulting \emph{credit assignment problem} asks how a global output error can be attributed to local, embodied parameters (e.g., a specific channel conductance or beam stiffness). In practice, training strategies for Physical Neural Networks (PNNs) fall into three broad paradigms. At a high level, these strategies can be grouped into ex-situ digital-twin optimization, hybrid digital--physical training, and in-materio learning based on physical gradient approximation or related local update rules. In addition, when gradients are unavailable or unreliable, some substrates rely on closed box search procedures such as evolutionary optimization. Which of these approaches is most suitable depends strongly on substrate characteristics, including tunability, observability, relaxation time, and fabrication variability. The corresponding substrate-specific realizations are therefore discussed in the ``Training and Design Paradigms'' subsections of Sections \ref{sec:dna} to \ref{sec:iontronics}.

\subsubsection{In-Silico Training (Digital Twin / Ex-Situ Optimization)}
In-silico training decouples learning from execution by treating the device as an inference engine. A differentiable \emph{digital twin} (e.g., implemented in PyTorch or JAX) is trained with standard backpropagation, and the converged parameters are then transferred to the physical instance in a one-shot programming step. The main advantage is immediate access to mature optimizers and architectures (e.g., Adam/RMSprop, CNNs/RNNs) without changing the hardware. The central limitation is the \emph{reality gap}: unmodeled noise, tolerances, and nonlinearities can cause a model to overfit simulation artifacts and fail to transfer reliably \cite{Wright2022}. Moreover, accurate deployment typically requires calibration of the twin to the specific fabricated instance, reducing plug-and-play operation \cite{Hughes2019}. This strategy is therefore particularly attractive for substrates whose forward dynamics can be modeled with sufficient fidelity and whose parameters can be programmed reproducibly, but that provide little support for direct in-device adaptation.

\subsubsection{Hybrid Digital--Physical Training (Reservoir Computing)}
A second class comprises hybrid digital--physical schemes, in which optimization remains partly external while the physical substrate directly performs the forward computation. The most established example is Reservoir Computing (RC), which avoids training the substrate itself by exploiting its intrinsic nonlinear dynamics. The physical system acts as a fixed dynamical reservoir that projects inputs into a high-dimensional spatiotemporal state; only a simple linear readout (often digital linear regression) is trained to map these states to targets. Conceptually, the reservoir functions like a physical kernel, turning nonlinear separability into a linear readout problem. Because the reservoir is fixed, training is computationally light and typically admits a closed-form solution (no epoch-wise backpropagation). The approach is also substrate-agnostic: reservoirs have been demonstrated in diverse media such as water-bucket dynamics \cite{Fernando2003}, soft robotic bodies \cite{Nakajima2015}, and optical delay systems \cite{Vandoorne2014}. The trade-off is task sub-optimality: since the internal features are not optimized end-to-end, comparable accuracy may require substantially larger reservoirs \cite{Tanaka2019}. Hybrid schemes are thus especially appealing for substrates with rich native dynamics but limited fine-grained weight programmability, whereas they are less suitable when peak task-specific performance requires end-to-end optimization of the internal state. As a side remark: in this review, reservoir computing is treated as a cross-cutting training and architectural paradigm within physical neural computing rather than as a separate material class.

\begin{figure*}[t]
    \centering
    \includegraphics[width=\linewidth]{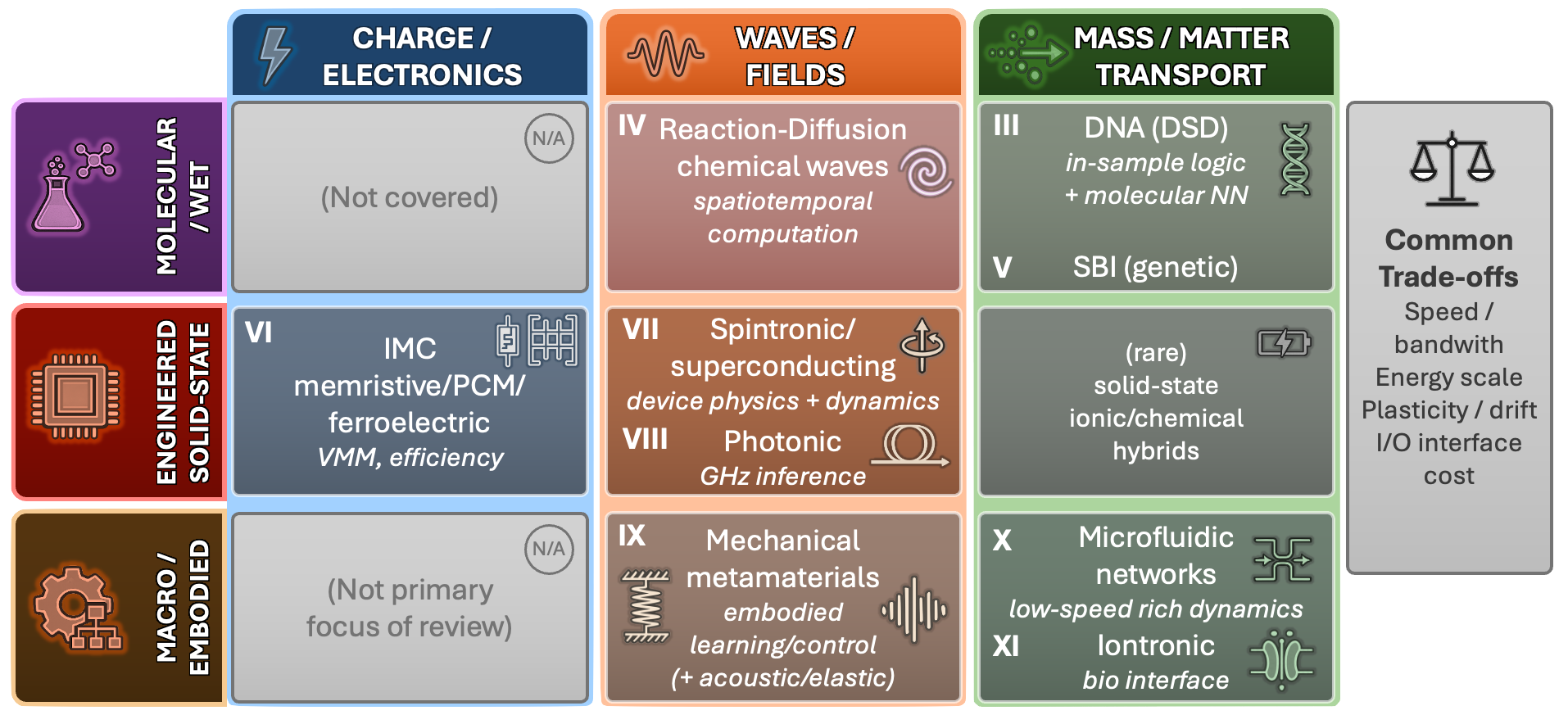}
    \caption{Landscape of physical substrates for computation and learning. The horizontal axis organizes approaches by the dominant coupling carrier (charge and electronics, waves and fields, and mass and matter transport), while the vertical axis groups them by substrate class and physical scale (molecular and wet, engineered solid-state, and macro and embodied systems). Each box summarizes a representative paradigm and indicates the corresponding chapter(s). Across this landscape, material physics determines which neural primitives are natively available in a substrate, whereas architecture determines how these primitives are arranged into scalable computational organizations.}
    \label{fig:taxonomy}
\end{figure*}

\subsubsection{In-Materio Learning (Physical Gradients / In-Situ Updates)}
The most ambitious paradigm is to perform learning \emph{within} the physics, thereby closing the reality gap by letting the device compute (or approximate) its own gradients. A prominent framework is \emph{Equilibrium Propagation (EqProp)}: the system is observed in a ``free'' phase (inputs only) and in a ``clamped'' phase (inputs plus a small output nudging), and differences between the resulting physical states yield an estimate proportional to the gradient \cite{Scellier2017}. In principle, this enables local, contrastive updates without storing a global gradient matrix \cite{Stern2021}. In practice, it raises substantial engineering challenges: implementing fast, stable, continuously tunable weights is difficult, and the need to physically relax (often twice per update) can make learning slow compared to purely digital training for small- to medium-scale systems \cite{Dillavou2022}. More broadly, this category includes in-situ update rules that use physically measured state differences or other local surrogate signals to approximate credit assignment directly in the substrate. Overall, PNN training remains less mature than large-scale digital backpropagation, but the combination of hybrid and in-situ approaches demonstrates that physically embodied learning is feasible under realistic constraints \cite{Momeni2025}. Such approaches are most promising for substrates that offer stable, repeatable, and sufficiently local update mechanisms, and remain difficult to realize when observability is poor, relaxation is slow, or device drift dominates the dynamics.

%%----------------------------------

\subsection{Taxonomy of Physical Substrates}
\label{subsec:taxonomy}

Physical neural substrates span everything from molecular test tubes to macroscopic mechanical lattices. To navigate this diversity without over-indexing on ``wet'' versus ``dry'' materials, we classify substrates by computational characteristics that mirror core abstractions in computer science: the \emph{data type} used for signal encoding, the \emph{memory architecture} used for plasticity, and the \emph{execution model} implied by their temporal dynamics.

\subsubsection{Signal Encoding (The Data Type)}
Most PNNs are intrinsically analog: information is carried by continuous thermodynamic variables (e.g., concentration, pressure, strain) whose effective precision is limited by noise rather than by a fixed word length. This maps naturally to continuous-time recurrent models, where graded states evolve under coupled dynamics; Hopfield’s analysis of networks with graded responses provides an early and influential formalization of this continuous regime \cite{Hopfield1984}. A smaller but important class of substrates is engineered for bistability, enabling digital-style encodings where the presence/absence of a molecular species or a snapped mechanical mode represents a Boolean state; DNA strand displacement is a canonical example, and Qian and Winfree demonstrated that it can implement arbitrary Boolean logic circuits \cite{QianWinfree2011}. Finally, event-driven encodings represent information primarily in \emph{timing} rather than amplitude, as in spiking or excitable media where propagating wavefronts act as discrete events; collision-based computing in reaction--diffusion systems illustrates how such impulse-based dynamics can support logic operations \cite{adamatzky2005reaction}.

\subsubsection{Plasticity Implementation (The Memory Architecture)}
In digital learning systems, weights are explicit variables updated by an external algorithm. In physical substrates, by contrast, ``memory'' is realized only insofar as material properties can be configured and retained. Some platforms are effectively read-only: geometry fixes the connectivity and coupling strengths during fabrication, yielding ASIC-like inference engines commonly used in reservoir-style setups; passive photonic reservoirs are a representative example \cite{Vandoorne2014}. Other systems are programmable but only through external intervention (e.g., mechanical adjustment, chemical titration), enabling reconfiguration without true autonomy. The strongest notion of plasticity arises when the substrate locally adapts in response to its own signal flux, providing a material analogue of Hebbian update rules; memristive devices exemplify this state-dependent memory, as highlighted in the foundational memristor formulation by Strukov et al. \cite{Strukov2008}.

\subsubsection{Temporal Dynamics (The Execution Model)}
A final differentiator is how computation uses time. In equilibrium-style systems, the input initializes the substrate, and the computation is the relaxation to a steady state; intermediate trajectories are not part of the ``output'' and are treated as transient. Molecular computing via thermodynamic equilibration is a classic illustration, including Adleman’s use of DNA to solve a Hamiltonian path instance by letting the system settle under designed constraints \cite{Adleman1994}. In dynamic systems, by contrast, information is encoded in the evolving trajectory itself, so outputs depend on input history through inertia, hysteresis, or internal state; this corresponds to sequential processing with fading memory. Jaeger’s ``echo state'' formulation provides a compact criterion for when such driven dynamics can support time-series processing reliably \cite{Jaeger2001}.

%The following chapters survey the field along the major physical regimes introduced here, moving from molecular and biochemical substrates to engineered solid-state platforms and embodied fluidic or mechanical systems. Figure \ref{fig:taxonomy} provides a compact overview of this landscape and situates the reviewed approaches within a common comparative mapping. While Figure \ref{fig:taxonomy} organizes the field primarily by substrate class and dominant physical carrier (as described in Subsections \ref{subsec:physics} and \ref{subsec:taxonomy}, learning strategies cut across these material categories. As already explained in Section \ref{subsec:training}, in the reviewed literature, ex-situ digital-twin optimization, hybrid reservoir/readout training, in-materio adaptation, and, where needed, closed box evolutionary search recur as the main learning regimes; details are  presented in the “Training and Design Paradigms” subsections of Sections \ref{sec:dna} to \ref{sec:iontronics}.

The following chapters survey the field along the major physical regimes introduced here, moving from molecular and biochemical substrates to engineered solid-state platforms and embodied fluidic or mechanical systems. Figure \ref{fig:taxonomy} provides a compact overview of this landscape and situates the reviewed approaches within a common comparative mapping. While Figure \ref{fig:taxonomy} organizes the field primarily by substrate class and dominant physical carrier (as described in Subsections \ref{subsec:physics} and \ref{subsec:taxonomy}), two further cross-cutting distinctions are important. First, material properties such as transport mechanism, switching behavior, intrinsic nonlinearity, stability, and temporal response determine which neural primitives are natively available in a given substrate, including weighted summation, thresholding, memory, or plasticity. Second, device and network architecture determine how these primitives are organized into computation, for example through crossbar arrays for parallel vector--matrix multiplication, coupled oscillators or reservoirs for dynamical processing, or spatially distributed cells, droplets, and reaction domains for propagating local interactions. As already explained in Section \ref{subsec:training}, learning strategies form an additional dimension that cuts across these material categories: in the reviewed literature, ex-situ digital-twin optimization, hybrid reservoir/readout training, in-materio adaptation, and, where needed, closed box evolutionary search recur as the main learning regimes; substrate-specific realizations are discussed in the ``Training and Design Paradigms'' subsections of Sections \ref{sec:dna} to \ref{sec:iontronics}.

% PART II ----------------------------------------------------------

\begin{figure*}[t]  
    \centering
    \includegraphics[width=\textwidth]{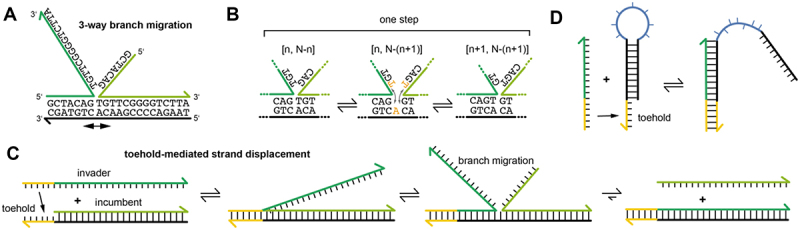}
%    \caption{Schematic representation of DNA strand displacement (toehold-mediated strand displacement). The diagram illustrates the process in four steps: (A) Initial state, where the template strand (blue) and protector strand (red) are hybridized, exposing a 'toehold'. (B) Binding of the input strand (green) to the toehold. (C) Branch migration, where the input strand successively displaces the protector strand. (D) Final state with the completely released protector strand.}
    \caption{Mechanisms of nucleic acid strand displacement. (A) Three-way branch migration: Two identical strands compete for binding to a common complementary template. (B) Stepwise kinetics: The migration proceeds as an unbiased random walk via the dissociation and reformation of single base pairs at the branch point. (C) Toehold-mediated strand displacement (TMSD): An invader strand binds to a single-stranded overhang ("toehold") to efficiently initiate branch migration and displace the incumbent strand. (D) Secondary structure invasion: An invader opens a hairpin loop via displacement, activating a previously sequestered sequence domain (blue). Reprinted from \cite{Simmel2023-eq}, licensed under CC BY-NC 4.0.}
    \label{fig:strand_displacement}
\end{figure*}

\section{Molecular Computing with DNA}
\label{sec:dna}

DNA is often viewed merely as a passive storage medium for genetic information (comparable to a hard drive). However, it possesses thermodynamic properties that make it also a powerful substrate for active computation. By stripping DNA of its biological context and treating it purely as a programmable polymer, researchers have engineered systems capable of executing complex neural logic in a test tube.

\subsection{The Substrate Principle}
The dominant mechanism for implementing molecular circuits is DNA strand displacement (DSD), a process driven by the thermodynamic tendency of DNA strands to maximize base-pairing. By designing "toehold" sequences that initiate branch migration, an input strand can displace a pre-hybridized output strand from a gate complex, effectively releasing a signal into the solution \cite{Chen2023,Simmel2023-eq} (see Fig.~\ref{fig:strand_displacement}). This mechanism functions analogously to signal transmission at a synapse, but operates purely through enthalpy-driven kinetics.

While early models abstracted DSD as digital logic, contemporary approaches exploit the continuous nature of reaction rates to implement analog computation, where signal strength is encoded in molar concentrations rather than binary presence. Recent advances in "molecular commutation" further abstract away the structural details, utilizing inverse design algorithms to compute optimal dissociation constant matrices for a desired network behavior \cite{202601.0088}. This shifts the design paradigm from manually assembling logic gates to mathematically optimizing the free-energy landscape of the reaction network.

\subsection{Mapping Neural Primitives}
To implement neural architectures in DNA, abstract mathematical operations must be mapped isomorphically to specific biochemical reaction kinetics.

\subsubsection{Weights and Signals}
Information is encoded in the continuous molar concentration of specific DNA strands. Synaptic weighting is realized through the stoichiometry of strand displacement reactions: the concentration of intermediate ``gate'' complexes modulates the gain between an input strand and the release of a downstream output strand \cite{QianWinfree2011}. Alternatively, DNA origami offers a spatially addressable substrate where binary or analog weights can be encoded via the physical arrangement of molecules on a nanoscale scaffold \cite{rothemund2006folding,zhan2023recent}. Summation occurs naturally through the parallel contribution of multiple reactant pathways to the production of a common output species.

\subsubsection{Activation and Non-Linearity}
To enable decision boundaries and deep logic, linear hybridization kinetics are augmented with molecular sequestration. By introducing high-affinity ``sink'' strands that bind and neutralize signal strands, the system enforces a concentration threshold below which no output is generated. This competitive annihilation process implements a function mathematically equivalent to the Rectified Linear Unit (ReLU) \cite{Soloveichik2010}. More advanced "winner-take-all" mechanisms group signals to compete against each other, allowing for efficient non-linear decision making in classification tasks \cite{Huang2025}.

\subsection{Architectures and I/O Interfaces}
The topology of DNA neural networks is not defined by physical placement but by the specificity of molecular interactions in a well-mixed solution. Research has evolved from static logic gates to dynamic systems capable of temporal processing and cellular integration.

Early architectures focused on implementing static input-output mappings akin to Boolean logic or feed-forward classifiers. The ``Winner-Take-All'' (WTA) networks demonstrated by Cherry and Qian utilize competitive hybridization to implement non-linear classification boundaries, successfully classifying 100-bit patterns (handwritten digits) \cite{CherryQian2018}. Similarly, 'seesaw gate' networks, which established a framework for scaling up digital circuit computation \cite{QianWinfree2014}, exploit reversible strand displacement to realize analog weighted summation and thresholding, enabling the construction of sigmoid-like transfer functions purely through mass-action kinetics \cite{QianWinfree2011}.

A long-time bottleneck for DNA computing has been signal degradation and ``crosstalk'' in deep networks. The CALCUL system (Classified Allosteric-toehold based Continuous and Ultra-accurate Learning) addresses this by introducing spatial isolation via magnetic beads. This separation allows for layer-by-layer processing and the implementation of convolutional operations (weight sharing) that were previously hindered by the ``soup'' nature of bulk solutions. This architecture has demonstrated high-accuracy classification of image patterns by physically preventing unwanted feedback loops \cite{Liu2025}.

To process time-varying signals or solve optimization problems, architectures have incorporated feedback loops and autocatalysis. Building on the foundational associative memory models proposed by Hopfield \cite{Hopfield1982}, DNA-based Hopfield networks utilize the energy landscape of the reaction system to store state information. Recent work demonstrated a discrete Hopfield network capable of solving combinatorial optimization tasks (such as Sudoku puzzles) by relaxing into a stable energy minimum representing the solution \cite{11045176}.

Data entry is typically achieved by introducing specific strand concentrations \cite{QianWinfree2011} or via light-sensitive molecular switches \cite{OHagan2023}. The computational result is read out through fluorescence kinetics using fluorophore-quencher pairs, where the intensity of the emitted light corresponds to the final output activation level \cite{QianWinfree2011}.

\subsection{Training and Design Paradigms}

Training strategies for DNA neural networks are evolving from static design to dynamic adaptation. At present, the dominant paradigm is ex-situ optimization via in-silico modeling and inverse design, whereas fully in-materio adaptation remains an emerging capability.

The currently dominant approach relies on in silico optimization, where the desired network behavior is modeled mathematically. Algorithms compute the optimal matrix of dissociation constants or reaction rates, which are subsequently translated into DNA sequences using inverse design tools [38]. The physical molecules are then synthesized to embody these pre-calculated weights.

Emerging paradigms demonstrate autonomous learning directly within the substrate. Recent experimental work has realized supervised learning in test tubes, where ‘‘training strands’’ (representing data and labels) interact with the network to physically adjust the concentration of weight molecules. This allows the system to update its decision boundaries purely through thermodynamic equilibration, without external digital processing \cite{cherry2025supervised}. This makes DNA particularly attractive when reaction pathways can be modeled and compiled into molecular programs with high specificity, but less suitable for fast, repeated, and fine-grained in-situ weight adaptation because kinetics remain slow and parameter tuning is tightly coupled to sequence design.

%\subsection{Representative Demonstrations}
%Experimental validations have confirmed the substrate's versatility. Cherry and Qian's implementation of a molecular classifier for handwritten digits (MNIST) proved that sophisticated non-linear decision boundaries can be engineered into an amorphous liquid medium \cite{CherryQian2018}. In the domain of biosensing, DNA logic circuits have been applied to multiplexed diagnostics, integrating multiple microRNA signals to compute diagnostic decisions with higher specificity than single-marker assays \cite{Sanjabi2019,Okumura2022-jx,Takiguchi2025}. Inside living cells, synthetic gene circuits can leverage transcriptional regulation to implement neuromorphic computation, for example via ``perceptgenes'' in bacteria \cite{Rizik2022} and via CRISPR-dCas9/gRNA architectures with RNA sequestration that realize perceptron-like threshold units \cite{Leon2025}.

\subsection{Representative Demonstrations}
Representative demonstrations of molecular neural computation span cell-free pattern classification, multiplexed diagnostic decision-making, and intracellular neuromorphic control. Cherry and Qian's molecular classifier for handwritten digits (MNIST) showed that complex non-linear decision boundaries can be realized in a well-mixed liquid medium, demonstrating that DNA strand-displacement networks can perform genuine pattern classification rather than merely simple logic operations \cite{CherryQian2018}. In biosensing, molecular logic circuits have been used for multiplexed diagnostic decision-making by integrating several microRNA inputs into a single molecular output, thereby improving specificity over single-marker assays \cite{Sanjabi2019,Okumura2022-jx,Takiguchi2025}. Moving from cell-free systems into living cells, synthetic gene circuits further show that neuromorphic computation can be embedded into cellular regulation: ``perceptgenes'' in bacteria implement transcriptionally regulated input integration \cite{Rizik2022}, while CRISPR-dCas9/gRNA architectures combined with RNA sequestration realize perceptron-like threshold units for classification and regression tasks \cite{Leon2025}. Taken together, these studies show that the same general molecular design principles can support pattern recognition, diagnostic inference, and intracellular computation across increasingly application-near settings.

\subsection{Engineering Constraints}
The transition from theoretical molecular algorithms to physical implementation imposes specific thermodynamic and kinetic constraints that define the operating regime of DNA computing.

The most distinct feature of the DNA substrate is the extreme disparity between information density and propagation speed. Molecular systems offer storage densities theoretically approaching $10^{19} \text{ bits}/\text{cm}^3$, with a single microliter capable of containing trillions of independent processing threads \cite{QianWinfree2011}. This allows for massive parallelism at thermodynamic limits of energy efficiency. However, operation bandwidth is diffusion-limited, typically restricting complex strand-displacement cascades to timescales of minutes or hours to reach equilibrium \cite{CherryQian2018}.

Another long-time bottleneck for molecular logic has been the ``single-use'' nature of the hardware. However, recent developments in 2025 have begun to mitigate this constraint. Sun et al.\ demonstrated that non-complementary DNA neural networks can decouple the ``weight'' molecules from the input signals, allowing inputs to be chemically washed from the system while preserving the trained parameters \cite{Sun2025}. Furthermore, ``heat-rechargeable'' circuits have been proposed that utilize thermal cycling not merely for resetting, but to actively drive the system into high-energy non-equilibrium states, effectively recharging the circuit's computational potential \cite{song2025heat}.

\subsection{Position within Physical Neural Computing}
DNA-based neural systems occupy an extreme position in the landscape of physical neural computing. They operate at the smallest physical scales and lowest characteristic bandwidths, while offering exceptionally high intrinsic parallelism. Architecturally, they realize chemically embedded computation: weighted summation and nonlinearity are implemented directly by reaction stoichiometry, distinguishing them from electronic or photonic substrates, where physical processes accelerate numerical operations but do not replace symbolic computation. Their relevance lies in a complementary regime characterized by direct processing of molecular inputs, biochemical compatibility, and extreme energy efficiency.

%-----------------------------------------------

\section{Chemical and Reaction-Diffusion Systems}
\label{sec:chemical}

While DNA computing typically relies on thermodynamic equilibrium and the static storage of genetic information, chemical computing exploits the non-equilibrium dynamics of dissipative structures. By utilizing reaction-diffusion (RD) systems and active matter, computation is shifted from the passive readout of code to the active propagation of information in space and time \cite{adamatzky2005reaction}. In this paradigm, the material substrate behaves as a continuous excitable medium, capable of processing information through wave propagation and synchronization rather than symbolic manipulation.

\subsection{The Substrate Principle}
Unlike the discrete logic of silicon or the hybridization kinetics of DNA, chemical computers operate by physically embodying partial differential equations. The fundamental ``machine language'' of this substrate is the Reaction-Diffusion equation:
\begin{equation}
    \frac{\partial u}{\partial t} = D\nabla^{2}u + R(u)
\end{equation}
where $u$ represents the vector of chemical concentrations. The diffusion term $D\nabla^{2}u$ mediates spatial coupling (signal transport), while the reaction term $R(u)$ dictates the non-linear local kinetics (processing) \cite{adamatzky2005reaction}.

Historically investigated in macroscopic vessels, recent advances have shifted towards ``On-Chip Chemical Computing''. Agostini et al. demonstrated the integration of oscillating Belousov-Zhabotinsky (BZ) reaction chambers directly onto silicon microchips, effectively creating hybrid architectures where the chemical medium performs massive parallel processing while electronic overlays handle I/O \cite{agostini2025towards}. A defining thermodynamic feature of these systems is their dissipative nature: computation is a continuous dynamic process maintained by a constant flux of reagents. Unlike non-volatile memory, if the energy supply ceases, the system relaxes to equilibrium and the computational state is lost \cite{Baltussen2024}.

\subsection{Mapping Neural Primitives}
To engineer intelligence into fluid substrates, the abstract mathematical operations of artificial neural networks must be mapped to the kinetic laws governing chemical reactions.

\subsubsection{Signals and Weights}
In chemical neural networks, information is typically encoded in the molar concentration of molecular species or the phase of an oscillatory wave. Synaptic weights, which modulate signal transmission, are physically instantiated by reaction rate constants ($k$) and diffusion coefficients ($D$). While traditional homogeneous systems rely on intrinsic kinetic parameters, recent microfluidic implementations allow for spatially programmable weights by altering channel geometries or utilizing responsive hydrogels that modulate local diffusion rates \cite{Zhu2025}.

A key advantage of this chemical signaling is its inherent robustness. Comparative studies suggest that chemical synapses exhibit superior anti-interference characteristics compared to electrical coupling, naturally filtering high-frequency noise and stabilizing synchronization in a manner that is difficult to replicate in standard electronics \cite{li2025anti_interference_neuronal}.

\subsubsection{Activation and Non-Linearity}
The linear accumulation of signals is handled natively by diffusion and mixing. The essential non-linearity required for deep computation arises from the Law of Mass Action and enzyme kinetics. Since reaction rates are proportional to the product of reactant concentrations, chemical systems provide a native multiplication operation, enabling the computation of polynomial functions \cite{Soloveichik2010}.

Furthermore, saturation effects in enzymatic reactions yield concentration curves mathematically identical to sigmoidal transfer functions. Sharp thresholding behaviors, equivalent to the Rectified Linear Unit (ReLU), are implemented via chemical titration mechanisms, where a signal molecule must neutralize a specific inhibitor concentration before triggering a downstream reaction \cite{Hjelmfelt1991}.

\subsection{Architectures and I/O Interfaces}
The architectural landscape of chemical computing has shifted from rigid, geometrically constrained logic circuits to amorphous, self-organizing dynamical systems.

Early implementations focused on collision-based computing, where traveling chemical wavefronts in excitable media (such as the above-mentioned BZ reaction) act as information carriers. Logical operations are realized through the precise geometrical routing of these waves: annihilation of colliding wavefronts implements NOT or XOR functions, while channel junctions perform OR logic \cite{adamatzky2005reaction,TothShowalter1995}. While conceptually foundational, these systems face scalability challenges due to the need for precise spatial structuring and the slow propagation speed of diffusion waves.

To overcome the constraints of explicit circuit design, the field has largely adopted the reservoir computing paradigm. Here, the ``messy'' complexity of non-linear reaction networks, such as the autocatalytic Formose reaction or electrochemical oscillators, serves as a high-dimensional projection space. The chemical substrate naturally maps low-dimensional inputs (e.g., flow rates) into a vast state space of intermediate concentrations, allowing a simple trained linear readout to extract complex temporal features \cite{Baltussen2024,Cucchi2021}. This approach %exploits the ``edge of chaos,''
utilizes the intrinsic thermodynamic relaxation of the medium as a computational resource.

Moving beyond bulk solutions, compartmentalized architectures utilize networks of interacting droplets or ``active matter'' swarms. In droplet-based systems, individual reaction vessels act as discrete nodes coupled via diffusion across lipid interfaces, forming programmable chemical lattices \cite{Villar2013-gb}. Conversely, active matter reservoirs employ self-propelled particles (e.g., bacteria or Janus colloids) whose collective hydrodynamics create reconfigurable network topologies, introducing advection as a signal transport mechanism to potentially surpass diffusion limits \cite{gaimann2025}.

\subsection{Training and Design Paradigms}
Training chemical systems presents a unique challenge: physical atoms do not store gradients, and the ``credit assignment problem'' is complicated by the opacity of the reaction mixture. Accordingly, chemical substrates are currently dominated by ex-situ digital-twin modeling and hybrid reservoir-style training, while closed box evolutionary optimization is often used when internal kinetics are too complex for direct gradient-based design.

To bridge the ``reality gap'' between simulation and wetware, recent approaches utilize Physics-Informed Neural Networks (PINNs) and Neural Ordinary Differential Equations (ODEs) \cite{Chen2018NeuralODE_arXiv}. Frameworks such as SPIN-ODE embed the governing chemical rate equations directly into the training loop, inferring the underlying kinetics from observational trajectories. This allows the construction of ``digital twins'' that are robust to noise and experimental variability, enabling offline optimization of control parameters that transfer reliably to the physical reactor \cite{peng2025spinode,raissi2019pinn}. This paradigm is particularly attractive when reaction kinetics can be modeled with sufficient fidelity, but it remains limited by partial observability, stochastic fluctuations, and calibration effort for specific experimental instances.

In the reservoir regime, training is restricted to the readout layer (typically a mass spectrometer or optical sensor), which is optimized via linear regression. For the substrate itself, evolutionary algorithms are often employed to select reagents or droplet topologies that maximize the ``richness'' or entropy of the chemical dynamics, effectively evolving the material's computational capacity without requiring a differentiable model of the complex internal kinetics \cite{Parrilla-Gutierrez2020-oo}. Such hybrid and closed box strategies are well matched to chemically rich but only weakly observable substrates, although they generally sacrifice end-to-end optimality and fine-grained in-situ adaptation.

\subsection{Representative Demonstrations}
\begin{figure}
    \centering
    \includegraphics[width=\columnwidth]{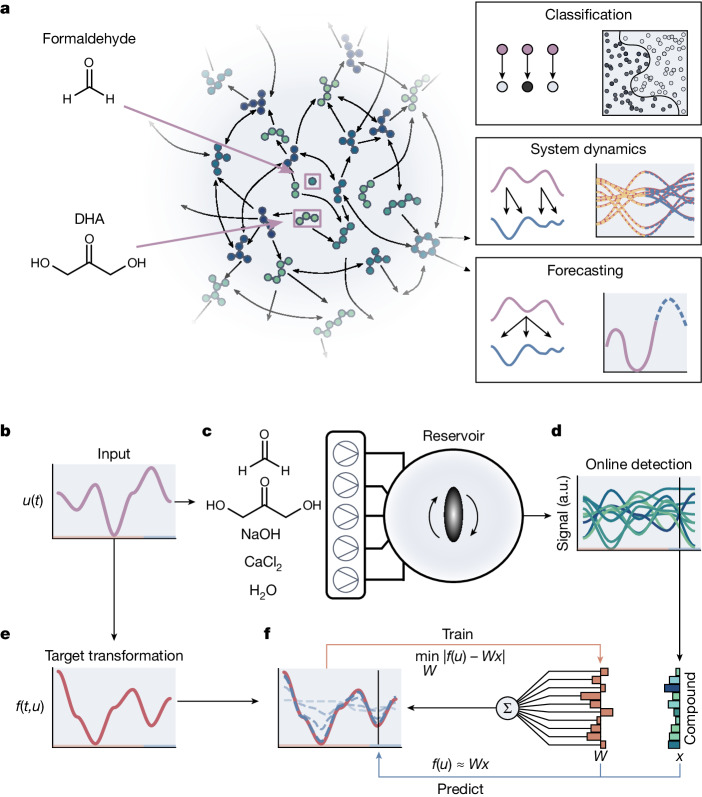}
    \caption{Architecture of a Chemical Reservoir Computer based on the Formose Reaction. (a) The chemical reaction network functions as a physical reservoir (a fixed, non-linear hidden layer). It projects low-dimensional inputs (reactants) into a high-dimensional chemical state space. This intrinsic non-linearity enables the linear separability of complex computational tasks, such as classification and time-series forecasting. (b--f) Signal Flow and Readout Training. A time-dependent input signal $u(t)$ modulates the inflow rates of reagents into a continuous stirred-tank reactor (CSTR). The resulting non-linear chemical dynamics (the state vector $x$) are monitored in real-time via mass spectrometry. Consistent with the reservoir computing paradigm, training is restricted to the readout layer: a linear weight matrix $W$ is optimized via regression to map the high-dimensional chemical state to the target output function ($y \approx Wx$). Reprinted from \cite{Baltussen2024}, licensed under CC BY 4.0.}
    \label{fig:formose_reservoir}
\end{figure}

Representative demonstrations of chemical neural computation span four regimes: temporal reservoir computing, embodied chemo-mechanical control, spatial analog problem solving, and structured oscillatory processing. The already mentioned ``Formose Reservoir'' exemplifies the use of molecular chaos for computation. Baltussen et al. showed that the autocatalytic polymerization of formaldehyde, previously regarded as an uncontrollable side reaction, exhibits rich transient dynamics that can serve as a physical reservoir. By modulating the inflow rates of reagents in a continuous stirred-tank reactor, the system generates characteristic molecular states that enable chaotic time-series prediction and non-linear classification, outperforming linear models by exploiting the intrinsic complexity of the chemical substrate \cite{Baltussen2024}. Figure \ref{fig:formose_reservoir} illustrates this principle and highlights how low-dimensional inputs are mapped into a high-dimensional chemical state space from which only the readout layer is trained.

In the domain of soft robotics, chemical intelligence enables control without microcontrollers. Self-oscillating gels driven by the Belousov--Zhabotinsky reaction have been engineered to function simultaneously as sensor, controller, and actuator. These materials exhibit autonomous peristaltic motion and phototactic behavior, thereby embodying the control loop directly in the chemo-mechanical properties of the material \cite{Zhu2025}.

The classical application of excitable chemical media remains spatial computing. Reaction-diffusion waves naturally explore geometric spaces in parallel. Steinbock et al. and later Lagzi et al. demonstrated that chemical wavefronts can solve maze-navigation and Voronoi-tessellation problems: the wave propagates through all possible paths simultaneously, and the first wavefront reaching the exit identifies the optimal trajectory \cite{Steinbock1995,Lagzi2010}.

In contrast to the unstructured dynamics of the Formose system, Parrilla-Gutierrez et al. demonstrated a programmable chemical processor based on a structured Belousov--Zhabotinsky cell array. Their system realized a chemical autoencoder for pattern recognition and encoded addressable memory states directly in the oscillatory phase and amplitude of the reaction vessels \cite{Parrilla-Gutierrez2020-oo}.

%----------------------------------------------

\subsection{Engineering Constraints}
The practical deployment of chemical neural networks is governed by thermodynamic necessities and interface limitations that distinguish them sharply from solid-state electronics.

Unlike DNA computing (which often relies on equilibrium states) or CMOS (where static power can be minimized), reaction-diffusion systems are inherently dissipative structures. Computation is a dynamic process maintained only by a continuous flux of free energy (reagents). This creates a fueling constraint: the system requires a constant supply of chemical ``fuel'' to sustain the non-equilibrium states necessary for wave propagation and oscillation. Consequently, these systems are best suited for environments where chemical energy is abundant and naturally replenished, such as within biological fluids or waste streams, rather than in battery-constrained portable devices \cite{Baltussen2024}.

The internal parallelism of a chemical droplet is immense (processing on the order of $10^{20}$ molecular interactions simultaneously). However, accessing this state remains a bottleneck. Reading out the complex spectrum of a Formose reservoir typically requires mass spectrometry, a slow and bulky process that negates the miniaturization benefits. However, the recent move toward ``lab-on-a-chip'' architectures that integrate electronic sensors directly with BZ reaction chambers offers a path to mitigate this transduction tax, enabling real-time electrical readout of chemical states \cite{agostini2025towards}.

\subsection{Position within Physical Neural Computing}
Chemical neural systems occupy a distinct position between molecular and macroscopic physical computation. They operate at spatial scales ranging from micrometers to centimeters and at time scales set by diffusion and reaction kinetics, resulting in low bandwidth but rich spatiotemporal dynamics.

%In contrast to DNA-based networks, which encode computation in discrete molecular interactions, chemical systems perform neural operations through continuous concentration fields and propagating wavefronts. Weighted interactions and non-linear activation emerge from diffusion coupling and chemical feedback loops, making these substrates particularly suited for spatial pattern processing. Their primary relevance lies not in high-speed inference, but in domains where computation over space and time is intrinsic to the problem, such as morphological pattern formation, distributed sensing, and unconventional control in chemically active environments.

In contrast to DNA-based networks, chemical systems implement neural operations through continuous concentration fields and propagating wavefronts rather than discrete molecular interactions. Weighted interactions and nonlinear activation arise from diffusion coupling and chemical feedback loops. These properties make chemical substrates especially suitable for spatial pattern processing. Their main relevance is therefore not high-speed inference, but applications in which computation over space and time is intrinsic to the problem, such as morphological pattern formation, distributed sensing, and control in chemically active environments.

%------------------------------------------------------------------

\section{Synthetic Biological Intelligence}
\label{sec:sbi}

%Synthetic Biological Intelligence (SBI) represents a qualitative transition from chemical computation to computation embodied in living matter. Whereas molecular and chemical substrates function as programmable but passive media, SBI exploits the intrinsic agency of biological systems: metabolism, homeostasis, self-repair, and plasticity become computational resources. In this paradigm, learning is not externally imposed by algorithmic weight updates alone, but emerges from biologically conserved mechanisms for adaptation and free-energy minimization \cite{Smirnova2023}.
Synthetic Biological Intelligence (SBI) extends chemical computation into living matter. Unlike molecular or chemical substrates, which are programmable but passive, SBI exploits intrinsic biological functions such as metabolism, homeostasis, self-repair, and plasticity as computational resources. As a result, learning is not only imposed externally through weight updates, but can also emerge from biological mechanisms of adaptation and free-energy minimization \cite{Smirnova2023}.

\subsection{The Substrate Principle}
Research in SBI currently focuses on three distinct substrate classes representing increasing levels of biological complexity.
At the genetic level, engineered bacterial consortia implement distributed logic. Early work by Elowitz and Leibler demonstrated synthetic oscillatory networks (``Repressilator''), while Tamsir et al. and Danino et al. realized synchronized quorum sensing and Boolean gates within bacterial populations \cite{Elowitz2000,Tamsir2011,Danino2010}.
At the morphological level, the slime mold \textit{Physarum polycephalum} acts as a macroscopic, multinucleate single cell capable of solving spatial optimization tasks by dynamically reconfiguring its protoplasmic tube network \cite{Tero2010,Adamatzky2010}.
Most recently, ``Organoid Intelligence'' (OI) has emerged, utilizing three-dimensional neural cultures derived from human induced pluripotent stem cells (iPSCs). These organoids develop dense and interconnected neuronal networks with spontaneous spiking activity and long-term potentiation. Among the SBI substrates discussed here, they most closely mimic the cytoarchitecture of the human cortex \cite{Smirnova2023,Cai2023-lr}.

\subsection{Mapping Neural Primitives}
Living substrates realize neural primitives through physical growth, electrophysiology, and biochemical regulation.

\subsubsection{Weights and Plasticity}
In \textit{Physarum}, the ``weight'' of a connection is physically instantiated by the diameter of the protoplasmic tube, which determines the cytoplasmic flux. This adaptive resizing follows a positive feedback loop: $\Delta w_{ij} \propto \text{flux}_{ij}$ \cite{Tero2010}. In neuronal organoids, weights correspond to synaptic efficacy. Their adaptation follows biologically conserved rules such as spike-timing-dependent plasticity (STDP), in which the temporal correlation between pre- and post-synaptic activity determines potentiation or depression \cite{Bi1998}.

\subsubsection{Activation and Dynamics}
Activation in slime molds is continuous and oscillatory, driven by rhythmic contraction waves. Organoids, conversely, operate in the spiking regime, encoding information in discrete action potentials and population firing rates. In genetic circuits, activation is transcriptional, defined by the concentration of repressor or activator proteins binding to DNA promoters \cite{Elowitz2000}. Going beyond digital logic, Daniel et al. demonstrated that these transcriptional networks can also be engineered to perform synthetic analog computation, such as addition and division in the logarithmic domain, directly within living cells \cite{Daniel2013}.

\subsubsection{Optimization via Homeostasis}
Unlike artificial systems that minimize an explicit loss function, biological networks appear to optimize for homeostasis and predictability. The Free Energy Principle proposes that living systems adapt their internal states to minimize the difference between predicted and sensed sensory inputs (sensory surprise), effectively performing active inference to maintain their structural integrity \cite{Friston2010}.

\subsection{Architectures and I/O Interfaces}
Architectures in SBI are typically emergent rather than lithographically defined, requiring sophisticated interfaces to guide computation.

\textit{Physarum} naturally forms spatial graphs. When food sources are arranged to represent nodes in a geometric problem, the organism relaxes into a Steiner tree configuration, effectively computing the shortest path that connects all points while balancing metabolic cost against transport efficiency \cite{Tero2010}.

Brain organoids are typically interfaced using High-Density Microelectrode Arrays (HD-MEAs). These devices provide thousands of bidirectional channels that allow for both the recording of population spikes and the delivery of patterned electrical stimulation \cite{Obien2015}. The organoid functions as a recurrent reservoir, transforming input stimulation patterns into high-dimensional spatiotemporal readouts \cite{Cai2023-lr}.

To enable systematic experimentation at scale, Jordan et al.\ introduced an open and remotely accessible neuroplatform that integrates long-term electrophysiological recording, automated microfluidic control, continuous environmental regulation, and closed-loop stimulation for large numbers of neural organoids \cite{Jordan2024-jz}.

\subsection{Training and Design Paradigms}
Since the internal state of a living cell is largely opaque (``black box''), training relies on closed-loop interaction and environmental shaping. Accordingly, living substrates are currently dominated by hybrid closed-loop training and ex-situ circuit design, whereas direct in-materio credit assignment remains largely out of reach.

The most promising training paradigm involves placing the biological culture in a closed feedback loop with a simulated environment (``DishBrain''). Sensory inputs are provided through electrical stimulation, and motor outputs are read from neural activity. The culture then receives structured feedback: predictable stimuli for correct behavior and noisy stimuli for incorrect behavior. This closed loop encourages the system to reorganize its connectivity toward more predictable states \cite{Kagan2022}. This paradigm is attractive because it exploits the adaptive plasticity of living neural tissue without requiring an explicit internal model. However, it offers only limited interpretability, weak control over credit assignment, and modest reproducibility across biological instances.

For bacterial systems, ``training'' corresponds to the iterative design of plasmid sequences. Combinatorial logic is implemented by layering transcriptional repressors (e.g., NOR gates), allowing for the construction of asynchronous digital circuits inside living cells \cite{Tamsir2011,Brenner2008}. To scale such designs, Nielsen et al. introduced ``Cello'', a design automation framework that compiles abstract Verilog specifications directly into functional DNA sequences, thereby solving the problem of manual interference between gates \cite{Nielsen2016}. In such systems, the dominant paradigm is therefore closer to ex-situ digital design and compilation than to learning in the neural-network sense: biological function is programmed into the genetic architecture before deployment. This is powerful for constructing reliable logic behavior, but less suitable for fast online adaptation or repeated gradient-based weight updates.

\subsection{Representative Demonstrations}
%Demonstrations of SBI have successfully moved from passive observation to goal-directed behavior.

%Kagan et al. demonstrated that human cortical neurons cultured on an MEA could learn to play the arcade game ``Pong''. Through the application of the Free Energy Principle (providing predictable feedback for successful hits), the system demonstrated rapid learning and adaptation in minutes, outperforming control groups without feedback \cite{Kagan2022}. The architecture of the experiment is shown in Figure \ref{fig:kagan_pong}.

\begin{figure}
    \centering
    \includegraphics[width=\columnwidth]{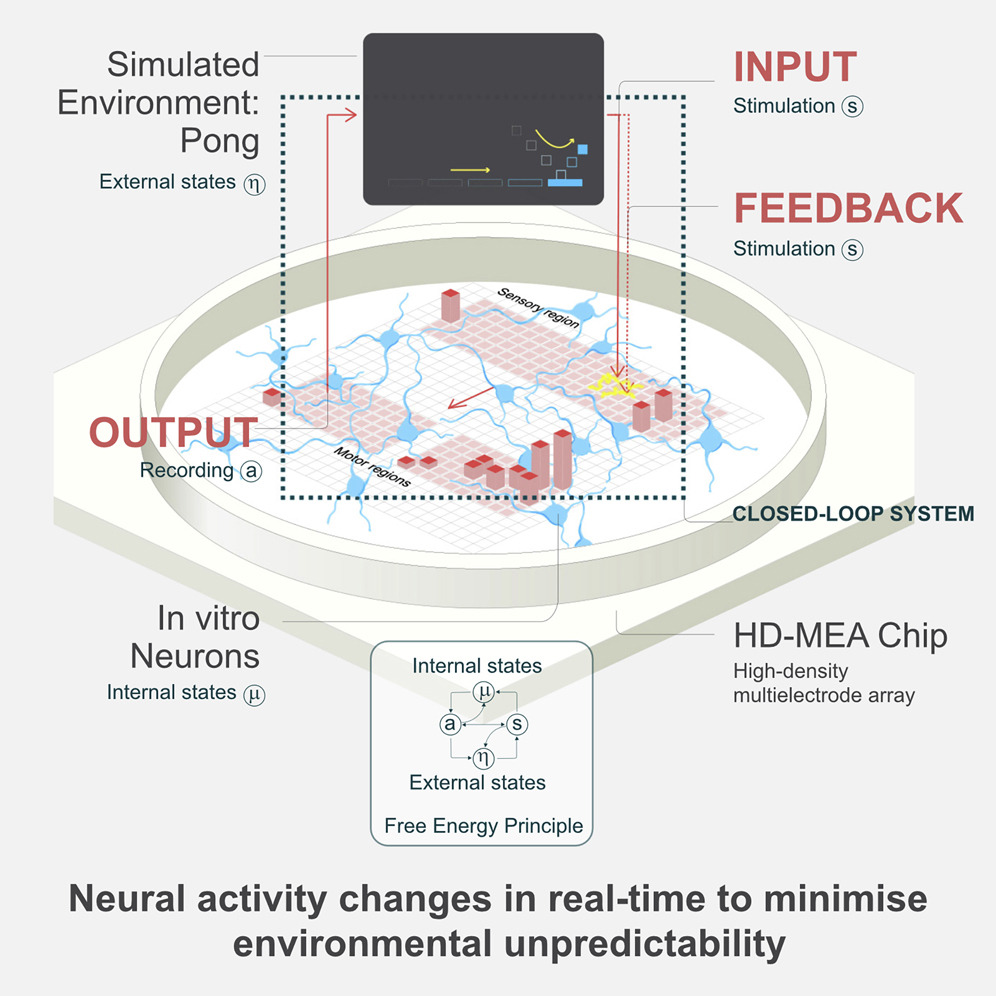}
    \caption{Schematic of the closed-loop ``Pong'' experiment. Human cortical neurons cultured on a high-density multielectrode array (HD-MEA) are bidirectionally coupled to a simulated game environment: game states are encoded as electrical stimulation patterns, neural responses are recorded and decoded as motor output, and structured feedback is returned to the culture. The setup illustrates how synthetic biological intelligence can realize real-time learning through closed-loop interaction aimed at reducing environmental unpredictability. Reprinted from \cite{Kagan2022}, licensed under CC BY 4.0.}
    \label{fig:kagan_pong}
\end{figure}

Representative demonstrations of SBI span two characteristic regimes: closed-loop learning in neuronal cultures and self-organized spatial optimization in living networks. Kagan et al. demonstrated that human cortical neurons cultured on a multielectrode array can be embedded into a real-time sensorimotor loop with the arcade game ``Pong'' \cite{Kagan2022}. 
Electrical stimulation encoded the game state, neural activity was decoded as motor output, and structured feedback drove rapid adaptation within minutes, showing that living neural tissue can perform goal-directed learning when coupled to an interactive environment. Figure \ref{fig:kagan_pong} illustrates this architecture and highlights that sensing, computation, and adaptation emerge jointly from the biological culture and its interface.

%In a classic experiment, \textit{Physarum} was used to replicate the design of the Tokyo railway system. By placing oat flakes at locations corresponding to major cities, the organism grew a network that matched the efficiency and fault tolerance of the actual engineered rail network \cite{Tero2010}.

In a complementary form of SBI, \textit{Physarum was used to reproduce the topology of the Tokyo railway system} \cite{Tero2010}. When oat flakes were placed at locations corresponding to major cities, the organism grew a transport network that balanced efficiency, connectivity, and fault tolerance, closely matching the engineered rail layout. This demonstration is representative because it shows that living matter can solve a nontrivial spatial optimization problem through growth and adaptive reconfiguration rather than through explicitly programmed digital control.

\subsection{Engineering Constraints}
The utilization of living matter introduces unique engineering challenges centered on maintenance and reproducibility.
First, the energetic advantage of biological tissue (which operates at orders of magnitude lower power than silicon) is offset by the significant overhead of life support systems (sterile environments, nutrient perfusion).
Second, biological variability remains a critical bottleneck. Unlike transistors, no two organoids are identical in cytoarchitecture or connectome, making standardized performance difficult to guarantee.
Third, the "I/O bandwidth" is constrained by the bio-electronic interface; while the tissue processes information massively in parallel, the number of electrodes available for reading and writing state remains limited compared to the number of neurons \cite{Smirnova2023,Obien2015}.

\subsection{Position within Physical Neural Computing}
SBI occupies the extreme end of the physical intelligence spectrum, characterized by maximal adaptivity and minimal programmability. It offers unparalleled learning efficiency and structural plasticity but lacks the speed and deterministic reliability of electronic hardware. As such, SBI is best viewed as a complementary paradigm for applications requiring high sensitivity, biocompatibility, or the study of intelligence itself in its native substrate.

%-----------------------------------------------------------------

\section{Resistive and Ferroelectric In-Memory Computing}
\label{sec:memristive_pcm}

Memristive, phase-change, and ferroelectric systems implement computation through the physics of non-volatile conductance and polarization modulation. 
Unlike von Neumann architectures that shuttle data between distinct processing and storage units, these substrates realize ``in-memory computing'' (IMC), where the memory element itself performs the computation. 
By organizing resistive switching devices, which are typically based on filamentary metal-oxides (ReRAM), chalcogenide phase-change materials (PCM), or ferroelectric oxides (FeRAM), into dense crossbar arrays, vector-matrix multiplication (VMM) is executed directly through the relaxation of electrical currents. 
This offers a path to massive parallelism and energy efficiency that circumvents the memory wall \cite{Strukov2008,Sebastian2020}.
Recently, this paradigm has expanded to ``in-sensor computing,'' where ferroelectric materials simultaneously sense physical stimuli and perform synaptic weighting, eliminating the analog-to-digital conversion overhead at the edge \cite{Tan2024_Vision}.

\subsection{The Substrate Principle}
The fundamental principle relies on the tunable electrical state of a two-terminal device, governed by history-dependent hysteresis loops.

In ReRAM, the conductance is modulated by the field-driven migration of oxygen vacancies, which form conductive filaments. 
In PCM, programming pulses induce a reversible phase transition between a highly conductive crystalline state and a highly resistive amorphous state \cite{Burr2014}. 
Beyond resistive switching, nanoscale spintronic oscillators have also been shown to serve as non-linear neurons \cite{Torrejon2017}.

Ferroelectric devices (FeFETs, FTJs) store information in the remanent polarization of the material lattice.  
Unlike ReRAM, which relies on atomic filament formation, ferroelectrics rely on the displacement of ions within the unit cell. 
Applying an electric field switches the polarization domains. 
Crucially, partial switching of these domains allows for the storage of continuous analog values. 
In their review, Khan et al. identify this technology as a key enabler for future non-volatile logic, arguing that HfO$_2$-based FeFETs provide the necessary scalability to merge logic and memory at the transistor level \cite{Khan2020-sh}.
Recent advances in flexible electronics utilize ultra-thin Indium Tin Oxide (ITO) channels combined with ferroelectric gates to realize transparent, high-performance synaptic devices suitable for wearable applications \cite{Li2024-aw}.
Furthermore, the movement of ``Domain Walls'' (interfaces between regions of different polarization) exhibits rich, non-linear dynamics that can be exploited for reservoir computing, where the position of the wall naturally encodes the reservoir state \cite{EverschorSitte2024}.

When arranged in a crossbar topology, all these devices exploit Kirchhoff’s circuit laws to perform analog arithmetic. 
An input voltage vector applied to the rows generates currents proportional to the conductance (weight) at each crosspoint (Ohm’s Law). 
These currents sum naturally along the columns (Kirchhoff’s Current Law), physically computing the dot product 
$\sum_j V_i G_{ij}$ in a single time step, regardless of the array size \cite{Sebastian2020}.

\subsection{Mapping Neural Primitives}
To function as a neural network, the analog physics of the crossbar must be mapped to the linear and non-linear operations of deep learning.

\subsubsection{Weights and Summation}
Synaptic weights are encoded as the analog conductance values ($G$) or polarization states ($P$) of the devices. 
Since conductance is strictly positive, signed weights are typically realized using a differential pair architecture, where the effective weight is the difference between two conductances ($W = G^+ - G^-$). 
In ferroelectric tunnel junctions (FTJs), the weight is determined by the tunneling probability, modulated by the height of the ferroelectric barrier potential.

\subsubsection{Activation and Nonlinearity}
Unlike chemical or mechanical substrates, where non-linearity is often intrinsic to the material, memristive arrays typically perform only the linear transformation. 
The non-linear activation function (e.g., ReLU, Sigmoid) is usually implemented at the array periphery using mixed-signal circuits. 
However, ferroelectric materials offer a distinct advantage here: the intrinsic $P-E$ (Polarization-Electric Field) hysteresis loop is naturally sigmoidal. 
Recent reviews highlight how this property, combined with topological textures like domain walls, allows for fully passive non-linear layers suitable for reservoir computing \cite{EverschorSitte2024}.

\subsection{Architectures and I/O Interfaces}
The architectural paradigm is shifting from pure digital accelerators to heterogeneous systems that integrate sensing and computing.

To mitigate device variability and limited dynamic range, architectures employ redundancy. 
A single synaptic weight may be represented by multiple physical devices (multi-memristive synapses). 
Arbitration logic or averaging mechanisms are then used to combine these conductances, effectively increasing the bit-precision and reducing the impact of stochastic switching noise inherent to the nanoscale physics \cite{Boybat2018}.

\begin{figure}
    \centering
    \includegraphics[width=\columnwidth]{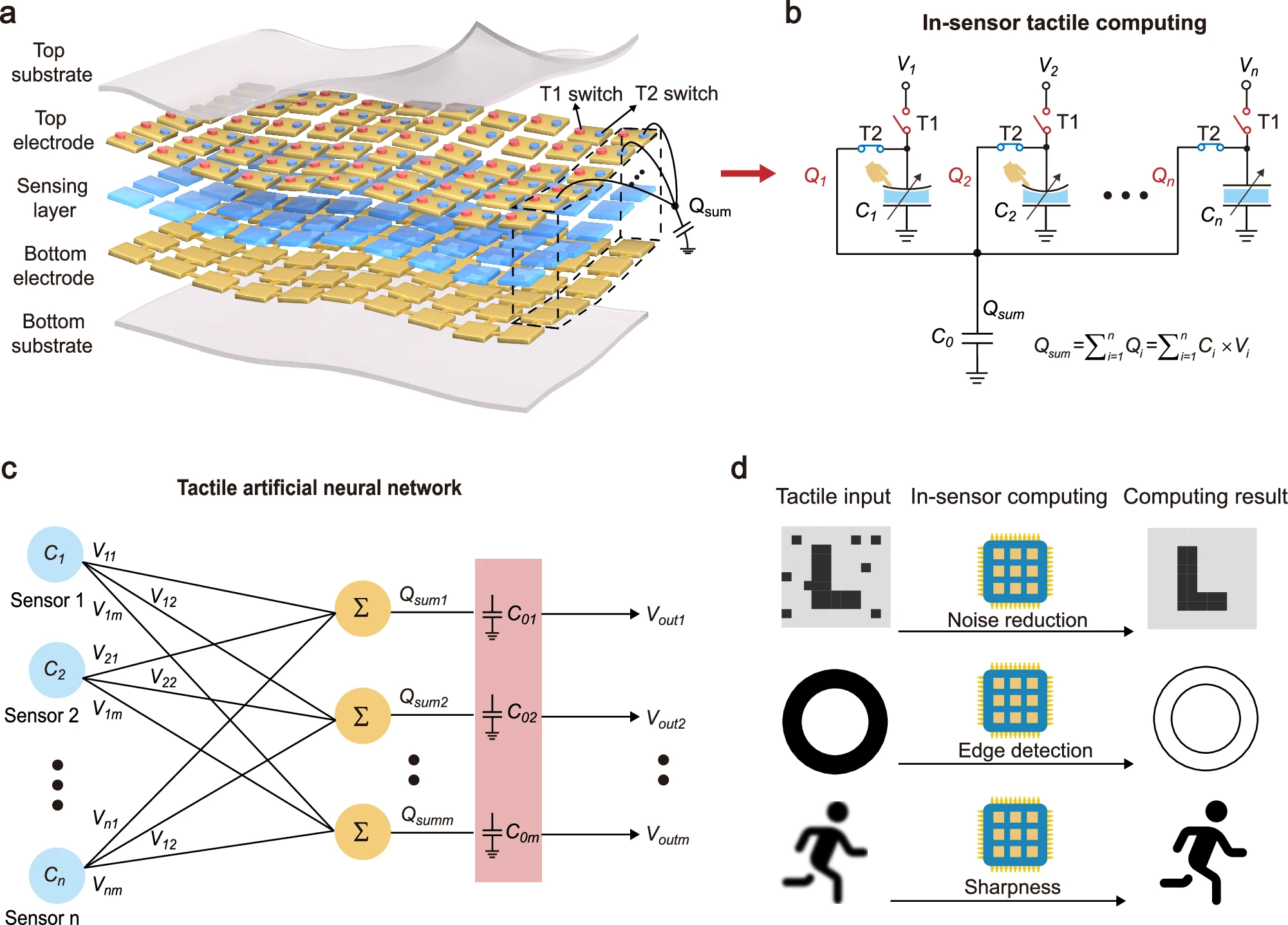}
    \caption{Capacitive in-sensor tactile computing architecture. (a) Schematic of the flexible capacitive sensor array integrated with electrical switches. (b) Circuit diagram demonstrating in-situ multiplication and accumulation (MAC) operations within a sensor subregion. (c) Equivalent tactile neural network mapping input sensor capacitance vectors $C$ to voltage weights $V$. (d) Examples of low-level sensory tasks, including noise reduction, edge detection, and sharpness extraction, computed directly within the sensor layer. The figure illustrates how early neural processing can be embedded directly into the sensing substrate, thereby reducing data transfer to external processing units. Reprinted from \cite{Chen2025}, licensed under CC BY 4.0.}
    \label{fig:chen_insitu}
\end{figure}

As already discussed above, a critical bottleneck in standard neuromorphic hardware is the transduction tax. 
Ferroelectric and capacitive architectures address this via ``in-sensor computing.'' 
Chen et al. demonstrated a system where the sensor elements themselves perform early-stage neural processing. 
By utilizing a flexible capacitive pressure sensor array, the system executes in situ analog multiplication and accumulation directly within the sensor network. 
This architecture bypasses the need for redundant data transfer to central processing units, effectively merging the ``skin'' and the ``brain'' of the device \cite{Chen2025}. The idea of this approach is shown in Figure \ref{fig:chen_insitu}.

For standard crossbars, the interface remains a challenge. 
Input vectors are converted via DACs, and output currents are digitized by ADCs. 
Optimizing the energy and area of these peripheral circuits is critical, as they can dominate the total power budget \cite{Sebastian2020}.

\subsection{Training and Design Paradigms}
Training requires navigating the ``non-idealities'' of the substrate, such as asymmetric update/decay curves and cycle-to-cycle variability. Accordingly, resistive and ferroelectric substrates support a comparatively broad spectrum of training strategies, ranging from ex-situ device-aware optimization to hybrid in-situ updates and reservoir-style use of intrinsic device dynamics.

In the ex-situ training paradigm, the network is trained on a GPU using a high-precision digital model. 
The resulting weights are mapped onto the physical conductances. 
``Device-aware'' training techniques explicitly model noise and drift during the training phase, making the network robust to the imperfections of the hardware deployment \cite{Joshi2020}. This approach is attractive because it leverages mature digital optimization while compensating for substrate imperfections, but it remains limited by write variability, conductance drift, and the reality gap between simulated and fabricated devices.

In-situ approaches update the weights directly on the chip. 
A powerful hybrid approach is ``Mixed-Precision Computing,'' where the crossbar performs the bulk of the low-precision forward/backward pass arithmetic, while a digital unit accumulates the high-precision gradient updates. 
This strategy combines the efficiency of analog physics with the numerical stability of digital algorithms \cite{LeGallo2018}. Such hybrid schemes are especially well matched to crossbar arrays because they exploit native analog parallelism while offloading numerically delicate update accumulation to digital control.

A particularly memristor-friendly learning regime is \emph{physical reservoir computing}, where the memristive substrate is not trained as a weight array but exploited for its intrinsic nonlinear, history-dependent dynamics. In a representative demonstration, \cite{Du2017MemristorRC} experimentally implemented reservoir computing using volatile (``dynamic'') WO$_x$ memristors whose internal ionic relaxation provides short-term memory, mapping temporal input streams into separable conductance-state trajectories. Only a simple readout is trained (e.g., linear regression), enabling temporal inference without iterative high-precision conductance programming. More recently, memristive reservoir-style models have also been explored at the algorithmic and modeling level for time-series processing \cite{Pistolesi2025}. This regime is particularly attractive for temporal tasks, although it generally sacrifices full end-to-end optimization of the internal device states.

Building on this memristive-reservoir perspective, \cite{horuz2026mars} shows how memristive-inspired temporal dynamics can be reorganized into a parallelized reservoir architecture for scalable time-series processing. Empirically, this substrate-aware reformulation achieves competitive, and in several cases superior, accuracy relative to modern gradient-based sequence models while reducing training from minutes or hours to the order of seconds, and in some cases to sub-second runtimes. Taken together, these results point to a broader co-design direction for physical neural computing: efficiency on memristive and in-memory substrates may depend not only on improved devices or calibration schemes, but also on model architectures whose computational structure is shaped from the outset by the dynamics and constraints of the target substrate.

%\subsection{Representative Demonstrations}
%Experimental milestones have validated both the learning and inference capabilities of these hardware substrates.
%Prezioso et al. demonstrated the first integrated ``memristor perceptron'' capable of classifying simple patterns using in-situ training \cite{Prezioso2015}. 
%Scaling up, Burr et al. utilized large-scale PCM arrays to perform high-accuracy inference on the MNIST and CIFAR datasets, introducing critical drift-compensation techniques \cite{Burr2014,Joshi2020}.
%In the domain of flexible electronics, Li et al.\ realized high-performance FeFETs suitable for transparent, wearable applications \cite{Li2024-aw}. 
%Most recently, Chen et al. implemented a fully integrated in-sensor tactile computing system capable of noise reduction and edge detection. Their approach demonstrated a power consumption over 22 times lower than conventional mixed-signal electronic systems, paving the way for energy-autonomous robotics \cite{Chen2025}.

\subsection{Representative Demonstrations}
Representative demonstrations of resistive and ferroelectric neural hardware span four regimes: in-situ learning in small arrays, large-scale analog inference, flexible synaptic devices for wearable electronics, and integrated in-sensor computing. Prezioso et al. demonstrated one of the first integrated memristor perceptrons, showing that simple pattern classification can be achieved directly in hardware with in-situ training rather than by offline weight transfer alone \cite{Prezioso2015}. Scaling up, Burr et al. used large PCM crossbar arrays to perform high-accuracy inference on MNIST and CIFAR while introducing drift-compensation strategies that addressed one of the key non-idealities of resistive memory hardware \cite{Burr2014,Joshi2020}. In the domain of flexible electronics, Li et al. realized high-performance FeFET devices based on ultra-thin indium tin oxide channels, demonstrating that synaptic functionality can be combined with transparency, flexibility, and suitability for wearable systems \cite{Li2024-aw}. At the system level, Chen et al. implemented an integrated in-sensor tactile computing architecture that performs low-level operations such as noise reduction and edge detection directly within the sensor layer, reducing data movement and achieving more than a 22-fold lower power consumption than conventional mixed-signal approaches \cite{Chen2025}.

\subsection{Engineering Constraints}
Despite their promise, these systems face significant physical limitations. Particularly in PCM, stored weight values relax over time due to conductance drift, degrading accuracy. In contrast, ferroelectric devices (FeFETs) typically exhibit superior retention characteristics and lack the significant resistance drift of PCM, though they suffer from ``imprint'' effects. Furthermore, endurance poses a challenge for ReRAM and PCM, which suffer from limited cycling endurance (often $10^8$--$10^9$ cycles). Ferroelectric oxides, however, demonstrate significantly higher endurance ($>10^{10}$ cycles) and lower switching energies, making them preferable for applications requiring frequent weight updates \cite{Sebastian2020}. Finally, voltage drops (IR drop) along the nanowires of large crossbars distort the effective voltage seen by devices, limiting the maximum feasible array size.

\subsection{Position within Physical Neural Computing}
Resistive and ferroelectric systems occupy the most technologically mature position in the landscape of physical neural computing. 
They are the primary candidates for solid-state ``In-Memory Computing,'' offering a direct path to break the von Neumann bottleneck. 
While PCM and ReRAM offer high density, the inclusion of ferroelectrics introduces superior linearity and energy efficiency, particularly for edge sensing applications. 
Their success, however, is inseparable from algorithm-hardware co-design: they are not ideal analog calculators, but noisy physical systems that require robust algorithmic wrappers to function reliably.

%--------------------------------------------------------------

\section{Spintronic and Superconducting Neural Systems}
\label{sec:spintronics_superconducting}

While resistive memory architectures (Section \ref{sec:memristive_pcm}) focus on density and in-memory multiplication, a distinct class of physical substrates targets the extremes of temporal resolution and energy efficiency. 
Spintronic and superconducting systems diverge from charge-based electronics by utilizing magnetic state variables as the carriers of information, namely the electron spin and the magnetic flux quantum.
These substrates are uniquely characterized by their ability to support ultrafast, oscillator-based, and spike-based computing paradigms that mimic the pulse-coded communication of biological neurons at speeds approaching the gigahertz and terahertz regimes \cite{Grollier2020}. 
We group them here as the ``Magnetic and Quantum-Electronic Class,'' unified by their reliance on collective quantum phenomena (ferromagnetism and superconductivity) to realize non-volatile dynamics with minimal dissipation \footnote{Here, ``quantum'' refers to quantum-mechanical device physics (e.g., flux quantization in superconducting circuits), not to qubit-based quantum neural computing that relies on superposition/entanglement, which is excluded in Section \ref{sec:scope}.}.

\subsection{The Substrate Principle}
This class exploits two fundamental physical regimes: room-temperature magnetic dynamics and cryogenic macroscopic quantum coherence.

Spintronics utilizes the intrinsic angular momentum (spin) of electrons. 
The core computational element is often the Spin-Torque Nano-Oscillator (STNO). 
When a DC current passes through a magnetic junction, it transfers angular momentum, inducing a steady precession of the magnetization. 
This converts a constant electrical input into a microwave frequency output, acting as a tunable non-linear oscillator \cite{Markovic2020}.
Moving beyond single devices, topologically stable, vortex-like spin textures such as magnetic Skyrmions exhibit strong non-linear dynamics that can be harnessed for neuromorphic and reservoir computing schemes \cite{Marrows2024,Pinna2020,Yokouchi2022}.

Superconducting computing operates at cryogenic temperatures (typically 4 K), where resistance vanishes, and quantized magnetic flux pulses (Single Flux Quanta, SFQ) represent the information carriers. Josephson Junctions (JJs) inherently generate such quantized SFQ voltage pulses and exhibit natural spiking dynamics analogous to biological neurons when biased appropriately. SFQ circuits have been proposed and demonstrated as neuromorphic hardware because they can operate at very high clock rates (tens to hundreds of GHz) with extremely low energy per pulse, orders of magnitude below typical CMOS spiking implementations \cite{Schneider2020_SFQSciRep,Russek2016_SFQ,Schneider2025_SelfTraining}.

\subsection{Mapping Neural Primitives}
The mapping of neural functions in these substrates shifts from static weights to dynamic synchronization and pulse interactions.

\subsubsection{Weights and Synchronization}
In spintronics, synaptic weights are often encoded in the coupling strength between oscillating STNOs. 
Computation emerges from the synchronization (phase-locking) of these oscillators, mimicking the temporal binding hypothesis of neuroscience \cite{Grollier2020}. 
In skyrmion reservoirs, the “weight” is implicit in the complex dynamical response of the magnetic textures; input signals perturb the skyrmion fabric, and the resulting high-dimensional non-linear modes encode the information \cite{Pinna2020,Raab2022_Brownian}.

\subsubsection{Activation and Spiking}
Superconducting Josephson Junctions provide the most physically faithful realization of a spiking neuron. 
A JJ integrates incoming flux pulses (current) until a critical current $I_c$ is exceeded. 
At this threshold, the junction switches, releasing exactly one flux quantum (a voltage spike) and resetting. 
This ``Integrate-and-Fire'' dynamics is intrinsic to the device physics and does not require complex circuitry to emulate \cite{Russek2016_SFQ}. 
Similarly, spintronic devices utilize the non-linear threshold of magnetic precession to implement activation functions.

\subsubsection{Plasticity and Self-Organization}
\begin{figure}
    \centering
    \includegraphics[width=1\columnwidth]{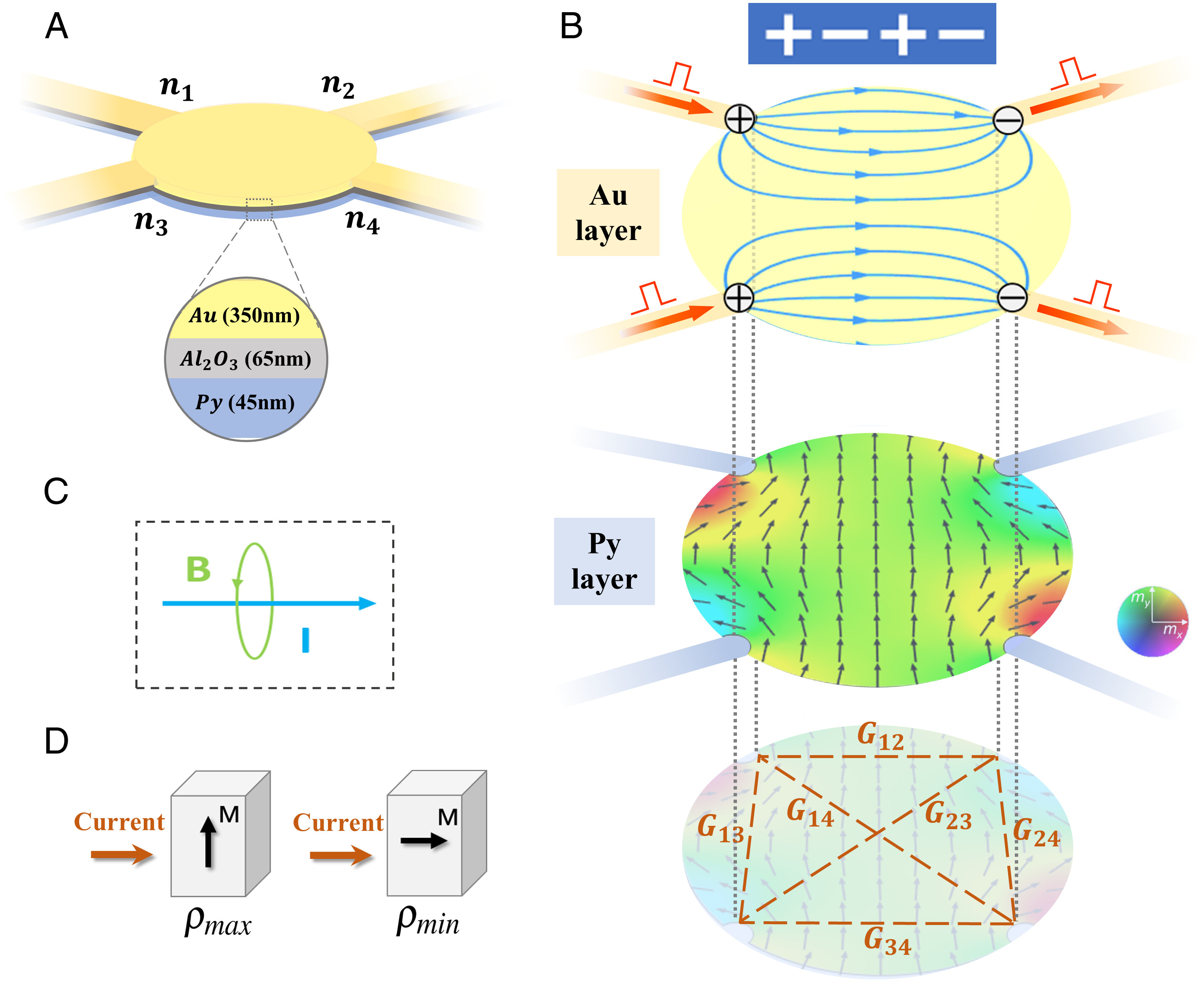}
%    \caption{Working principle of a 4-node self-learning magnetic Hopfield Neural Network (HNN). (A) Schematic of the experimental device structure. (B) Operational mechanism: The top Au layer acts as input neurons; voltage pulses generate current distributions that configure the spin orientation in the Py layer via the Oersted field (C). The resulting magnetic texture evolution defines the conductance matrix, mimicking synaptic weights. (D) The Anisotropic Magnetoresistance (AMR) effect is employed to detect state evolution based on the angle between current and magnetization. Reprinted from \cite{Niu2025_PNAS}, licensed under CC-BY-NC-ND 4.0.}
    \caption{Working principle of a 4-node self-learning magnetic Hopfield neural network (HNN). (A) Schematic of the experimental device structure. (B) Voltage pulses applied to the top Au layer encode the input or boundary conditions and generate current distributions that configure the spin orientation in the Py layer via the Oersted field (C). The resulting magnetic texture relaxation defines the effective conductance matrix, mimicking self-adjusting synaptic weights. (D) The anisotropic magnetoresistance (AMR) effect is used to read out the evolving network state through the angle-dependent relation between current and magnetization. The figure illustrates how associative memory and gradient-descent-like learning emerge directly from ferromagnetic energy minimization. Reprinted from \cite{Niu2025_PNAS}, licensed under CC-BY-NC-ND 4.0.} 
    \label{fig:magnetic_hnn}
\end{figure}

A major breakthrough in 2025 was the demonstration of intrinsic learning in magnetic materials. 
Niu et al. showed that magnetic textures can minimize their internal free energy in response to clamped boundary conditions. 
This physical relaxation process is mathematically isomorphic to the update rules of a Hopfield network, allowing the material to ``learn'' patterns via intrinsic gradient descent without external digital calculation \cite{Niu2025_PNAS}. The basic principle is shown with a sample 4-node Hopfield network in Figure \ref{fig:magnetic_hnn}.

\subsection{Architectures and I/O Interfaces}
Architectures in this domain favor reservoir computing and event-driven spiking networks over static feedforward layers.

Pinna et al. proposed that disordered (random) skyrmion textures can serve as a high-dimensional physical reservoir.
Inputs (e.g., current- or field-induced perturbations) locally distort the magnetic texture and excite non-linear collective dynamics.
The resulting interactions between skyrmions and the surrounding spin background (e.g., repulsion, deformation, mode coupling) embed the input history into a rich spatiotemporal state that can be linearly read out, for instance via electrical transduction such as magnetoresistive signals \cite{Pinna2020}.
This approach does not rely on explicitly training the substrate itself; instead, it exploits the intrinsic complexity of the skyrmion texture and trains only the readout layer.

Superconducting architectures arrange Josephson junctions into large-scale spiking neural networks (SNNs) operating with Single Flux Quantum (SFQ) pulses.
Signals propagate as ballistic quantized flux pulses along superconducting transmission lines with negligible resistive loss and minimal dispersion.
Such networks are particularly suited for processing ultrafast temporal data streams (e.g., from RF or cryogenic sensors) in real time, enabling operations such as coincidence detection and temporal pattern matching at clock rates far beyond conventional CMOS implementations \cite{Schneider2025_SelfTraining}.

\subsection{Training and Design Paradigms}
Training methods differ significantly between the two pillars of this chapter. At a high level, magnetic systems are increasingly explored in the in-materio regime, where optimization is partly delegated to the intrinsic energy landscape of the substrate, whereas superconducting neural systems more often rely on local in-situ update rules or hybrid training schemes.

For magnetic systems, the ``In-Materio'' learning paradigm is gaining traction. 
By mapping the loss function of a neural network to the Hamiltonian (energy function) of the ferromagnet, the training process becomes a physical relaxation to the ground state. 
This physics-driven strategy is attractive because it exploits the natural relaxation dynamics of the substrate and can partially replace a fully external optimizer loop. However, its practical effectiveness depends on how faithfully the physical energy landscape represents the computational objective and on whether the relaxation reliably reaches useful minima \cite{Niu2025_PNAS}.

In superconducting SNNs, training often utilizes Spike-Timing-Dependent Plasticity (STDP). 
This can be implemented using variable inductive loops (SQIFs) or by hybridizing with magnetic Josephson junctions, where the magnetic state (and thus the critical current) is modified by the timing of arriving flux quanta \cite{Grollier2020}. Such local update rules are well matched to event-driven superconducting hardware because they directly exploit spike timing and device-level dynamics, but they currently offer less flexibility than large-scale gradient-based digital optimization and remain constrained by hardware complexity, tunability, and reproducibility.

%\subsection{Representative Demonstrations}
%Several experimental studies illustrate how magnetic and superconducting dynamics can be harnessed for neuromorphic information processing.
%Pinna et al. investigated disordered (random) skyrmion textures as physical reservoirs and showed that their intrinsic non-linear dynamics generate high-dimensional spatiotemporal responses suitable for temporal signal processing tasks, supporting the use of magnetic textures as substrates for reservoir computing \cite{Pinna2020}.

%In the field of superconducting, Schneider et al. implemented a spiking neural network based on Josephson junctions and Single Flux Quantum (SFQ) signaling, demonstrating operation at multi–tens of gigahertz clock rates with attojoule-scale switching energies \cite{Schneider2025_SelfTraining}. The work provides a concrete example of how superconducting device physics can be exploited for high-speed, energy-efficient neuromorphic architectures.

%Finally, Niu et al. realized a magnetic Hopfield network in which synaptic adaptation is governed by the intrinsic energy landscape of a ferromagnetic system, thereby demonstrating associative memory and gradient-descent-like learning directly at the device level \cite{Niu2025_PNAS}.

\subsection{Representative Demonstrations}
Representative demonstrations in this class span three characteristic regimes: magnetic reservoir computing, ultrafast superconducting spiking networks, and intrinsic self-learning in magnetic energy landscapes. Pinna et al. investigated disordered (random) skyrmion textures as physical reservoirs and showed that their intrinsic non-linear dynamics generate high-dimensional spatiotemporal responses suitable for temporal signal processing tasks, supporting the use of magnetic textures as substrates for reservoir computing \cite{Pinna2020}. In the superconducting domain, Schneider et al. implemented a spiking neural network based on Josephson junctions and Single Flux Quantum (SFQ) signaling, demonstrating operation at multi--tens of gigahertz clock rates with attojoule-scale switching energies \cite{Schneider2025_SelfTraining}. This work provides a concrete example of how superconducting device physics can be exploited for high-speed, energy-efficient neuromorphic architectures. Finally, Niu et al. realized a magnetic Hopfield network in which synaptic adaptation is governed by the intrinsic energy landscape of a ferromagnetic system, thereby demonstrating associative memory and gradient-descent-like learning directly at the device level \cite{Niu2025_PNAS}. Figure \ref{fig:magnetic_hnn} illustrates this principle and shows how boundary-conditioned magnetic texture relaxation defines the effective conductance matrix and can be read out electrically via anisotropic magnetoresistance.

\subsection{Engineering Constraints}
The impressive speed and efficiency of these substrates come with distinct physical barriers. 
In spintronics, thermal fluctuations can destabilize small magnetic textures (superparamagnetism), imposing a trade-off between device miniaturization and information retention, although stochastic resonance can also be exploited for probabilistic computing \cite{Markovic2020}. 

In the superconducting domain, the ``cryogenic penalty'' remains a major hurdle, as circuits require cooling to temperatures around 4\,K. 
While the switching energy of SFQ pulses is in the attojoule range and spiking networks can operate at clock rates of tens to hundreds of gigahertz \cite{Schneider2025_SelfTraining}, the required cryogenic infrastructure (cryocoolers, thermal shielding, bias distribution, and room-temperature I/O electronics) dominates the overall energy budget. 
System-level analyses show that the cooling and support electronics typically consume from tens of watts up to kilowatts of electrical power, depending on system scale and cooling technology, overwhelming the energy dissipated by the Josephson junctions themselves \cite{Holmes2013_PowerBudgets}. 

As a result, superconducting neuromorphic hardware is currently viable primarily for large-scale high-performance computing installations or specialized environments such as quantum-computer control systems, rather than for mobile or embedded devices. 
In addition, impedance and voltage-level mismatches between superconducting SFQ circuits (millivolt-level signals) and room-temperature CMOS logic (volt-level signals) make interfacing complex and energetically costly, favoring architectures in which superconducting processors operate as cryogenic ``islands'' connected to conventional electronics via dedicated interface circuits \cite{Holmes2013_PowerBudgets}.

\subsection{Position within Physical Neural Computing}
Spintronic and superconducting systems occupy the ``high-performance'' corner of the physical computing landscape. 
They offer a stark contrast to the slow, diffusive nature of wetware and the static nature of resistive crossbars. 
Spintronics offers a path to non-volatile, radiation-hard, and oscillatory computing at room temperature. 
Superconducting logic, conversely, represents the ultimate physical limit of speed and efficiency, theoretically capable of matching the connectivity of the human brain with the speed of a supercomputer, provided the cryogenic barrier is accepted. 
Together, they constitute the magnetic and quantum-electronic frontier of PNNs.

%------------------------------------------------------------

\section{Photonic Neural Computing}
\label{sec:photonic}

Photonic computing replaces electrons with photons as the information carrier, leveraging the intrinsic properties of light—ultra-high bandwidth, massive parallelism via wavelength division multiplexing (WDM), and propagation at the speed of light—to perform computation. While electronic processors are throttled by capacitive charging and interconnect latency, photonic processors operate in the ``flight-time'' regime, where the latency is determined solely by the refractive index. This makes photonics the premier candidate for accelerating linear operations at speeds and energy efficiencies unattainable by CMOS \cite{Shastri2021,Wetzstein2020}.

\subsection{The Substrate Principle}
The fundamental physical mechanism exploited is wave interference. Optical propagation naturally implements linear transformations; when coherent modes overlap, their amplitudes sum constructively or destructively based on phase. This allows matrix-vector multiplication (which arguably is the computational bottleneck of neural networks) to be performed passively as light traverses a medium.
Information is encoded in amplitude, phase, polarization, or spatial mode. Unlike the charge-accumulation logic of electronics, the computation here is a transmission problem: the result is the scattering of the input field by the optical elements \cite{Shen2017,Miller2013}. However, since photons lack charge and interact only weakly (making ``optical transistors'' difficult), practical systems often adopt hybrid architectures: linear operations occur optically, while non-linear activations and control logic remain electronic \cite{Tait2017,Prucnal2016}.

\subsection{Mapping Neural Primitives}
Mapping neural networks to photonics requires translating linear algebra into optical components while addressing the inherent difficulty of implementing non-linearity.

\subsubsection{Weights and Summation}
In coherent architectures, a synaptic weight is physically represented by the transmission coefficient or phase shift of a component. Vector-Matrix Multiplication (VMM) is performed as light propagates through a programmable mesh of Mach-Zehnder Interferometers (MZIs) or diffractive surfaces. The summation is inherent to the physics of superposition at the detector or combiner ports \cite{Clements2016,Miller2013}.

\subsubsection{Activation and Nonlinearity}
Implementing non-linearity remains the primary engineering hurdle. While some approaches explore all-optical non-linearities using saturable absorbers or bistable resonators, these often require prohibitively high power densities. The dominant paradigm relies on Opto-Electronic (O-E-O) conversion: the optical signal is detected, thresholded electronically, and re-modulated onto a carrier. This introduces a bandwidth bottleneck but ensures stable operation \cite{Williamson2019,Tait2017}.

\subsection{Architectures and I/O Interfaces}
Photonic architectures generally fall into three categories, trading off programmability against throughput.

Programmable photonic integrated circuits (PICs) utilize meshes of MZIs to implement arbitrary unitary matrices, following the triangular decomposition scheme originally proposed by Reck et al.\ \cite{Reck1994}. In these ``Optical TPUs,'' thermo-optic or electro-optic phase shifters act as tunable weights. A landmark theoretical framework by Miller proved that such meshes can self-configure to perform any linear transformation, a concept later validated experimentally for high-speed inference \cite{Miller2013,Shen2017}.

\begin{figure}
    \centering
    \includegraphics[width=1\linewidth]{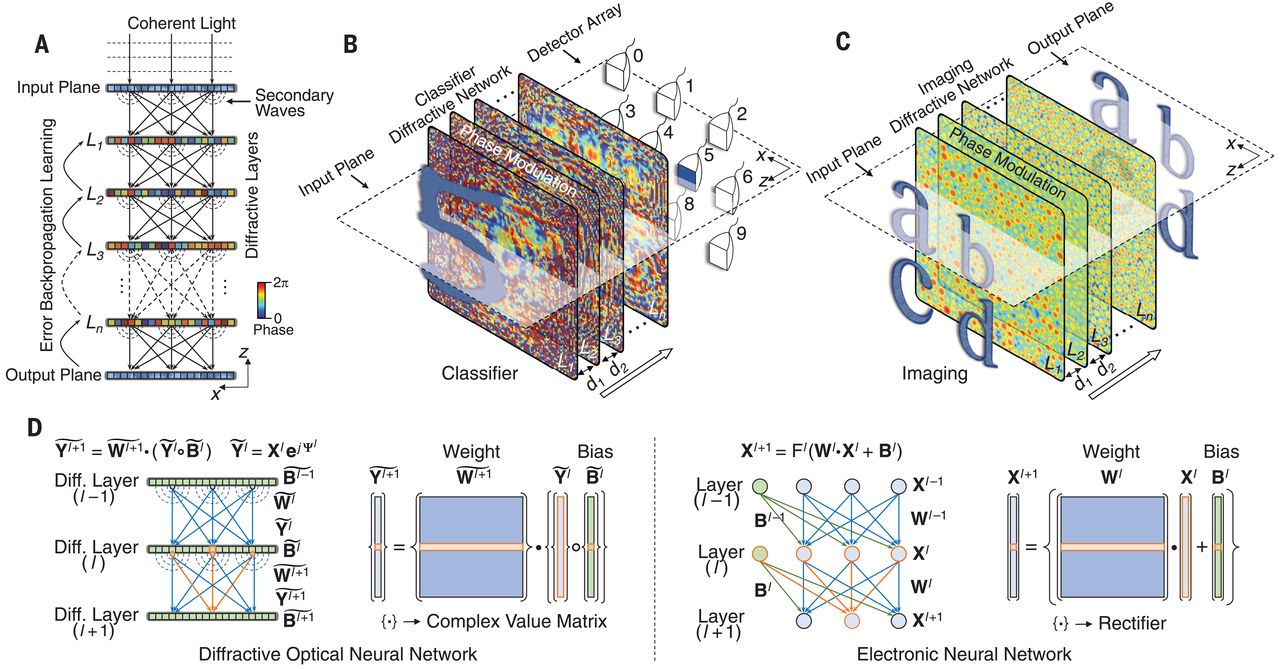}
%    \caption{Concept of All-Optical Diffractive Deep Neural Networks ($D^2NN$). 
%    (A) General architecture of a standard fully-connected digital neural network. 
%    (B) Physical implementation as a $D^2NN$. The network consists of multiple passive diffractive layers (e.g., 3D-printed polymer) that modulate the phase and amplitude of the incident light field. Each point on a layer acts as a neuron, connecting to the next layer via optical diffraction. The system performs inference (e.g., digit classification) at the speed of light by focusing the wave onto a specific detector region. 
    \caption{Diffractive deep neural networks (D$^2$NNs).
    (A) A D$^2$NN uses multiple trainable diffractive layers with complex-valued transmission or reflection coefficients to implement an optical function between input and output planes. Once the layer parameters have been optimized and the structure has been fabricated, inference is performed passively through wave propagation at the speed of light. (B) Example of a D$^2$NN for handwritten digit classification. (C) D$^2$NN functioning as an imaging lens. (D) Comparison with a conventional neural network, highlighting coherent complex-valued inputs and diffraction-based layer interconnections. The figure illustrates the key trade-off of this photonic regime: extremely fast and energy-efficient inference at the cost of limited post-fabrication programmability. Reprinted from \cite{Lin2018} with permission from the American Association for the Advancement of Science.} %with permission from AAAS.}} 
    %% License cost roughly 26 Euros
    \label{fig:diffractive_DNN}
\end{figure}

Diffractive architectures operate in free space. A series of passive transmissive layers (engineered metasurfaces) modulate the phase and amplitude of an incoming wavefront. As the wave propagates through these layers, the diffraction pattern physically computes the inference. These systems operate at the speed of light with minimal energy consumption, but act as fixed, read-only classifiers once fabricated \cite{Lin2018}. The aprooach is visualized in Figure \ref{fig:diffractive_DNN}.

For temporal processing, photonic reservoirs exploit the rich transient dynamics of optical feedback. Architectures using semiconductor lasers with delayed feedback or fiber-optic loops generate high-dimensional state spaces from time-series inputs. These systems excel at chaotic signal prediction and speech recognition, requiring training only at the electronic readout layer \cite{Appeltant2011,Paquot2012}.

\subsection{Training and Design Paradigms}
Training photonic circuits presents challenges due to phase noise and the ``analog gap'' between simulation and reality. Accordingly, photonic neural hardware is currently dominated by ex-situ digital training with calibrated deployment, while more recent work increasingly explores in situ gradient measurement directly on the physical chip.

The standard approach involves training a model digitally and mapping the weights to physical phase shifts. However, fabrication tolerances and thermal crosstalk require rigorous calibration routines to align the physical device with the digital model \cite{Shen2017}. Algorithms specifically designed to robustly configure MZI meshes in the presence of imperfections are essential for scaling \cite{Miller2013}. This paradigm is attractive because it leverages mature digital optimization while preserving the high-speed, low-latency inference capabilities of photonic hardware. On the other hand, it remains limited by calibration overhead and residual mismatch between simulated and fabricated devices.

More recent advances utilize \textit{in situ} training, where gradients are measured directly on the hardware. Techniques such as the ``adjoint variable method'' physically backpropagate light through the output ports to measure sensitivity, automatically accounting for the exact physical state of the chip, including defects \cite{Hughes2018}. Such approaches are particularly promising for reducing the analog gap and compensating for fabrication defects, although they still require stable optical control and precise measurement infrastructure to remain practical at scale.

%\subsection{Representative Demonstrations}
%Significant milestones have moved photonics from theory to prototype.
%Shen et al. demonstrated a programmable nanophotonic processor using an MZI mesh capable of vowel recognition, validating the coherent matrix multiplier concept \cite{Shen2017}.
%In the diffractive domain, Lin et al. utilized 3D-printed layers to classify MNIST digits purely through wave propagation, achieving passive ``zero-power'' inference \cite{Lin2018}.
%In the temporal domain, Appeltant et al. and Paquot et al. successfully employed delayed-feedback laser reservoirs for high-speed speech recognition and chaotic time-series prediction, proving the efficacy of photonic dynamics for recurrent tasks \cite{Appeltant2011,Paquot2012}.

\subsection{Representative Demonstrations}
Representative demonstrations of photonic neural computing span three characteristic regimes: programmable coherent matrix processors, passive diffractive inference, and dynamical reservoir computing. Shen et al. demonstrated a programmable nanophotonic processor based on a mesh of Mach--Zehnder interferometers that performed vowel recognition, providing one of the first convincing experimental validations that coherent optical interference can implement trainable matrix operations for neural inference \cite{Shen2017}. In the diffractive domain, Lin et al. realized a deep optical classifier using multiple 3D-printed diffractive layers that classified handwritten digits purely through wave propagation after fabrication, showing that a trained optical structure can perform passive inference at the speed of light with essentially no active computation during operation \cite{Lin2018}. Figure \ref{fig:diffractive_DNN} illustrates this principle and highlights both the layer-wise diffraction-based transformation and the fixed, read-only nature of the fabricated system. For temporal tasks, Appeltant et al. and Paquot et al. demonstrated delayed-feedback photonic reservoirs for speech recognition and chaotic time-series prediction, showing that optical dynamics can also support recurrent computation with fading memory rather than only static linear transforms \cite{Appeltant2011,Paquot2012}.

\subsection{Engineering Constraints}
The adoption of photonic computing is constrained by scale and sensitivity.
First, optical components are orders of magnitude larger than transistors (constrained by the wavelength of light), limiting the number of neurons per chip and thus the integration density.
Second, analog optical computing is limited by laser noise, detector shot noise, and thermal drift, typically capping effective precision at 4-8 bits \cite{Tait2017}.
Third, the transduction tax has to be taken into account again: the energy and latency cost of converting data between the electrical and optical domains can negate the benefits of optical processing unless the computational depth is sufficiently large \cite{Shastri2021}.

\subsection{Position within Physical Neural Computing}
Photonic neural computing occupies the ``high-speed, low-latency'' corner of the physical computing landscape. Unlike chemical or mechanical systems that excel at slow, adaptive dynamics, photonics excels at performing massive linear algebra operations at the bandwidth limit of the electromagnetic spectrum. Ideally suited for ultrafast signal processing and server-side acceleration, it complements substrates like memristors or wetware that offer higher density or plasticity but lower speed \cite{Prucnal2016,Tait2017}.

%----------------------------------------------------------------

\section{Mechanical Metamaterials}
\label{sec:mechanical}

Mechanical neural computing revisits mechanics, which is arguably the oldest substrate of calculation, through the modern lens of metamaterials. Unlike the rigid gears of the Babbage engine, contemporary mechanical computing relies on compliant mechanisms and the non-linear deformation of continuous media. By engineering the topology of lattice structures, information is processed not by the flow of electrons, but by the propagation of strain energy and the interactions of mechanical instabilities \cite{Lee2022}. This substrate is particularly relevant for ``Embodied Intelligence'' and morphological computation, where the computation is physically integrated into the structural load-bearing elements of a robotic system \cite{Pfeifer2007,Kim2013, ZahediAy2013, ZahediLangerAy2017}.

\subsection{The Substrate Principle}
The fundamental computational resource in mechanical systems is the elastic potential energy landscape. While linear elastic materials obey Hooke's Law (force proportional to displacement), neural computation requires non-linearity. This is achieved through geometric instabilities, most notably ``snap-through buckling.''
A bistable buckled beam possesses two energy minima separated by an energy barrier. Transitioning between these states allows for binary storage (memory) and thresholding behavior \cite{Bertoldi2017}. In metamaterials, thousands of such unit cells are coupled in a lattice; the global deformation of the material is thus the aggregate result of local interactions, functioning effectively as a massive network of mechanically coupled non-linear springs \cite{Coulais2017}.

\subsection{Mapping Neural Primitives}
To act as a neural network, the mechanical substrate must exhibit weighted connections, non-linear activation functions, and mechanisms for plasticity.

\subsubsection{Stiffness as Weight}
In a mechanical network, the ``signal'' is typically a force or displacement vector. The synaptic weight corresponds to the local stiffness ($k$) of the beams connecting nodes. A stiffer beam transmits force more effectively, effectively amplifying the signal, while a compliant beam attenuates it. In programmable metamaterials, these weights can be adjusted by altering the beam thickness during fabrication or, in active systems, by modulating the stiffness using piezoelectric or magnetic actuators \cite{Coulais2017,Lee2022}.

\subsubsection{Instability as Activation}
The non-linear activation function is realized through buckling phenomena and topological modes. A slender beam under compression initially acts linearly. However, once a critical load (Euler load) is exceeded, it snaps to a new configuration. This sudden change in stiffness acts as a physical sigmoid or Heaviside function, allowing the system to perform decision-making and logic operations purely through geometric deformation \cite{Kane2014,Ion2017}.

\subsubsection{Plasticity and Memory}
While many architectures are static (read-only), adaptive materials enable in-materio learning. 
Memory can be encoded in the reconfigurable geometry of multistable elements and via time-dependent relaxation or aging processes. 
Recent work demonstrates supervised learning through physical parameter updates in creased sheets, and reveals how viscoelastic relaxation and hysteretic rearrangements in disordered (amorphous) media generate history-dependent responses and memory 
\cite{Stern2020, Dykstra2019, Keim2020, Pashine2019}.

\subsection{Architectures and I/O Interfaces}
Architectures in mechanical computing are defined by their unit cell topology and dimensionality.

%\subsubsection{Lattice Metamaterials}
2D and 3D lattices of interconnected beams form the bulk of mechanical neural networks. These structures function as ``physical perceptrons,'' where an input force pattern applied to one boundary propagates through the lattice to produce a target displacement at an output boundary \cite{Coulais2017}.

%\subsubsection{Origami and Kirigami}
Alternatively, plate-based systems utilize origami (folding) and kirigami (cutting) principles. Here, the ``neurons'' are the intersection vertices of folds. The non-linear rotational stiffness of the creases allows these sheets to function as reconfigurable logic gates and multi-input classifiers with high packing density \cite{Waitukaitis2015,Zhai2021OrigamiKirigamiReview,Stern2020}.

%\subsubsection{Continuum Reservoirs}
Third, soft bodies and elastic solids with complex internal dynamics can be used as physical reservoirs. Similar to their fluidic counterparts, soft silicone arms or gels exploit their infinite degrees of freedom to project low-dimensional inputs (e.g., motor commands) into a high-dimensional dynamic state space, enabling temporal pattern recognition for soft robotic control \cite{Nakajima2014}.

\subsection{Training and Design Paradigms}
Training a mechanical material implies finding a configuration of stiffnesses or geometries that yields a desired force-displacement response. Accordingly, mechanical substrates are dominated by ex-situ inverse design, using either gradient-based differentiable mechanics or closed box evolutionary optimization, whereas direct in-materio adaptation remains comparatively rare.

The classical engineering approach formulates the desired input-output mapping as a constraint optimization problem. Algorithms such as SIMP (Solid Isotropic Material with Penalization) optimize the density distribution of material within a design domain to satisfy these functional constraints \cite{Sigmund2013}. This paradigm is well suited to mechanical systems because geometry and constitutive behavior can often be modeled explicitly, but it is limited by modeling assumptions and by the difficulty of capturing fabrication tolerances and material nonlinearities with high fidelity.

More recent \emph{differentiable mechanics} approaches embed differentiable simulators or differentiable surrogate models into an
optimization loop, enabling gradient-based inverse design of geometries and effective stiffness parameters \cite{Du2022DiffPD,Dold2023GraphLattices,Lee2022}. Alternatively, evolutionary algorithms can mutate geometry representations under task-level fitness functions, which is attractive when
accurate gradients are unavailable, or the design space is highly discrete
\cite{Miskin2013ArtificialEvolution,Dong2022VoxelEA}. Gradient-based methods are typically more sample-efficient when accurate simulators are available, whereas evolutionary strategies are more robust to discontinuous design spaces and non-differentiable objectives, albeit usually at higher search cost.

\subsection{Representative Demonstrations}
\begin{figure}
    \centering
    \includegraphics[width=\columnwidth]{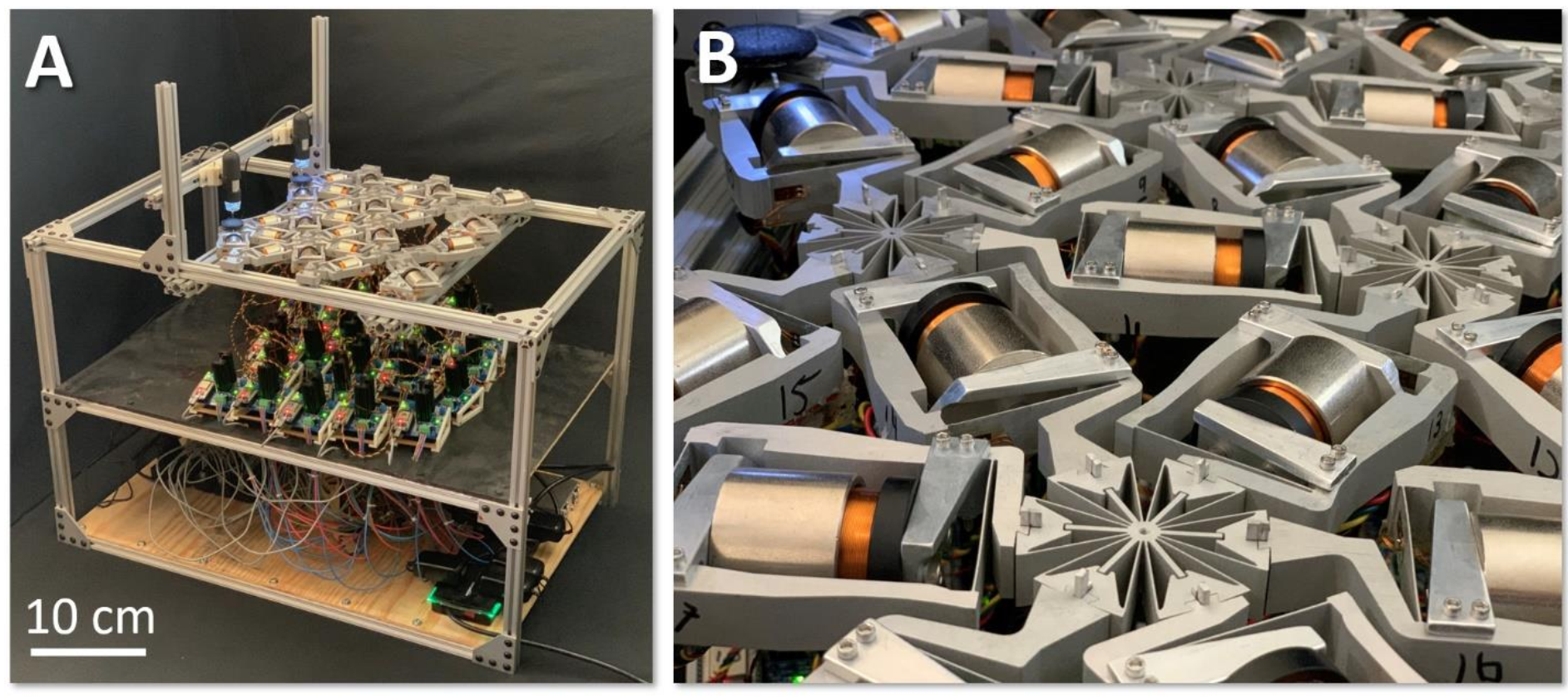}
    \caption{Mechanical neural network (MNN) fabricated for the experimental study of Lee et al. \cite{Lee2022}. (A) Full-view photograph of the lattice-based mechanical network. (B) Close-up of the beam structure whose local stiffness values act as trainable mechanical weights. The figure illustrates the physical realization of a mechanical perceptron-like architecture in which desired behaviors, such as shape morphing, are achieved by optimizing the geometry and stiffness distribution of the lattice. Reprinted from \cite{Lee2022} with permission from the American Association for the Advancement of Science.}
    %\caption{Mechanical neural network (MNN) fabricated for the experimental study of \cite{Lee2022}. (A) Full-view and (B) close-up photographs. Reprinted from \cite{Lee2022}.} %% buy license!!
    \label{fig:lee_mnn}
\end{figure}

Representative demonstrations of mechanical neural computation span three characteristic regimes: trainable lattice-based networks, in-situ gradient-based learning, and geometry-driven logic and classification.
A major milestone was the experimental realization of mechanical neural networks (MNNs) by Lee et al. They fabricated a lattice of tunable beams whose stiffnesses are adjusted analogously to ANN weights, enabling the structure to learn multiple mechanical behaviors (e.g., shape morphing) through repeated training/optimization and experimental validation \cite{Lee2022}. Figure \ref{fig:lee_mnn} shows the fabricated network and illustrates the physical realization of this trainable mechanical architecture.

Building on this, Li and Mao demonstrated that such networks can also be trained for regression and classification using in situ backpropagation, extracting gradient information directly from the mechanical system's physical response without relying on a digital twin \cite{Li2024AllMechanical}.

A complementary line of work exploits geometry and multistability more directly. Coulais et al. realized combinatorial logic gates using bistable mechanical units arranged in planar lattices \cite{Coulais2017}. More broadly, origami- and kirigami-based metamaterials demonstrate that nonlinear classifier-like input-output mappings can emerge directly from geometry, folding mechanics, and physical adaptation rather than from electronically controlled processing \cite{Stern2020,Zhai2021OrigamiKirigamiReview}.

\subsection{Engineering Constraints}
Mechanical computing faces limits imposed by material science and wave physics.
First, signals travel at the speed of sound within the material (typically km/s), which is orders of magnitude slower than electromagnetic signals. This limits mechanical systems to low-frequency control tasks (Hz to kHz range).
Second, repeated cycling of bistable elements can lead to plastic deformation or fatigue failure, altering the computational properties (weights) over time.
Third, in viscoelastic materials, energy is dissipated as heat (damping). This limits the depth of the network, as the signal-to-noise ratio degrades with distance from the input source unless active amplification is introduced.

\subsection{Position within Physical Neural Computing}
Mechanical metamaterials occupy the niche of ``structural intelligence.'' They are unlikely to compete with electronic or photonic substrates in speed or accuracy, but they uniquely enable computation without electricity and intrinsic coupling to physical environments. By embedding intelligence directly into load-bearing structures, they represent a paradigm complementary to molecular or electronic approaches, ideal for applications requiring extreme robustness and morphological adaptation \cite{Pfeifer2007}.

% ---------------------------------------------------------

\section{Microfluidic Neural Networks (Pressure-Driven)}
\label{sec:microfluidics}

Microfluidic systems implement computation by routing liquids or gases through top-down engineered networks of micrometer-scale channels. Unlike chemical or DNA-based substrates, where the computational structure emerges bottom-up from molecular interactions, microfluidic neural systems constitute rigid devices whose topology is defined by lithographically fabricated geometries and boundary conditions \cite{Whitesides2006}. Consequently, they function as deterministic hydraulic integrated circuits that process information through the controlled transport of mass rather than charge, enabling the direct manipulation of physical matter \cite{Prakash2007}. As detailed in a recent review by Law et al., this field is rapidly expanding towards fully fluidic neuromorphic architectures that integrate flow-based logic with fluidic memristive elements \cite{Law2025} .

\subsection{The Substrate Principle}
Microfluidic neural computing operates strictly in the low Reynolds number regime ($\mathrm{Re} \ll 1$), where viscous forces dominate over inertia. In this domain, fluid motion is governed by the Stokes equations, resulting in flow that is laminar, deterministic, and time-reversible in the absence of active switching elements \cite{Squires2005}. This physical regime permits a rigorous analogy between hydraulic and electronic transport, enabling the systematic design of fluidic logic using established circuit theory \cite{Oh2012PressureDriven}.

In this analogy, long, narrow channels act as hydraulic resistors, inducing pressure drops linearly proportional to the volumetric flow rate. Elastic chambers function as fluidic capacitors, storing volume under pressure and introducing temporal integration constants. Asymmetric channel structures, such as Tesla valves, rectify flow similar to diodes, while pressure-actuated flexible membranes implement active gating, functionally replicating the behavior of transistors \cite{Unger2000}. By composing these fundamental elements, microfluidic substrates can realize complex analog signal processing architectures completely within the fluidic domain.

\subsection{Mapping Neural Primitives}
To construct neural networks, the governing fluid dynamic equations must be mapped to the mathematical primitives of weighted summation, non-linear activation, and state retention.

\subsubsection{Synaptic Weights and Summation}
Synaptic weighting is physically encoded in the hydraulic resistance of the channel network. According to the Hagen-Poiseuille law, the resistance scales linearly with channel length and inversely with the fourth power of the hydraulic diameter. Therefore, weights can be statically defined during fabrication by optimizing channel geometries to restrict signal flow \cite{Bruus2008}. The summation operation emerges naturally from the conservation of mass at channel junctions, where merging fluid streams result in a combined pressure or flow rate proportional to the weighted inputs.

\subsubsection{Activation and Nonlinearity}
While low-Reynolds-number flow is inherently linear, neural computation requires non-linearity. This is introduced through fluid-structure interactions or multiphase dynamics. Pressure-gain valves and flexible membranes exhibit thresholding behaviors where flow is blocked until a critical pressure is exceeded, effectively implementing rectified linear unit (ReLU) or sigmoidal activation functions \cite{Unger2000}. Alternatively, ``Bubble Logic'' utilizes the non-linear interaction of immiscible fluids, where the presence or absence of a droplet in a channel acts as a discrete state variable capable of performing Boolean logic operations \cite{Prakash2007}.

\subsubsection{Memory Mechanisms}
Temporal dynamics and memory are realized through fluidic capacitance and inertia. Elastic deformation of channel walls allows for short-term storage of signal state (integration). For long-term memory, systems utilize bistable mechanisms such as trapped droplets, hysteretic valve switching, or high-viscosity non-Newtonian fluids that retain their state in the absence of continuous power, functioning as fluidic flip-flops \cite{Weaver2010,Preston2019}.

\subsection{Architectures and I/O Interfaces}
Microfluidic architectures typically follow one of two paradigms, dictated by the trade-off between control complexity and dynamic richness.

Layered channel networks implement fixed-weight mappings where the computation is defined by the circuit topology. These systems function as analog feedforward networks for classification or regression tasks, or as digital logic circuits for Boolean operations \cite{Dertinger2001}. While interpretable, they often require complex multilayer fabrication to handle signal routing (crossovers).

To bypass the need for precise component-level design, recent approaches utilize reservoir computing. Here, complex channel geometries, recirculation zones, or multiphase flows generate rich, high-dimensional transient dynamics. The fluidic body acts as a physical kernel, projecting inputs into a state space that is linearised by a simple readout layer. This approach is particularly robust against fabrication tolerances and has been successfully applied to temporal pattern recognition and soft robotic control \cite{Nakajima2014,Nakajima2015}.

The interface between the macro- and microworld represents a significant engineering bottleneck. Inputs are encoded as continuous pressure levels, volumetric flow rates, or discrete droplet sequences, typically requiring external solenoid valves or syringe pumps. Readout is performed via integrated pressure transducers, flow sensors, or optical imaging of dye concentrations, often imposing the transduction tax discussed above that limits the overall system bandwidth \cite{Sackmann2014}.

\subsection{Training and Design Paradigms}
Unlike electronic neural networks, microfluidic systems generally do not provide native mechanisms for rapid, repeated in-situ parameter updates during operation. Consequently, training strategies have historically relied on in-silico optimization. Accordingly, microfluidic substrates are dominated by ex-situ digital-twin design, complemented by gradient-based inverse design and, where necessary, closed box evolutionary optimization. Direct in-materio adaptation remains limited because channel geometries are typically fixed once fabricated and fluidic dynamics evolve on relatively slow timescales.

The dominant paradigm is offline training, where a differentiable model of the fluid dynamics (a digital twin) is trained digitally. The resulting optimal geometries (channel widths, lengths) are then transferred to the physical substrate via lithography. This approach effectively treats the microfluidic chip as an inference-only ASIC. This strategy is attractive because fluid transport is often sufficiently structured to support simulation-driven optimization, but it remains limited by fabrication tolerances, model mismatch, and the absence of rapid post-fabrication reconfiguration.

Recent advances have integrated finite-element fluid solvers directly into automatic differentiation pipelines, enabling gradient-based optimization of channel geometries for complex flow control tasks \cite{Holl2020}. Complementary approaches use evolutionary algorithms or machine learning to search the topological design space for networks that exhibit desired behaviors, such as oscillation or gait generation, without requiring an explicit gradient \cite{D2LC00254J}. Gradient-based methods are typically more sample-efficient when accurate fluid models are available, whereas evolutionary search is better suited to highly discrete or topological design spaces, albeit usually at substantially higher search cost.

\subsection{Representative Demonstrations}
Representative demonstrations of pressure-driven microfluidic neural computation span three characteristic regimes: digital fluidic logic, embodied reservoir computing, and fully fluidic control for autonomous soft robotics.

Prakash and Gershenfeld demonstrated that bubbles in a microfluidic channel can interact non-linearly to perform universal Boolean logic (AND, OR, NOT) and signal synchronization. This established the field of digital microfluidic computing, where the bit is a physical packet of matter \cite{Prakash2007}. This established a fluidic computing paradigm in which the bit is not an electrical state but a physical packet of matter moving through a designed hydraulic circuit.

\begin{figure}
    \centering
    \includegraphics[width=1\linewidth]{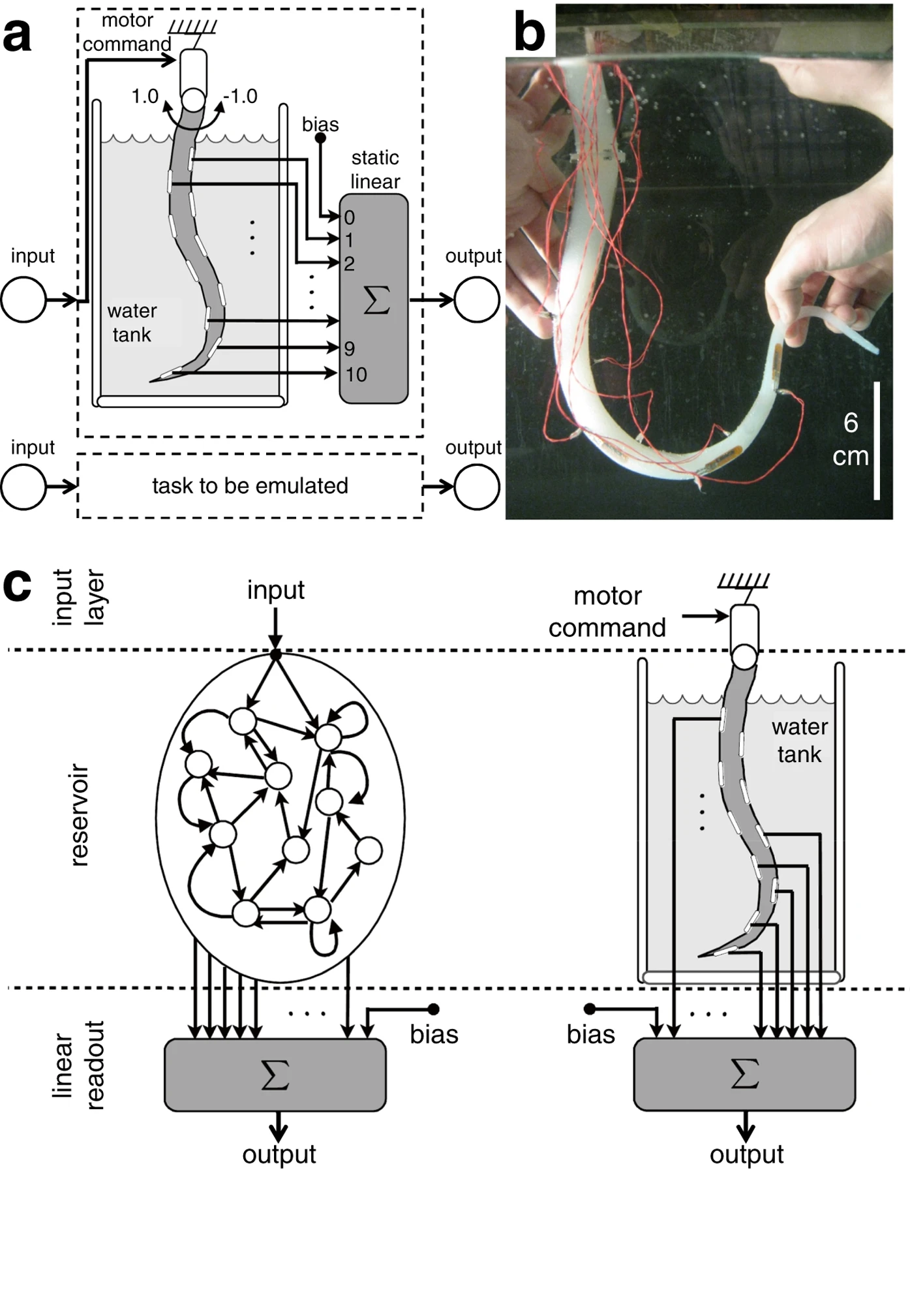}
    %\caption{Physical reservoir computing using a soft silicone arm. (a) The information processing loop: motor commands actuate the arm, while embedded bend sensors capture the complex body dynamics to serve as reservoir states. (b) The experimental soft arm prototype. (c) Comparison showing how the physical soft body replaces the abstract reservoir layer of a conventional Echo State Network. Reprinted from \cite{Nakajima2015}, licensed unter CC-BY 4.0.}
    \caption{Physical reservoir computing using a soft silicone arm. (a) Information-processing loop: motor commands actuate the arm, while embedded bend sensors capture the resulting body dynamics as reservoir states. (b) Experimental prototype of the soft arm. (c) Conceptual comparison showing how the physical soft body replaces the abstract reservoir layer of a conventional Echo State Network. The figure illustrates the central idea of embodied reservoir computing: computation emerges from the intrinsic transient dynamics of the compliant physical system, while only the readout is trained. Reprinted from \cite{Nakajima2015}, licensed unter CC-BY 4.0.}
    \label{fig:silicon_arm}
\end{figure}

A complementary regime is reservoir-style computation in compliant fluidic bodies. Nakajima et al. showed that the nonlinear dynamics of a soft silicone arm, as well as related fluidic networks, can be exploited as physical reservoirs for computation \cite{Nakajima2014,Nakajima2015}. Motor commands perturb the body, embedded bend sensors record the resulting transient dynamics, and only the readout layer is trained, enabling tasks such as chaotic time-series prediction and autonomous gait generation. Figure \ref{fig:silicon_arm} illustrates this principle and highlights how the soft body itself replaces the abstract reservoir layer of a conventional Echo State Network.

A third line of work demonstrates fully fluidic embodied autonomy. Wehner et al. integrated microfluidic logic into the ``Octobot'', a soft robot capable of self-powered movement without semiconductor electronics \cite{Wehner2016}. Building on such principles, Preston et al. realized pneumatic logic gates and ring oscillators that were embedded directly into soft robotic bodies to achieve closed-loop locomotion and obstacle navigation without electronic processors \cite{Preston2019}. These systems show that fluidic substrates can implement not only isolated logic operations or reservoir dynamics, but also autonomous control loops in which sensing, computation, and actuation are tightly integrated in the material system itself.

%\begin{figure}
%    \centering
%    \includegraphics[width=\columnwidth]{figures/41586_2016_Article_BFnature19100_Fig1_HTML.png}
%    \caption{Integrated fabrication of the entirely soft Octobot. The assembly process combines a pre-fabricated microfluidic controller with embedded 3D printing (EMB3D) of catalytic and fugitive inks directly into the elastomer body, creating the internal fuel and actuation networks required for autonomous operation. Reprinted from \cite{Wehner2016}.} %% license is very expensive, maybe excahnge image, z.B. von https://www.nature.com/articles/srep10487 und dann in der vorherigen subsection
%    \label{fig:octobot}
%\end{figure}

\subsection{Engineering Constraints}
The practical deployment of microfluidic neural networks is defined by a trade-off between integration and speed.
First, due to viscosity and compressibility, signal propagation speeds are limited to the speed of sound in the fluid or slower. This restricts operation frequencies to the Hz to kHz range, making them unsuitable for high-speed data processing but ideal for biological timescales.
Second, integration density is limited. While individual channels are small, the control infrastructure (valves, pumps) scales poorly. Large-scale integration requires multilayer soft lithography, which increases fabrication complexity.
Unlike electronics, however, microfluidics is inherently robust to electromagnetic interference (EMI) and radiation, making it attractive for extreme environments \cite{Preston2019}.

\subsection{Position within Physical Neural Computing}
Microfluidic neural systems occupy an intermediate regime between chemical computing and solid-state hardware. They offer richer dynamics and stronger embodiment than electronic substrates, while remaining deterministic and engineerable compared to biochemical systems. Their primary relevance lies in applications requiring tight integration of sensing, computation, and actuation in fluidic or biological environments, such as ``lab-on-a-chip'' diagnostics or soft robotics, rather than in high-speed inference \cite{Whitesides2006}.

\section{Iontronics and Nanofluidic Neural Computing}
\label{sec:iontronics}

While microfluidics mimics hydraulic machinery, iontronics mimics the nervous system. By transitioning from the microscale to the nanoscale ($< 100$ nm), surface-to-volume ratios increase drastically, entering a regime where fluid transport is dominated by surface charges rather than bulk continuum mechanics. The information carrier shifts from the fluid mass to the specific ions ($\text{Na}^+, \text{K}^+, \text{Cl}^-$) dissolved within it \cite{Bocquet2020}. This substrate provides the technological bridge between solid-state electronics and biological wetware, enabling functional ``artificial axons'' and ionic transistors that operate using the brain's native signaling medium \cite{Robin2021}.

\subsection{The Substrate Principle}
The computational engine of iontronics is the Electrical Double Layer (EDL). When an electrolyte contacts a charged surface, counter-ions accumulate to screen the surface charge, forming a layer with a characteristic thickness known as the Debye length ($\lambda_D$). In channels comparable in size to the Debye length ($h \approx 2\lambda_D$), the EDLs from opposing walls overlap, rendering the entire channel unipolar \cite{Daiguji2005}.

Crucially, transport at this scale defies classical hydrodynamics. Kavokine et al. demonstrated that in carbon nanotubes, water flows with vanishingly low friction due to quantum mechanical coupling between fluid fluctuations and the electron gas of the wall. This ``quantum friction'' effect enables transport rates orders of magnitude higher than continuum predictions, providing the high-speed ionic throughput necessary for neuromorphic spiking \cite{Kavokine2022}.

\subsection{Mapping Neural Primitives}
Iontronics maps neural functions to the electro-kinetic equations (Poisson-Nernst-Planck) governing ion flux.

\subsubsection{Synaptic Weights as Ionic Conductance}
In an iontronic circuit, the synaptic weight $w_{ij}$ is physically instantiated as the ionic conductance $G$ of a nanopore. According to Ohm's law for electrolytes ($I = G \cdot V$), the signal is the product of conductance and voltage. Unlike in bulk fluids, this conductance is highly nonlinear and history-dependent. Recent work by Robin et al. demonstrated ``Ion-Shuttling Memristors'' using 2D nanofluidic channels. Here, the reversible intercalation of ions into graphite layers or the formation of ion clusters creates a hysteretic conductance response, enabling non-volatile analog weight storage similar to Long-Term Potentiation (LTP) in biological synapses \cite{Robin2023}. Extending this capability, Ismail et al.\ recently demonstrated fully programmable memristors based on two-dimensional nanofluidic channels, proving that stable analog weights can be precisely tuned in aqueous environments \cite{Ismail2025}.

\subsubsection{Activation and Rectification}
The non-linear activation function required for deep learning is realized through geometric asymmetry. Conical nanopores act as Ion Current Rectifiers (ICR), where the interplay between electro-osmotic flow and the EDL creates a diode-like current-voltage characteristic. This provides a physical implementation of the Rectified Linear Unit (ReLU), permitting current flow in only one direction above a threshold voltage \cite{Daiguji2005}.

\subsection{Architectures and I/O Interfaces}
A key goal of neuromorphic engineering is the physical realization of the ``Integrate-and-Fire'' neuron. Robin, Kavokine, and Bocquet demonstrated that this behavior emerges naturally in nanofluidic slits. Their system couples two distinct transport channels, one as a model for the fast sodium and the other for the slow potassium channels of a biological axon, through an electronic feedback loop. Under a constant voltage input, ions accumulate (the ``integration'') until a threshold is reached, triggering a negative differential resistance event that discharges the potential in a sharp spike (the ``firing''). This device produces self-sustained trains of action potentials structurally identical to the Hodgkin-Huxley model, proving that spiking dynamics are an emergent property of confined ion transport \cite{Robin2021}.

The primary architectural advantage of iontronics lies in its interface. Since the computation uses the same ions ($\text{Na}^+, \text{K}^+$) as living cells, iontronic devices can communicate directly with biological tissue. This avoids the need for transduction layers required by electronic devices and minimizes the risk of toxic redox reactions, paving the way for ``smart prosthetics'' where the computational hardware is chemically indistinguishable from the biological neural network it regulates \cite{Bocquet2020}.

\subsection{Training and Design Paradigms}
Training and design in iontronic neural systems follow the general paradigms of physical learning, but are shaped by the fact that “weights” and state variables emerge from coupled electrokinetics rather than from explicitly addressable parameters. As consolidated in recent review literature, effective design therefore requires \emph{physico--chemical co-design}: geometry (e.g., slit height/length, tapering), interfacial properties (surface charge, functionalization), and electrolyte conditions (ionic strength, pH, valence) jointly set electric-double-layer overlap, selectivity, and concentration polarization, which in turn determine nonlinearity and the volatility/retention trade-off \cite{Daiguji2005,Bocquet2020,Luo2025Iontronics}. Accordingly, iontronic systems instantiate all three major categories introduced in Section~II.B, namely ex-situ digital-twin design, hybrid digital--physical training, and in-materio adaptation. In practice, three strategies dominate. (i) \emph{In-silico / digital-twin workflows} use reduced transport models to guide geometry and operating points and then rely on empirical calibration to compensate fabrication tolerances and surface-chemistry drift \cite{Bocquet2020,Luo2025Iontronics}. (ii) \emph{Hybrid training via reservoir computing} exploits transient ionic dynamics as a fixed high-dimensional projection, while only a simple readout is trained digitally; this is particularly attractive because it is robust to device variability and avoids requiring accurate gradients inside wet, nanoscale hardware \cite{Kamsma2024,Stepney2024}. (iii) \emph{In-materio adaptation} leverages history-dependent conductance arising from ionic rearrangements and interfacial state changes to realize synapse-like plasticity primitives; here, “training” is largely implemented through stimulation and programming protocols (pulse trains, duty cycles) and through electrolyte tuning \cite{Robin2023,Ismail2025}. It should be noted that the emergence of \emph{programmable} nanofluidic memristors strengthens this co-design perspective by showing that qualitatively different hysteresis regimes can be accessed within the same hardware through controlled changes in electrolyte composition and operating conditions, turning device programming into an explicit design handle for neuromorphic functionality \cite{Ismail2025}. Overall, digital-twin workflows benefit from model-guided design but remain sensitive to calibration drift and surface-chemistry variability; reservoir-style approaches are well matched to rich transient ion dynamics but sacrifice end-to-end optimization of internal states; and in-materio adaptation offers substrate-native plasticity, yet currently remains constrained by limited controllability, observability, and update speed.

\subsection{Representative Demonstrations}
\begin{figure}
    \centering
    \includegraphics[width=1\linewidth]{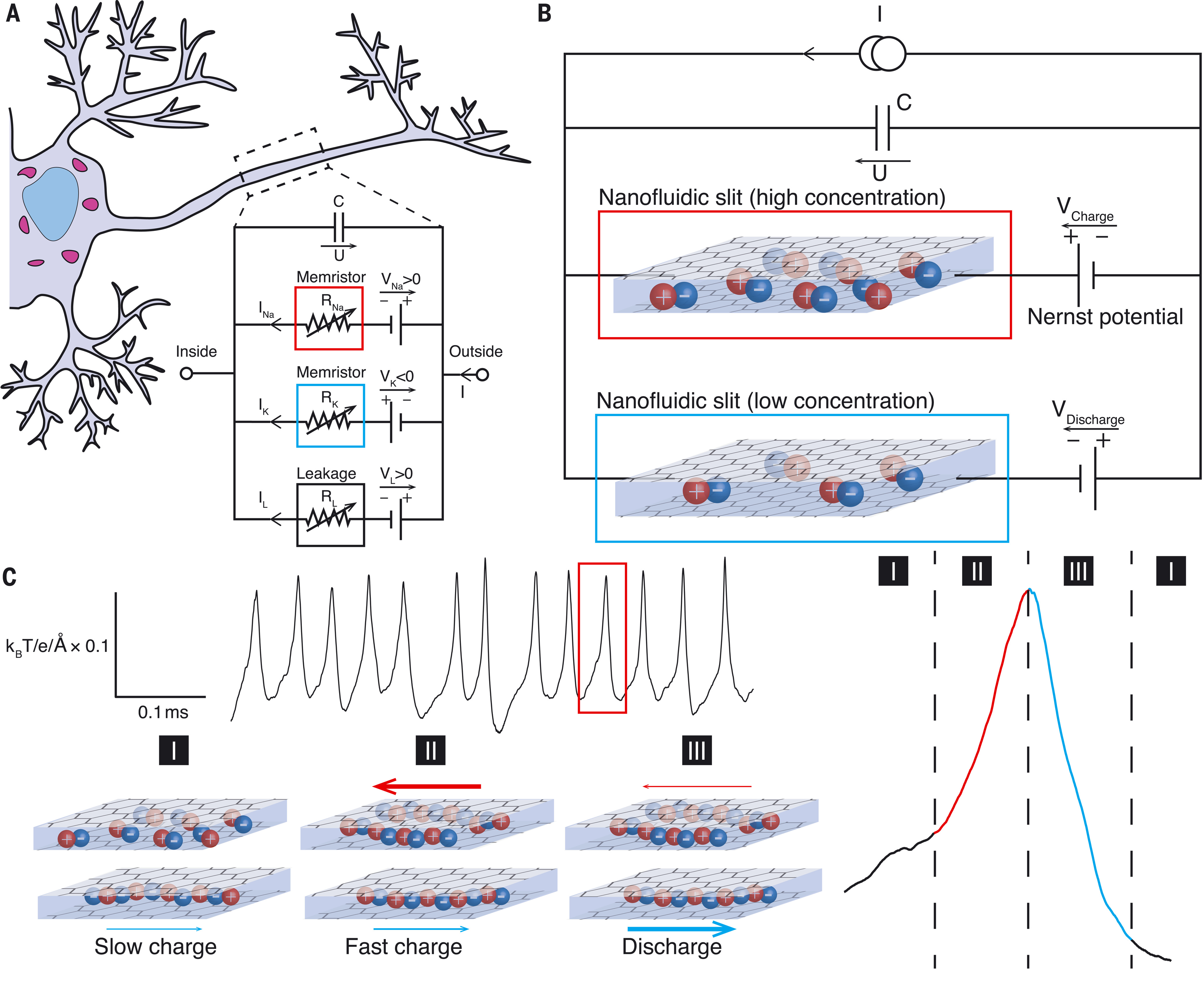}
    \caption{Artificial Hodgkin-Huxley-like neuron based on 2D ionic memristors.
(A) Circuit representation of the Hodgkin-Huxley model showing the ionic branches and corresponding Nernst potentials. (B) Brownian-dynamics simulation of the ionic prototype, consisting of two slits connected to reservoir pairs with different ionic concentrations. (C) Spontaneous voltage spikes generated by the prototype, with the right panel illustrating the quiescent, initiation, and discharge phases of a single spike. The figure exemplifies how nanofluidic transport asymmetry and ionic coupling can reproduce biologically familiar excitability in a fully physical device. Reprinted from \cite{Robin2021} with permission from the American Association for the Advancement of Science.}
    \label{fig:artificial_axon}
\end{figure}

Representative demonstrations in iontronic neural computing now span ionic spiking elements, synapse-like memory devices, reservoir-style processing, configurable memristive primitives, and biointerfaced spiking components.
At the device-physics end, Hodgkin--Huxley-like spiking dynamics have been reproduced in angstrom-scale nanofluidic slit architectures, establishing an “artificial axon” concept where coupled ionic transport nonlinearities yield autonomous spike trains \cite{Robin2021}. Figure~\ref{fig:artificial_axon} illustrates this principle and highlights how differential ionic selectivity and coupled transport give rise to quiescent, initiation, and discharge phases that closely resemble biological spiking.

At the learning-synapse end, two-dimensional nanofluidic channels have demonstrated long-term memory and synapse-like dynamics under electrochemical control, motivating the view of iontronics as a route to neuromorphic functionality that is directly rooted in aqueous transport physics \cite{Robin2023}. 
On the systems side, a particularly clear engineering demonstration is fluidic reservoir computing with iontronic nanochannels, which achieves handwritten-digit classification with an experimentally reported accuracy of 81\% under a simple readout protocol, highlighting how transient concentration polarization can act as short-term memory without full in-situ weight programming \cite{Kamsma2024}. 

Finally, programmability at the device level has been strengthened by nanofluidic memristors whose hysteresis type can be switched by tuning electrolyte chemistry and operating conditions, suggesting a practical route to configurable iontronic primitives for larger circuits \cite{Ismail2025}. Bio-interfacing directions are illustrated by nanofluidic spiking synapses based on PEDOT:PSS (a popular composite material in this field) confined in nanopores, providing spiking behavior in an aqueous environment compatible with bioelectronic integration \cite{Xu2024}.

\subsection{Engineering Constraints}
The practical application of iontronics is defined by a fundamental trade-off between bio-compatibility and computational speed. The primary limitation is the inherent mobility of the charge carriers; ions move approximately $10^5$ times slower than electrons in silicon. Consequently, iontronic circuits operate in the kHz regime, which renders them unsuitable for general-purpose number crunching but perfectly matches the millisecond timescale of biological action potentials.

Furthermore, at the nanoscale, systems are subject to significant flicker noise and thermal fluctuations (Brownian motion). While traditionally viewed as a defect in deterministic logic, this stochasticity is increasingly exploited as a computational resource for probabilistic inference \cite{Robin2021}. A significant manufacturing hurdle remains the fabrication of circuits at the single-nanometer scale %—such as drilling precise nanopores in 2D materials like MoS$_2$—
which currently limits large-scale integration compared to established CMOS processes \cite{Robin2023}.

\subsection{Position within Physical Neural Computing}
Iontronics occupies a unique niche as the ``missing link'' between abiotic electronics and biotic wetware. It is the only substrate that speaks the native language of biology. Its future lies not in competing with silicon for FLOPs, but in domains requiring high-fidelity bio-integration, acting as the computational interface between artificial intelligence and the human nervous system.

% PART IV ----------------------------------------------------------
%\part{Synthesis and Outlook}

% ------------------------------------------------------------------

\section{Cross-Substrate Benchmarking}
\label{sec:benchmarking}

Comparing physical neural substrates is inherently difficult: reported performance figures depend not only on the substrate physics, but also on the experimental interface (encoding/decoding), the training protocol, and what is counted as system energy (core physics only vs.\ peripheral actuation and transduction). Rather than proposing a definitive benchmark, we introduce, in this section, a \textit{first-order comparative framework}. This approach aligns with the methodology advocated by Stepney, whose recent tutorial emphasizes the need to rigorously define the ``physical reservoir'' properties to avoid false positives in computational capacity \cite{Stepney2024}. We therefore propose a framework that (i) uses a small set of standardized tasks to assess capability, and (ii) positions substrates in a common, physically interpretable landscape defined by characteristic speed, energy scale, and plasticity. The goal is to make trade-offs explicit and to identify where each substrate is a natural fit, not to crown a universal winner.

\subsection{Standardized tasks: capability tests instead of a single scoreboard}
\label{sec:physical_turing_tests}

A single benchmark cannot fairly represent the diversity of physical substrates (e.g., a DNA network optimized for molecular diagnostics versus a photonic circuit optimized for GHz signal processing). We therefore propose two minimal capability domains that recur across the literature and collectively probe the key neural primitives discussed throughout this review: (1) \emph{static classification}, requiring separability in a high-dimensional state space, and (2) \emph{dynamic prediction/control}, based on fading memory and nonlinear temporal processing. These are not meant as rigid standards, but as \emph{capability tests} that allow comparing whether a substrate can execute core neural functions at all. We do not claim that the proposed framework is complete; rather, we view it as a starting point that makes otherwise implicit trade-offs explicit and debatable.

The canonical reference for static classification is MNIST-style image classification (28$\times$28) or compressed variants thereof, which stresses parallel input encoding and stable linear/nonlinear separation. Demonstrations exist for DNA strand-displacement classifiers on compressed MNIST-like inputs \cite{CherryQian2018}, for deep physical neural networks trained with physics-aware backpropagation in wave-based platforms \cite{Wright2022}, and for integrated photonic neural networks based on coherent interference \cite{Shen2017}. In electronic ``computational memory'', memristive/PCM crossbars have demonstrated large-scale neural inference and learning with weights stored as conductances \cite{Burr2014,LeGallo2018,Joshi2020}.

In the second domain, temporal tasks such as Mackey--Glass prediction or NARMA-$k$ emulation probe fading memory and nonlinearity; closed-loop control tasks additionally probe stability under feedback. Reservoir computing has long provided substrate-agnostic baselines (e.g., liquid/wave reservoirs) \cite{Fernando2003,Appeltant2011}. Mechanical and soft robotic systems can embody dynamics directly (e.g., compliant bodies as reservoirs) \cite{Nakajima2015}, while microfluidic and pneumatic circuits have demonstrated autonomous oscillations and gait-like pattern generation driven by the system physics \cite{Preston2019,Drotman2021}. Bio-hybrid systems (neuronal cultures) have shown closed-loop learning in real-time control environments \cite{Kagan2022}. Photonic reservoirs offer high-bandwidth temporal processing via delay dynamics \cite{Paquot2012}.

On a practical note, the dominant limitation for many substrates is not the internal physics but the I/O mapping: providing 784 independent analog inputs (MNIST) is trivial in electronics, but a major engineering challenge for pressure-driven microfluidics; conversely, supplying molecular inputs to a silicon classifier often requires expensive transduction that wetware can avoid. These asymmetries motivate the second element of the framework: a physically grounded landscape view.

To make these capability tests more practically usable, Table~\ref{tab:benchmark_protocol} summarizes a minimal reference protocol. The protocol is intentionally modest: it is not meant to prescribe a single benchmark for all substrates, but to define a common reporting baseline. Substrate-specific adaptations are unavoidable, for example when a molecular system cannot expose hundreds of independent inputs or when a mechanical body naturally processes low-dimensional continuous signals. However, deviations from the reference protocol should be stated explicitly, so that differences in performance can be attributed to the substrate rather than to hidden choices in encoding, noise injection, or energy accounting.

\begin{table*}[t]
\centering
\caption{Minimal reference protocol for cross-substrate benchmarking of physical neural systems. The proposed values are intended as reporting defaults rather than rigid standards; substrate-specific deviations should be stated explicitly.}
\label{tab:benchmark_protocol}
\footnotesize
\begin{tabularx}{\textwidth}{@{}p{2.9cm} p{4.2cm} p{4.2cm} X@{}}
\toprule
\textbf{Protocol element} &
\textbf{Static classification task} &
\textbf{Dynamic prediction/control task} &
\textbf{Reporting requirement} \\
\midrule
Input dimensionality &
Reference input: MNIST-style $28\times 28$ images where feasible; compressed variants with $d \in \{16,64,256\}$ for substrates with limited I/O. &
Low-dimensional time series with $d \in [1,16]$ input channels; sequence length $T=100$--$1000$ samples or substrate-specific physical time horizon. &
Report the physical encoding of each input dimension, compression method if used, and whether dimensions are applied in parallel or sequentially. \\

Noise levels &
Evaluate clean inputs and additive input/readout noise at representative SNR levels, e.g., 40\,dB, 20\,dB, and 10\,dB where experimentally meaningful. &
Evaluate process or readout perturbations at matched SNR levels; for closed-loop tasks additionally report disturbance amplitude and duration. &
State whether noise is injected digitally, physically at the input, internally by the substrate, or only at readout; report mean and variance over repeated trials. \\

Training regime &
Report whether parameters are obtained by ex-situ digital training, device-aware training, in-situ adaptation, reservoir readout training, or closed box optimization. &
Same categories; additionally report whether temporal memory is intrinsic to the substrate or provided by external delay/state variables. &
Specify train/test split, number of training examples or episodes, number of physical evaluations, and whether the same physical instance is reused. \\

Performance metric &
Classification accuracy, confusion matrix, and robustness degradation under noise/compression. &
Prediction error (e.g., NRMSE) for time-series tasks; task reward, tracking error, or stability margin for control tasks. &
Report mean, standard deviation, number of runs, and whether variability is dominated by device noise, fabrication spread, biological variability, or environmental drift. \\

Latency and throughput &
Time from input presentation to stable or readable output, including settling or relaxation time. &
Sampling interval, memory horizon, closed-loop reaction time, and maximum stable control/update rate. &
Separate intrinsic substrate time constants from encoding, actuation, readout, and digital post-processing latency. \\

Energy accounting &
Energy per inference or per classified sample. &
Energy per prediction step, control step, or physical episode. &
State the accounting boundary: core substrate only, substrate plus actuation/readout, or full system including ADC/DAC, pumps, lasers, cryogenics, incubation, or life support. \\

I/O and interface overhead &
Number and type of physical input/output channels; transduction required for applying and reading inputs. &
Same, with additional reporting of feedback-loop interfaces for control tasks. &
Report whether the dominant cost lies in the substrate dynamics or in peripheral sensing, actuation, conversion, and stabilization hardware. \\
\bottomrule
\end{tabularx}
\end{table*}

This protocol also clarifies the interpretation of the landscape comparison below: speed, energy, and plasticity should not be read as isolated material constants, but as system-level quantities whose meaning depends on the stated task, input encoding, noise regime, and accounting boundary.

\subsection{Quantitative comparison: the physics--compute landscape}
\label{sec:physics_compute_landscape}

Table~\ref{tab:landscape} aggregates order-of-magnitude characteristics derived from representative demonstrations discussed across Parts 2 and 3. We emphasize that these values are \emph{illustrative} rather than definitive: they indicate characteristic regimes, not optimized engineering limits. Where possible we cite experimental demonstrations; where literature varies widely (e.g., optical energy per MAC depending on whether tuning and detection are included), we report broad ranges and state what is being counted.

\begin{sidewaystable}[p]
\centering
\caption{Comparative landscape of physical neural substrates.}
\label{tab:landscape}

\small

\resizebox{\textheight}{!}{%
\begin{tabular}{@{}llllllp{6cm}@{}}
\toprule
\textbf{Substrate} &
\textbf{Info carrier} &
\textbf{Speed (Hz)} &
\textbf{Energy (J/op)} &
\textbf{Plasticity} &
\textbf{Demonstrated capability} &
\textbf{Dominant bottleneck / natural niche} \\
\midrule

Silicon (GPU/TPU) &
Electrons &
$\sim10^{9}$ &
$\sim10^{-12}$ &
Simulated &
Broadly solved &
Excellent generality; heavy transduction for ``non-digital'' inputs. \\

Memristive / PCM in-memory &
Electrons (conductance states) &
$\sim10^{6}$--$10^{9}$ &
$\sim10^{-15}$--$10^{-12}$ &
Device / mixed-precision &
Large-scale inference and learning demos &
Array nonidealities, ADC/DAC overhead; best for dense in-memory MAC near sensors \cite{Burr2014,LeGallo2018,Joshi2020,Sebastian2020}. \\

Ferroelectric / In-Sensor &
Polarization domains &
$\sim10^{7}$--$10^{9}$ &
$\sim10^{-15}$ &
Hysteresis / Domain walls &
In-sensor tactile computing &
Imprint effects;
best for zero-latency edge sensing \cite{Chen2025}. \\

Spintronic / Superconducting &
Spin / Flux quanta &
$\sim10^{9}$--$10^{11}$ &
$\sim10^{-19}$--$10^{-16}$ &
Intrinsic / STDP &
Skyrmion reservoirs; 50\,GHz SNNs &
Cryogenics (SC), thermal stability;
best for ultra-low power HPC \cite{Pinna2020,Schneider2025_SelfTraining}. \\

Photonic neural hardware &
Photons (phase/amplitude) &
$\sim10^{9}$--$10^{11}$ &
(core $\ll10^{-12}$; system-dependent) &
Mostly configured (hybrid) &
High-speed classification; photonic reservoirs &
Nonlinearity and opto-electronic interface; best for ultrafast inference and RF/optical signal processing \cite{Shen2017,Tait2017,Paquot2012,Prucnal2016}. \\

Mechanical / acoustic &
Phonons / elastic waves &
$\sim10^{3}$--$10^{6}$ &
$\sim10^{-9}$--$10^{-6}$ &
Coupled / emergent &
Wave-based classification; embodied dynamics &
Manufacturability, damping, sensing; best for smart structures and embedded vibration processing \cite{Wright2022,Nakajima2015}. \\

Microfluidic / pneumatic &
Fluid mass / pressure &
$\sim10^{0}$--$10^{2}$ &
$\sim10^{-6}$--$10^{-3}$ &
In-situ / fixed logic &
Oscillators, gait generation, soft control &
I/O fan-in and actuation counted in energy; best for untethered soft robotics in harsh EM environments \cite{Preston2019,Drotman2021}. \\

Iontronic / nanofluidic &
Ions (charge transport) &
$\sim10^{2}$--$10^{4}$ &
$\sim10^{-16}$--$10^{-12}$ &
Memristive / STDP-like &
Spiking primitives (axon-like) &
Nanofabrication precision and noise; best for bio-interfaces and low-power spiking primitives \cite{Robin2021,Robin2023}. \\

DNA (wet molecular) &
Molecules / concentrations &
$\sim10^{-5}$--$10^{-3}$ &
$\sim10^{-19}$--$10^{-16}$ &
Static / slow adaptive &
Compressed MNIST-like classification &
Latency and readout; best for in-sample diagnostics where transduction dominates silicon \cite{CherryQian2018}. \\

Chemical (reaction--diffusion, BZ) &
Wavefronts / chemical potential &
$\sim10^{-2}$--$10^{0}$ &
$\sim10^{-8}$--$10^{-5}$ &
Emergent &
Spatial computing (maze, Voronoi) &
Programmability and cascadability; best for analog spatial geometry problems \cite{TothShowalter1995,Steinbock1995}. \\

Living (cells / neurons) &
Ions + biochemical regulation &
$\sim10^{1}$--$10^{3}$ &
$\sim10^{-15}$--$10^{-12}$ &
True (STDP, adaptation) &
Closed-loop learning/control (e.g., Pong) &
Reproducibility and maintenance; best for adaptive bio-hybrid controllers and neuroscience-in-the-loop \cite{Kagan2022}. \\

\bottomrule
\end{tabular}
}

\end{sidewaystable}

Following our analysis, we can say that the following observations are robust across the surveyed literature.

\paragraph{Thermodynamic and biochemical limits enable extreme energy efficiency, but at low bandwidth.}
Molecular systems (DNA) and iontronic devices can approach very low physical energy scales per state transition, but their usable throughput is constrained by diffusion, reaction kinetics, and stochastic fluctuations \cite{CherryQian2018,Robin2021}. This makes them attractive when \emph{transduction dominates} and sample rates are intrinsically slow (e.g., diagnostics), rather than for high-frequency control.

\paragraph{Wave and field substrates excel at bandwidth, but pay for nonlinearity and interfacing.}
Photonic and mechanical wave systems implement linear transformations ``for free'' via propagation and interference, supporting extremely high bandwidth for inference and signal processing \cite{Shen2017,Wright2022}. However, strong, compact nonlinearities and stable large-scale programmability remain system-level challenges, often pushing architectures toward hybrid electro-optic control \cite{Prucnal2016,Tait2017}.

\paragraph{In-memory electronic substrates sit at a favorable midpoint if co-designed with algorithms.}
Memristive and PCM crossbars offer dense, scalable MAC at moderate-to-high bandwidth while reducing memory movement. Yet their advantages are inseparable from mixed-precision and device-aware algorithm design, because drift, noise, and nonlinear updates violate ideal arithmetic \cite{LeGallo2018,Joshi2020,Sebastian2020}. In this sense, they represent the most ``engineering-ready'' route to physical neural hardware at scale.

\subsection{Synthesis: computational impedance matching}
\label{sec:impedance_matching}

The landscape view suggests that physical neural substrates are not drop-in replacements for silicon, but specialized accelerators whose value emerges when the physics of the substrate matches the physics of the problem. We term this \emph{computational impedance matching}: choose the substrate that minimizes unavoidable transduction and that naturally implements the dominant dynamics of the task.

\begin{itemize}
\item \textbf{Chemical-to-decision pipelines.} If the input is already molecular (biomarkers, metabolites), DNA or biochemical networks can compute \emph{in situ} without expensive sensing and digitization; long reaction times are acceptable when sample rates are low \cite{CherryQian2018}.
\item \textbf{High-bandwidth inference.} If the task is ultrafast signal processing (RF, optical communications), photonic inference layers can exploit propagation bandwidth, with electronics providing nonlinearity and control \cite{Shen2017,Prucnal2016}:
\item \textbf{Dense edge inference and adaptive sensing.} If the bottleneck is moving weights and activations between memory and compute, memristive/PCM computational memory can provide energy-efficient MAC close to sensors, provided algorithms are co-designed for device nonidealities \cite{LeGallo2018,Joshi2020}.
\item \textbf{Embodied control.} If the task is to generate or stabilize physical motion in soft systems, microfluidic/pneumatic or mechanical substrates can embody the control dynamics directly, often counting ``computation'' as mechanical work \cite{Preston2019,Drotman2021,Nakajima2015}.
\item \textbf{Bio-hybrid adaptivity.} If continuous adaptation under uncertainty is paramount, living neural tissue can learn in closed-loop settings, at the cost of reproducibility and maintenance overhead \cite{Kagan2022}.
\end{itemize}

In short, a useful benchmark for physical intelligence must not only score accuracy on a task, but must also characterize the interface cost and the physics-algorithm fit. The framework above provides a starting point for that comparison and highlights where further standardized reporting (especially for end-to-end energy and I/O costs) is needed.

\section{Conclusion and Outlook}
\label{sec:outlook}

In this review, we have surveyed a broad spectrum of physical substrates for neural computation, ranging from molecular reaction networks and living cellular systems to solid-state electronic, photonic, mechanical, and fluidic hardware. Taken together, these developments suggest that neural computation is not tied to a particular material platform, but rather emerges whenever a physical system provides three ingredients: rich nonlinear dynamics, a mechanism for weighted interaction between degrees of freedom, and a process for adaptation or effective training \cite{Sebastian2020,Prucnal2016,Smirnova2023}. The diversity of substrates explored in the preceding parts demonstrates that these ingredients can be realized through fundamentally different physical mechanisms: charge transport, wave interference, chemical kinetics, mechanical deformation, fluid flow, or biological plasticity \cite{CherryQian2018,Paquot2012,Wright2022,Nakajima2015,Kagan2022}.

\subsection{From algorithms to physical design principles}

A recurring theme across all parts of this article is that performance in physical neural systems does not arise from abstract learning algorithms alone, but from careful co-design of algorithms, architectures, and materials \cite{LeGallo2018,Joshi2020,Shen2017}. In digital accelerators, the physical layer is typically treated as an implementation detail. In contrast, physical neural computing elevates the substrate to a first-class design variable: device physics determines which operations are cheap, which sources of noise are unavoidable, and which forms of plasticity are accessible \cite{Sebastian2020,Friston2010}.

We see the following general design principles emerge from the surveyed literature:

\begin{itemize}
  \item \textbf{Exploit native dynamics.} Systems achieve their highest efficiency when the dominant computation (e.g., linear transformation, diffusion, wave propagation, or relaxation) coincides with the natural dynamics of the substrate, rather than being emulated through control circuitry \cite{Paquot2012,Nakajima2015}.
  \item \textbf{Locality of interaction.} Many substrates support only local coupling (chemical reactions, mechanical stress, nearest-neighbor electrical conduction). Architectures that embrace this locality are more scalable and robust than those attempting to impose global connectivity \cite{CherryQian2018,Adamatzky2010}.
  \item \textbf{Tolerance to imprecision.} Noise, drift, and variability are intrinsic to most physical substrates. Successful systems therefore rely on learning rules, redundancy, or reservoir-style computation that remain functional under significant uncertainty \cite{LeGallo2018,Boybat2018,Paquot2012}.
  \item \textbf{Interface-aware design.} In many demonstrations, the energetic and temporal cost of sensing, actuation, and signal conversion dominates the cost of the internal computation. End-to-end efficiency must therefore be evaluated at the level of the complete physical-digital loop \cite{CherryQian2018,Shen2017,Preston2019}.
\end{itemize}

These principles suggest that the traditional separation between ``algorithm'' and ``hardware'' becomes blurred in physical intelligence: the computational model must be chosen with the substrate in mind, and vice versa \cite{Sebastian2020}.

\subsection{The role of benchmarking and comparative evaluation}

The benchmarking framework we introduced in Section \ref{sec:benchmarking} provides a first step toward systematic comparison of heterogeneous substrates \cite{Fernando2003,Appeltant2011}. While no single task or metric can capture the full richness of physical neural systems, standardized capability tests and physically interpretable performance dimensions (speed, energy scale, plasticity, and interface cost) enable more meaningful cross-domain discussion than isolated demonstrations \cite{Wright2022,Shen2017,Burr2014,Kagan2022}.

In the long term, progress in this field will benefit from community practices analogous to those in conventional machine learning: shared datasets adapted to physical constraints, reference tasks for static and dynamic processing, and transparent reporting of what is included in energy and latency measurements \cite{Sebastian2020,Joshi2020}. In the absence of such conventions, performance comparisons between such disparate technologies like molecular classifiers, photonic inference engines, and in-memory electronic accelerators will lack the rigorous framework necessary for commensurate evaluation and cumulative progress.

\subsection{Application domains and realistic expectations}

The preceding chapters make clear that physical neural systems are unlikely to displace digital computing as a universal platform. Their value instead lies in domains where their physical properties provide decisive advantages and where the required levels of reliability, stability, and environmental robustness remain compatible with the substrate:

\begin{itemize}
  \item \textbf{In-sample molecular decision making,} where chemical or DNA-based networks can operate directly on biochemical inputs without costly transduction \cite{CherryQian2018}.
  \item \textbf{Ultrafast signal processing,} where photonic systems exploit wave propagation to perform linear inference at bandwidths unattainable electronically \cite{Shen2017,Prucnal2016}.
  \item \textbf{Energy-constrained edge inference,} where memristive and phase-change memories reduce data movement by colocating storage and computation \cite{LeGallo2018,Joshi2020}.
  \item \textbf{Embodied control in soft and adaptive machines,} where mechanical or fluidic substrates naturally integrate computation with actuation \cite{Nakajima2015,Preston2019,Drotman2021}.
  \item \textbf{Bio-hybrid systems,} where living neural tissue provides continuous adaptation in uncertain environments \cite{Kagan2022,Smirnova2023}.
\end{itemize}

At the same time, these niches differ strongly in their path to deployment. Some substrates are most promising for highly specialized laboratory or medical settings, where direct operation in the native physical domain outweighs limited speed or reusability; others are closer to technological translation because they can be fabricated, packaged, and interfaced with greater repeatability. Realistic expectations therefore require matching each substrate not only to a computational advantage, but also to its achievable operating stability over time.

We believe that recognizing these niches is essential to avoid unrealistic expectations and to guide investment toward problems where physical intelligence can offer genuine benefits over mature digital technologies. Equally important is to acknowledge that the most compelling application domains are those in which the reliability requirements are aligned with the physical operating regime of the substrate rather than inherited uncritically from conventional digital hardware.

\subsection{Reliability, stability, and transition to real-world applications}

%Despite rapid progress, several fundamental challenges remain that must be addressed to transition these systems toward practical applications. A primary concern is scalability, as many physical demonstrations are currently constrained by fabrication limits, control complexity, or inherent biological variability \cite{Sebastian2020,Smirnova2023}. This issue is closely linked to problems of programmability and reproducibility; particularly in wetware and bio-hybrid systems, achieving identical behavior across nominally identical devices remains difficult to guarantee \cite{Smirnova2023,Kagan2022}. Furthermore, the development of efficient learning rules for training at scale constitutes a major research gap, especially when dealing with non-ideal and only partially observable substrates \cite{LeGallo2018,Friston2010,Momeni2025}. Finally, long-term stability remains a critical bottleneck, as drift, fatigue, and aging effects inherent to almost all non-digital substrates complicate deployment beyond controlled laboratory settings \cite{Joshi2020,Robin2023}. Overcoming these multifaceted obstacles will necessitate a concerted interdisciplinary effort across machine learning, materials science, device physics, biology, and control theory.

Despite rapid progress, the transition of physical neural systems from laboratory demonstrations to practical applications depends not only on computational performance, but on whether reliable and stable operation can be maintained under realistic conditions. Across substrates, the central challenge is that computation is embodied in matter whose properties drift, age, fluctuate, or depend sensitively on the environment. As a result, deployment requires more than proof-of-principle functionality: it requires calibration strategies, tolerance to variability, and application scenarios whose operational demands match the physical limits of the substrate.

These constraints appear in different forms across the PNN landscape. In molecular and chemical systems, reaction kinetics, cross-talk, reagent depletion, and resetability limit long-term reuse and favor applications such as in-sample diagnostics, programmable assays, or lab-on-chip decision making, where direct biochemical operation can justify lower throughput and stricter environmental control \cite{CherryQian2018}. In living and bio-hybrid systems, biological variability, maintenance overhead, and limited interface bandwidth remain major barriers to standardized deployment, suggesting that near-term impact is more likely in closed-loop experimental platforms, biointerfaces, and scientific model systems than in robust mass-market computing technologies \cite{Kagan2022,Smirnova2023}.

By contrast, solid-state substrates such as memristive, phase-change, and ferroelectric systems are in several respects closer to practical edge-AI integration, but they remain constrained by device drift, stochastic switching, endurance limits, and the energy and area costs of peripheral interfacing \cite{LeGallo2018,Joshi2020}. Photonic systems offer exceptional bandwidth and low-latency inference, yet their transition to real-world use depends critically on calibration stability, thermal robustness, and efficient electro-optical I/O \cite{Shen2017,Prucnal2016}. Mechanical, metamaterial, microfluidic, and iontronic approaches are particularly attractive where sensing, computation, and actuation or biochemical interfacing must be co-designed, but their prospects depend on controlling fatigue, fouling, packaging complexity, and environmental sensitivity over extended operating periods \cite{Nakajima2015,Preston2019,Drotman2021,Robin2023}.

A primary concern is therefore scalability in the broader engineering sense: not merely increasing system size, but doing so while preserving reproducibility, controllability, and acceptable failure rates \cite{Sebastian2020,Smirnova2023}. This issue is closely linked to programmability and observability; particularly in wetware and bio-hybrid systems, achieving identical behavior across nominally identical devices remains difficult to guarantee \cite{Smirnova2023,Kagan2022}. Furthermore, the development of efficient learning rules for training at scale constitutes a major research gap, especially when dealing with non-ideal and only partially observable substrates \cite{LeGallo2018,Friston2010,Momeni2025}. Finally, long-term stability remains a critical bottleneck, as drift, fatigue, and aging effects inherent to almost all non-digital substrates complicate deployment beyond controlled laboratory settings \cite{Joshi2020,Robin2023}.

In our view, the most credible transition paths will therefore be substrate-specific rather than universal. PNN platforms are most likely to succeed first in application niches where their distinctive physics provides a decisive systems advantage and where reliability can be ensured either by environmental control, by hybrid digital supervision, or by a task formulation that tolerates limited precision and gradual drift. Overcoming these multifaceted obstacles will necessitate a concerted interdisciplinary effort across machine learning, materials science, device physics, biology, and control theory.

\subsection{Perspective}

Physical neural computing reframes intelligence as a property that can be distributed across matter, rather than confined to abstract algorithms executed on standardized hardware \cite{Prucnal2016,Smirnova2023}. Whether implemented in molecules, photons, deformable bodies, or living cells, neural computation appears as a general organizational principle of dynamical systems far from equilibrium \cite{Friston2010}. We expect that the most successful systems will be hybrid, combining multiple of these physical substrates rather than relying on a single universal platform. In such systems, digital components will often remain essential not only for orchestration and training, but also for calibration, error mitigation, and the stabilization of otherwise fragile physical dynamics.

From this perspective, current demonstrations should not be judged primarily by their raw performance relative to GPUs or CPUs, but by the new regimes of computation they make accessible: computation at the scale of chemistry, at the speed of light, or embedded directly into adaptive materials \cite{CherryQian2018,Shen2017,Nakajima2015}. As these regimes mature, they may reshape how intelligent systems are engineered, not by replacing digital computers, but by complementing them with substrates whose physics performs part of the computation by default \cite{Sebastian2020}. Their long-term impact, however, will depend on whether these gains can be translated into sufficiently reliable, stable, and maintainable operation outside carefully controlled laboratory settings.

In this sense, we believe that the study of physical neural networks is not only a search for alternative hardware, but also an exploration of the physical limits and manifestations of learning itself.

\section*{Acknowledgments}
The authors acknowledge the use of large language models, specifically Google Gemini and OpenAI ChatGPT, in the preparation of this manuscript. These tools were used to support the literature review on the state of the art for Sections \ref{sec:dna} through \ref{sec:iontronics}. Wherever possible, the papers were accessed and checked individually (the vast majority). Otherwise, the abstracts were used in combination with AI summaries to extract the relevant information. In addition, AI was utilized to assist with the LaTeX formatting of tables and figures and to perform linguistic refinement and stylistic smoothing of the entire text. All AI-generated suggestions and research outputs were critically reviewed, verified, and edited by the authors, who maintain full responsibility for the technical accuracy and integrity of the final content.

{
\printbibliography

@Article{ZahediAy2013,
AUTHOR = {Zahedi, Keyan and Ay, Nihat},
TITLE = {Quantifying Morphological Computation},
JOURNAL = {Entropy},
VOLUME = {15},
YEAR = {2013},
NUMBER = {5},
PAGES = {1887--1915},
URL = {https://www.mdpi.com/1099-4300/15/5/1887},
ISSN = {1099-4300},
DOI = {10.3390/e15051887}
}

@Article{ZahediLangerAy2017,
AUTHOR = {Ghazi-Zahedi, Keyan and Langer, Carlotta and Ay, Nihat},
TITLE = {Morphological Computation: Synergy of Body and Brain},
JOURNAL = {Entropy},
VOLUME = {19},
YEAR = {2017},
NUMBER = {9},
ARTICLE-NUMBER = {456},
URL = {https://www.mdpi.com/1099-4300/19/9/456},
ISSN = {1099-4300},
DOI = {10.3390/e19090456}
}

@article{CherryQian2018,
  author       = {Cherry, Kevin M. and Qian, Lulu},
  title        = {Scaling up molecular pattern recognition with {DNA}-based winner-take-all neural networks},
  journal      = {Nature},
  year         = {2018},
  volume       = {559},
  number       = {7714},
  pages        = {370--376},
  doi          = {10.1038/s41586-018-0289-6}
}

@article{Stepney2024,
  author       = {Stepney, Susan},
  title        = {Physical reservoir computing: a tutorial},
  journal      = {Natural Computing},
  year         = {2024},
  volume       = {23},
  number       = {4},
  pages        = {665--685},
  doi          = {10.1007/s11047-024-09997-y}
}

@article{Law2025,
  author       = {Law, Cheryl S. and Wang, Juan and Nielsch, Kornelius and Abell, Andrew D. and Bisquert, Juan and Santos, Abel},
  title        = {Recent advances in fluidic neuromorphic computing},
  journal      = {Applied Physics Reviews},
  year         = {2025},
  volume       = {12},
  number       = {2},
  pages        = {021309},
  doi          = {10.1063/5.0235267}
}

@article{10.1145/359576.359579,
author = {Backus, John},
title = {Can programming be liberated from the von {N}eumann style? a functional style and its algebra of programs},
year = {1978},
issue_date = {Aug. 1978},
publisher = {Association for Computing Machinery},
address = {New York, NY, USA},
volume = {21},
number = {8},
issn = {0001-0782},
%url = {https://doi.org/10.1145/359576.359579},
doi = {10.1145/359576.359579},
abstract = {Conventional programming languages are growing ever more enormous, but not stronger. Inherent defects at the most basic level cause them to be both fat and weak: their primitive word-at-a-time style of programming inherited from their common ancestor—the von Neumann computer, their close coupling of semantics to state transitions, their division of programming into a world of expressions and a world of statements, their inability to effectively use powerful combining forms for building new programs from existing ones, and their lack of useful mathematical properties for reasoning about programs.An alternative functional style of programming is founded on the use of combining forms for creating programs. Functional programs deal with structured data, are often nonrepetitive and nonrecursive, are hierarchically constructed, do not name their arguments, and do not require the complex machinery of procedure declarations to become generally applicable. Combining forms can use high level programs to build still higher level ones in a style not possible in conventional languages.Associated with the functional style of programming is an algebra of programs whose variables range over programs and whose operations are combining forms. This algebra can be used to transform programs and to solve equations whose “unknowns” are programs in much the same way one transforms equations in high school algebra. These transformations are given by algebraic laws and are carried out in the same language in which programs are written. Combining forms are chosen not only for their programming power but also for the power of their associated algebraic laws. General theorems of the algebra give the detailed behavior and termination conditions for large classes of programs.A new class of computing systems uses the functional programming style both in its programming language and in its state transition rules. Unlike von Neumann languages, these systems have semantics loosely coupled to states—only one state transition occurs per major computation.},
journal = {Commun. ACM},
month = aug,
pages = {613–641},
numpages = {29},
keywords = {von Neumann languages, von Neumann computers, programming languages, program transformation, program termination, program correctness, models of computing systems, metacomposition, functional programming, functional forms, combining forms, applicative state transition systems, applicative computing systems, algebra of programs}
}

@ARTICLE{58356,
  author={Mead, C.},
  journal={Proceedings of the IEEE}, 
  title={Neuromorphic electronic systems}, 
  year={1990},
  volume={78},
  number={10},
  pages={1629-1636},
  doi={10.1109/5.58356}
}

@Article{Strukov2008,
author={Strukov, Dmitri B.
and Snider, Gregory S.
and Stewart, Duncan R.
and Williams, R. Stanley},
title={The missing memristor found},
journal={Nature},
year={2008},
month={5},
day={01},
volume={453},
number={7191},
pages={80-83},
abstract={Basic electronics textbooks list three fundamental passive circuit elements: resistors, capacitors and inductors. But nearly forty years ago, Leon Chua predicted the existence of a fourth, the memristor --- in effect a nonlinear resistor with memory. A paper from the Hewlett-Packard research lab now reports that memristance arises naturally in nanoscale systems where solid-state electronic and ionic transport are coupled under an external bias voltage. This finding can help explain many examples of apparently anomalous hysteretic current--voltage behaviour observed in electronic devices during the past 50 years. Memristors may have a significant impact on future electronic circuits by dramatically increasing the functional density over that achieved by transistors.},
issn={1476-4687},
doi={10.1038/nature06932},
%url={https://doi.org/10.1038/nature06932}
}

@book{adamatzky2005reaction,
  author    = {Andrew Adamatzky and Ben De Lacy Costello and Tetsuya Asai},
  title     = {Reaction-Diffusion Computers},
  publisher = {Elsevier Science},
  year      = {2005},
  isbn      = {9780444520425},
  doi       = {10.1016/B978-0-444-52042-5.X5000-2},
  address   = {Oxford},
  url       = {https://www.sciencedirect.com/book/monograph/9780444520425/reaction-diffusion-computers}
}

@Article{Feynman1982,
author={Feynman, Richard P.},
title={Simulating physics with computers},
journal={International Journal of Theoretical Physics},
year={1982},
month={6},
day={01},
volume={21},
number={6},
pages={467-488},
issn={1572-9575},
doi={10.1007/BF02650179},
%url={https://doi.org/10.1007/BF02650179}
}

@book{sohr2001navierstokes,
  author    = {Hermann Sohr},
  title     = {The Navier-Stokes Equations: An Elementary Functional Analytic Approach},
  series    = {Birkhäuser Advanced Texts Basler Lehrbücher},
  publisher = {Birkhäuser Basel},
  year      = {2001},
  edition   = {1},
  isbn      = {978-3-0348-8255-2},
  doi       = {10.1007/978-3-0348-8255-2},
  address   = {Basel},
  %url       = {https://link.springer.com/book/10.1007/978-3-0348-8255-2}
}

@article{Adamatzky_2011,
   title={On computing in fine-grained compartmentalised {B}elousov–{Z}habotinsky medium},
   volume={44},
   ISSN={0960-0779},
   %url={http://dx.doi.org/10.1016/j.chaos.2011.03.010},
   DOI={10.1016/j.chaos.2011.03.010},
   number={10},
   journal={Chaos, Solitons, and Fractals},
   publisher={Elsevier BV},
   author={Adamatzky, Andrew and Holley, Julian and Bull, Larry and De Lacy Costello, Ben},
   year={2011},
   month=oct, pages={779–790} }

@Article{Wright2022,
author={Wright, Logan G.
and Onodera, Tatsuhiro
and Stein, Martin M.
and Wang, Tianyu
and Schachter, Darren T.
and Hu, Zoey
and McMahon, Peter L.},
title={Deep physical neural networks trained with backpropagation},
journal={Nature},
year={2022},
month={1},
day={01},
volume={601},
number={7894},
pages={549-555},
abstract={Deep-learning models have become pervasive tools in science and engineering. However, their energy requirements now increasingly limit their scalability1. Deep-learning accelerators2--9 aim to perform deep learning energy-efficiently, usually targeting the inference phase and often by exploiting physical substrates beyond conventional electronics. Approaches so far10--22 have been unable to apply the backpropagation algorithm to train unconventional novel hardware in situ. The advantages of backpropagation have made it the de facto training method for large-scale neural networks, so this deficiency constitutes a major impediment. Here we introduce a hybrid in situ--in silico algorithm, called physics-aware training, that applies backpropagation to train controllable physical systems. Just as deep learning realizes computations with deep neural networks made from layers of mathematical functions, our approach allows us to train deep physical neural networks made from layers of controllable physical systems, even when the physical layers lack any mathematical isomorphism to conventional artificial neural network layers. To demonstrate the universality of our approach, we train diverse physical neural networks based on optics, mechanics and electronics to experimentally perform audio and image classification tasks. Physics-aware training combines the scalability of backpropagation with the automatic mitigation of imperfections and noise achievable with in situ algorithms. Physical neural networks have the potential to perform machine learning faster and more energy-efficiently than conventional electronic processors and, more broadly, can endow physical systems with automatically designed physical functionalities, for example, for robotics23--26, materials27--29 and smart sensors30--32.},
issn={1476-4687},
doi={10.1038/s41586-021-04223-6},
%url={https://doi.org/10.1038/s41586-021-04223-6}
}

@ARTICLE{8259423,
  author={Davies, Mike and Srinivasa, Narayan and Lin, Tsung-Han and Chinya, Gautham and Cao, Yongqiang and Choday, Sri Harsha and Dimou, Georgios and Joshi, Prasad and Imam, Nabil and Jain, Shweta and Liao, Yuyun and Lin, Chit-Kwan and Lines, Andrew and Liu, Ruokun and Mathaikutty, Deepak and McCoy, Steven and Paul, Arnab and Tse, Jonathan and Venkataramanan, Guruguhanathan and Weng, Yi-Hsin and Wild, Andreas and Yang, Yoonseok and Wang, Hong},
  journal={IEEE Micro}, 
  title={Loihi: A Neuromorphic Manycore Processor with On-Chip Learning}, 
  year={2018},
  volume={38},
  number={1},
  pages={82-99},
  doi={10.1109/MM.2018.112130359}
}

@article{akopyan2015truenorth,
  author    = {Filipp Akopyan and Jun Sawada and Andrew S. Cassidy and Rodrigo Alvarez-Icaza and John V. Arthur and Paul A. Merolla and Nabil Imam and Yutaka Nakamura and Pallab Datta and Gi-Joon Nam and Brian Taba and Michael P. Beakes and Bernard Brezzo and Jente B. Kuang and Rajit Manohar and William P. Risk and Bryan L. Jackson and Dharmendra S. Modha},
  title     = {TrueNorth: Design and Tool Flow of a 65 mW 1 Million Neuron Programmable Neurosynaptic Chip},
  journal   = {IEEE Transactions on Computer-Aided Design of Integrated Circuits and Systems},
  volume    = {34},
  number    = {10},
  pages     = {1537--1557},
  year      = {2015},
  doi       = {10.1109/TCAD.2015.2474396}
}

@article{Hopfield1982,
author = {J J Hopfield },
title = {Neural networks and physical systems with emergent collective computational abilities.},
journal = {Proceedings of the National Academy of Sciences},
volume = {79},
number = {8},
pages = {2554-2558},
year = {1982},
doi = {10.1073/pnas.79.8.2554},
URL = {https://www.pnas.org/doi/abs/10.1073/pnas.79.8.2554},
eprint = {https://www.pnas.org/doi/pdf/10.1073/pnas.79.8.2554}
}

@techreport{Jaeger2001,
author = {Jaeger, Herbert},
year = {2001},
month = {1},
number = {GMD Report 148},
pages = {},
title = {The “Echo State” Approach to Analysing and Training Recurrent Neural Networks},
institution = {GMD Forschungszentrum Informationstechnik, St.
Augustin, Germany},
doi={10.24406/publica-fhg-291111}
}

@article{Tanaka2019,
title = {Recent advances in physical reservoir computing: A review},
journal = {Neural Networks},
volume = {115},
pages = {100-123},
year = {2019},
issn = {0893-6080},
doi = {https://doi.org/10.1016/j.neunet.2019.03.005},
url = {https://www.sciencedirect.com/science/article/pii/S0893608019300784},
author = {Gouhei Tanaka and Toshiyuki Yamane and Jean Benoit Héroux and Ryosho Nakane and Naoki Kanazawa and Seiji Takeda and Hidetoshi Numata and Daiju Nakano and Akira Hirose}
}

@InProceedings{Fernando2003,
author="Fernando, Chrisantha
and Sojakka, Sampsa",
editor="Banzhaf, Wolfgang
and Ziegler, Jens
and Christaller, Thomas
and Dittrich, Peter
and Kim, Jan T.",
title="Pattern Recognition in a Bucket",
booktitle="Advances in Artificial Life",
year="2003",
publisher="Springer Berlin Heidelberg",
address="Berlin, Heidelberg",
pages="588--597",
isbn="978-3-540-39432-7",
doi={10.1007/978-3-540-39432-7_63},
}

@ARTICLE{Scellier2017,
  title    = "Equilibrium Propagation: Bridging the Gap between Energy-Based
              Models and Backpropagation",
  author   = "Scellier, Benjamin and Bengio, Yoshua",
  journal  = "Front Comput Neurosci",
  volume   =  11,
  pages    = "24",
  month    =  5,
  year     =  2017,
  address  = "Switzerland",
  language = "en",
  doi= {10.3389/fncom.2017.00024}
}

@article{Stern2021,
  title = {Supervised Learning in Physical Networks: From Machine Learning to Learning Machines},
  author = {Stern, Menachem and Hexner, Daniel and Rocks, Jason W. and Liu, Andrea J.},
  journal = {Phys. Rev. X},
  volume = {11},
  issue = {2},
  pages = {021045},
  numpages = {18},
  year = {2021},
  month = {5},
  publisher = {American Physical Society},
  doi = {10.1103/PhysRevX.11.021045},
  url = {https://link.aps.org/doi/10.1103/PhysRevX.11.021045}
}

@article{Dillavou2022,
  title = {Demonstration of Decentralized Physics-Driven Learning},
  author = {Dillavou, Sam and Stern, Menachem and Liu, Andrea J. and Durian, Douglas J.},
  journal = {Phys. Rev. Appl.},
  volume = {18},
  issue = {1},
  pages = {014040},
  numpages = {12},
  year = {2022},
  month = {7},
  publisher = {American Physical Society},
  doi = {10.1103/PhysRevApplied.18.014040},
  url = {https://link.aps.org/doi/10.1103/PhysRevApplied.18.014040}
}

@article{Purcell1977,
    author = {Purcell, E. M.},
    title = {Life at low {Reynolds} number},
    journal = {American Journal of Physics},
    volume = {45},
    number = {1},
    pages = {3-11},
    year = {1977},
    month = {01},
    abstract = {Editor’s note: This is a reprint (slightly edited) of a paper of the same title that appeared in the book Physics and Our World: A Symposium in Honor of Victor F. Weisskopf, published by the American Institute of Physics (1976). The personal tone of the original talk has been preserved in the paper, which was itself a slightly edited transcript of a tape. The figures reproduce transparencies used in the talk. The demonstration involved a tall rectangular transparent vessel of corn syrup, projected by an overhead projector turned on its side. Some essential hand waving could not be reproduced.},
    issn = {0002-9505},
    doi = {10.1119/1.10903},
    %url = {https://doi.org/10.1119/1.10903},
    %eprint = {https://pubs.aip.org/aapt/ajp/article-pdf/45/1/3/11809839/3_1_online.pdf},
}

@book{Mead1989,
author = {Mead, Carver},
title = {Analog VLSI and neural systems},
year = {1989},
isbn = {0201059924},
publisher = {Addison-Wesley Longman Publishing Co., Inc.},
address = {USA},
doi={10.1007/978-1-4613-1639-8},
}

@article{Soloveichik2010,
author = {David Soloveichik  and Georg Seelig  and Erik Winfree },
title = {{DNA} as a universal substrate for chemical kinetics},
journal = {Proceedings of the National Academy of Sciences},
volume = {107},
number = {12},
pages = {5393-5398},
year = {2010},
doi = {10.1073/pnas.0909380107},
URL = {https://www.pnas.org/doi/abs/10.1073/pnas.0909380107},
eprint = {https://www.pnas.org/doi/pdf/10.1073/pnas.0909380107}
}

@article{Florijn2014,
  title = {Programmable Mechanical Metamaterials},
  author = {Florijn, Bastiaan and Coulais, Corentin and van Hecke, Martin},
  journal = {Phys. Rev. Lett.},
  volume = {113},
  issue = {17},
  pages = {175503},
  numpages = {5},
  year = {2014},
  month = {10},
  publisher = {American Physical Society},
  doi = {10.1103/PhysRevLett.113.175503},
  url = {https://link.aps.org/doi/10.1103/PhysRevLett.113.175503}
}

@article{Hjelmfelt1991,
author = {A Hjelmfelt  and E D Weinberger  and J Ross },
title = {Chemical implementation of neural networks and {T}uring machines.},
journal = {Proceedings of the National Academy of Sciences},
volume = {88},
number = {24},
pages = {10983-10987},
year = {1991},
doi = {10.1073/pnas.88.24.10983},
URL = {https://www.pnas.org/doi/abs/10.1073/pnas.88.24.10983},
eprint = {https://www.pnas.org/doi/pdf/10.1073/pnas.88.24.10983}}

@ARTICLE{Adleman1994,
  title    = "Molecular computation of solutions to combinatorial problems",
  author   = "Adleman, L M",
  journal  = "Science",
  volume   =  266,
  number   =  5187,
  pages    = "1021--1024",
  month    =  nov,
  year     =  1994,
  address  = "United States",
  language = "en",
  doi      = {10.1126/science.7973651},
}

@ARTICLE{Kirkpatrick1983,
  title    = "Optimization by simulated annealing",
  author   = "Kirkpatrick, S and Gelatt, Jr, C D and Vecchi, M P",
  journal  = "Science",
  volume   =  220,
  number   =  4598,
  pages    = "671--680",
  month    =  may,
  year     =  1983,
  address  = "United States",
  language = "en",
  doi = {10.1126/science.220.4598.671},
}

@ARTICLE{Hughes2019,
  title    = "Wave physics as an analog recurrent neural network",
  author   = "Hughes, Tyler W and Williamson, Ian A D and Minkov, Momchil and
              Fan, Shanhui",
  journal  = "Sci Adv",
  volume   =  5,
  number   =  12,
  pages    = "eaay6946",
  month    =  {12},
  year     =  2019,
  address  = "United States",
  language = "en",
  doi = {10.1126/sciadv.aay6946},
}

@Article{Nakajima2015,
author={Nakajima, Kohei
and Hauser, Helmut
and Li, Tao
and Pfeifer, Rolf},
title={Information processing via physical soft body},
journal={Scientific Reports},
year={2015},
month={5},
day={27},
volume={5},
number={1},
pages={10487},
issn={2045-2322},
doi={10.1038/srep10487},
url={https://doi.org/10.1038/srep10487}
}

@Article{Vandoorne2014,
author={Vandoorne, Kristof
and Mechet, Pauline
and Van Vaerenbergh, Thomas
and Fiers, Martin
and Morthier, Geert
and Verstraeten, David
and Schrauwen, Benjamin
and Dambre, Joni
and Bienstman, Peter},
title={Experimental demonstration of reservoir computing on a silicon photonics chip},
journal={Nature Communications},
year={2014},
month={3},
day={24},
volume={5},
number={1},
pages={3541},
issn={2041-1723},
doi={10.1038/ncomms4541},
url={https://doi.org/10.1038/ncomms4541}
}

@article{Hopfield1984,
author = {J J Hopfield },
title = {Neurons with graded response have collective computational properties like those of two-state neurons.},
journal = {Proceedings of the National Academy of Sciences},
volume = {81},
number = {10},
pages = {3088-3092},
year = {1984},
doi = {10.1073/pnas.81.10.3088},
URL = {https://www.pnas.org/doi/abs/10.1073/pnas.81.10.3088},
eprint = {https://www.pnas.org/doi/pdf/10.1073/pnas.81.10.3088}}

@ARTICLE{Steinbock1995,
  title    = "Navigating complex labyrinths: optimal paths from chemical waves",
  author   = "Steinbock, O and T{\'o}th, A and Showalter, K",
  journal  = "Science",
  volume   =  267,
  number   =  5199,
  pages    = "868--871",
  month    =  feb,
  year     =  1995,
  address  = "United States",
  language = "en",
  doi = {10.1126/science.267.5199.868},
}

@ARTICLE{Tero2010,
  title    = "Rules for biologically inspired adaptive network design",
  author   = "Tero, Atsushi and Takagi, Seiji and Saigusa, Tetsu and Ito,
              Kentaro and Bebber, Dan P and Fricker, Mark D and Yumiki, Kenji
              and Kobayashi, Ryo and Nakagaki, Toshiyuki",
  journal  = "Science",
  volume   =  327,
  number   =  5964,
  pages    = "439--442",
  month    =  jan,
  year     =  2010,
  address  = "United States",
  language = "en",
  doi = {10.1126/science.117789},
}

@ARTICLE{Kagan2022,
  title    = "In vitro neurons learn and exhibit sentience when embodied in a
              simulated game-world",
  author   = "Kagan, Brett J and Kitchen, Andy C and Tran, Nhi T and
              Habibollahi, Forough and Khajehnejad, Moein and Parker, Bradyn J
              and Bhat, Anjali and Rollo, Ben and Razi, Adeel and Friston, Karl
              J",
  journal  = "Neuron",
  volume   =  110,
  number   =  23,
  pages    = "3952--3969.e8",
  month    =  dec,
  year     =  2022,
  address  = "United States",
  language = "en",
  doi = {10.1016/j.neuron.2022.09.001},
}

@ARTICLE{Friston2010,
  title    = "The free-energy principle: a unified brain theory?",
  author   = "Friston, Karl",
  journal  = "Nat Rev Neurosci",
  volume   =  11,
  number   =  2,
  pages    = "127--138",
  month    =  jan,
  year     =  2010,
  address  = "England",
  language = "en",
  doi = {10.1038/nrn2787},
}

@ARTICLE{Smirnova2023,
  title    = "Organoid intelligence ({OI}) - The ultimate functionality of a
              brain microphysiological system",
  author   = "Smirnova, Lena and Morales Pantoja, Itzy E and Hartung, Thomas",
  journal  = "ALTEX",
  volume   =  40,
  number   =  2,
  pages    = "191--203",
  year     =  2023,
  address  = "Germany",
  language = "en",
  doi = {10.14573/altex.2303261},
}

@ARTICLE{Lee2022,
  title    = "Mechanical neural networks: Architected materials that learn
              behaviors",
  author   = "Lee, Ryan H and Mulder, Erwin A B and Hopkins, Jonathan B",
  journal  = "Sci Robot",
  volume   =  7,
  number   =  71,
  pages    = "eabq7278",
  month    =  oct,
  year     =  2022,
  address  = "United States",
  language = "en",
  doi = {10.1126/scirobotics.abq7278},
}

@ARTICLE{Bertoldi2017,
  title    = "Flexible mechanical metamaterials",
  author   = "Bertoldi, Katia and Vitelli, Vincenzo and Christensen, Johan and
              van Hecke, Martin",
  journal  = "Nature Reviews Materials",
  volume   =  2,
  number   =  11,
  pages    = "17066",
  month    =  oct,
  year     =  2017,
  doi = {10.1038/natrevmats.2017.66},
}

@ARTICLE{Lin2018,
  title    = "All-optical machine learning using diffractive deep neural
              networks",
  author   = "Lin, Xing and Rivenson, Yair and Yardimci, Nezih T and Veli,
              Muhammed and Luo, Yi and Jarrahi, Mona and Ozcan, Aydogan",
  journal  = "Science",
  volume   =  361,
  number   =  6406,
  pages    = "1004--1008",
  month    =  jul,
  year     =  2018,
  address  = "United States",
  language = "en",
  doi = {10.1126/science.aat8084	},
}

@ARTICLE{Whitesides2006,
  title    = "The origins and the future of microfluidics",
  author   = "Whitesides, George M",
  journal  = "Nature",
  volume   =  442,
  number   =  7101,
  pages    = "368--373",
  month    =  jul,
  year     =  2006,
  doi = {10.1038/nature05058},
}

@article{Unger2000,
author = {Marc A. Unger  and Hou-Pu Chou  and Todd Thorsen  and Axel Scherer  and Stephen R. Quake },
title = {Monolithic Microfabricated Valves and Pumps by Multilayer Soft Lithography},
journal = {Science},
volume = {288},
number = {5463},
pages = {113-116},
year = {2000},
doi = {10.1126/science.288.5463.113},
URL = {https://www.science.org/doi/abs/10.1126/science.288.5463.113},
eprint = {https://www.science.org/doi/pdf/10.1126/science.288.5463.113}
}

@article{Prakash2007,
author = {Manu Prakash  and Neil Gershenfeld },
title = {Microfluidic Bubble Logic},
journal = {Science},
volume = {315},
number = {5813},
pages = {832-835},
year = {2007},
doi = {10.1126/science.1136907},
URL = {https://www.science.org/doi/abs/10.1126/science.1136907},
eprint = {https://www.science.org/doi/pdf/10.1126/science.1136907}
}

@ARTICLE{Wehner2016,
  title    = "An integrated design and fabrication strategy for entirely soft,
              autonomous robots",
  author   = "Wehner, Michael and Truby, Ryan L and Fitzgerald, Daniel J and
              Mosadegh, Bobak and Whitesides, George M and Lewis, Jennifer A
              and Wood, Robert J",
  journal  = "Nature",
  volume   =  536,
  number   =  7617,
  pages    = "451--455",
  month    =  aug,
  year     =  2016,
  doi = {10.1038/nature19100},
}

@article{Preston2019,
author = {Daniel J. Preston  and Philipp Rothemund  and Haihui Joy Jiang  and Markus P. Nemitz  and Jeff Rawson  and Zhigang Suo  and George M. Whitesides },
title = {Digital logic for soft devices},
journal = {Proceedings of the National Academy of Sciences},
volume = {116},
number = {16},
pages = {7750-7759},
year = {2019},
doi = {10.1073/pnas.1820672116},
URL = {https://www.pnas.org/doi/abs/10.1073/pnas.1820672116},
eprint = {https://www.pnas.org/doi/pdf/10.1073/pnas.1820672116}
}

@ARTICLE{Daiguji2005,
  title    = "Nanofluidic diode and bipolar transistor",
  author   = "Daiguji, Hirofumi and Oka, Yukiko and Shirono, Katsuhiro",
  journal  = "Nano Lett",
  volume   =  5,
  number   =  11,
  pages    = "2274--2280",
  month    =  nov,
  year     =  2005,
  address  = "United States",
  language = "en",
  doi = {10.1021/nl051646y},
}

@article{Robin2021,
author = {Paul Robin  and Nikita Kavokine  and Lydéric Bocquet },
title = {Modeling of emergent memory and voltage spiking in ionic transport through angstrom-scale slits},
journal = {Science},
volume = {373},
number = {6555},
pages = {687-691},
year = {2021},
doi = {10.1126/science.abf7923},
URL = {https://www.science.org/doi/abs/10.1126/science.abf7923},
eprint = {https://www.science.org/doi/pdf/10.1126/science.abf7923}
}

@article{Robin2023,
author = {P. Robin  and T. Emmerich  and A. Ismail  and A. Niguès  and Y. You  and G.-H. Nam  and A. Keerthi  and A. Siria  and A. K. Geim  and B. Radha  and L. Bocquet },
title = {Long-term memory and synapse-like dynamics in two-dimensional nanofluidic channels},
journal = {Science},
volume = {379},
number = {6628},
pages = {161-167},
year = {2023},
doi = {10.1126/science.adc9931},
URL = {https://www.science.org/doi/abs/10.1126/science.adc9931},
eprint = {https://www.science.org/doi/pdf/10.1126/science.adc9931}
}

@ARTICLE{Nielsen2016,
  title    = "Genetic circuit design automation",
  author   = "Nielsen, Alec A K and Der, Bryan S and Shin, Jonghyeon and
              Vaidyanathan, Prashant and Paralanov, Vanya and Strychalski,
              Elizabeth A and Ross, David and Densmore, Douglas and Voigt,
              Christopher A",
  journal  = "Science",
  volume   =  352,
  number   =  6281,
  pages    = "aac7341",
  month    =  apr,
  year     =  2016,
  address  = "United States",
  language = "en",
  doi = {10.1126/science.aac7341},
}

@ARTICLE{Chen2025,
  title    = "Capacitive in-sensor tactile computing",
  author   = "Chen, Yan and Cao, Jie and Qiu, Jie and Yang, Dongzi and Liu,
              Mengyang and Zhang, Mengru and Li, Chenyang and Wu, Zhongyuan and
              Yu, Jie and Zhang, Xumeng and Chen, Xianzhe and Huang, Zhangcheng
              and Song, Enming and Wang, Ming and Liu, Qi and Liu, Ming",
  journal  = "Nat Commun",
  volume   =  16,
  number   =  1,
  pages    = "5691",
  month    =  jul,
  year     =  2025,
  address  = "England",
  language = "en",
  doi      = {10.1038/s41467-025-60703-7},

}

@Article{D5SC06176H,
author ="Hu, Yaxue and Zhang, Jinghui and Shen, Ke and Shen, Wei and Lee, Hian Kee and Tang, Sheng",
title  ="Intelligent molecular logic computing toolkits: nucleic acid-based construction{,} functionality{,} and enhanced biosensing applications",
journal  ="Chem. Sci.",
year  ="2025",
volume  ="16",
issue  ="43",
pages  ="20139-20180",
publisher  ="The Royal Society of Chemistry",
doi  ="10.1039/D5SC06176H",
%url  ="http://dx.doi.org/10.1039/D5SC06176H",
abstract  ="Molecular logic computing is revolutionizing biosensing by enabling intelligent{,} programmable detection{,} moving beyond simple target recognition to advanced molecular-level information processing. By employing biological molecules such as DNA/RNA{,} proteins/enzymes{,} or even whole biological cells as building blocks for creating molecular logic toolkits{,} logic operations have made rapid progress in molecular logic-based biosensing. In this review{,} we present a comprehensive overview of intelligent molecular logic operation toolkits and their contributions to advancing biosensing technologies. We first outline the design principles of these toolkits{,} detailing various types of logic gates{,} including Boolean{,} combinatorial{,} and sequential logic{,} as well as advanced feedback systems{,} fuzzy logic{,} and reversible logic. We delve into the construction of DNA-based{,} synthetic{,} and nanomaterial-based logical operation toolkits. Following this{,} we explore the functionalities of intelligent molecular logic computing toolkits{,} which encompass modular multi-signal integration{,} activatable lock–key (OFF–ON) reconfigurable control{,} programmable control{,} and logic-gated nanomachines. We also elaborate on the analytical mechanisms underpinning molecular logic-gated operations that utilize various detection platforms{,} including fluorescent{,} colorimetric{,} and electrochemical techniques{,} along with artificial intelligence-powered and smartphone-based detection platforms. Applications spanning genetic analysis{,} cancer analysis{,} pathogen identification{,} living cell logic analysis{,} and point-of-care diagnostics are highlighted. Finally{,} the future challenges associated with molecular logic toolkits in enhancing biosensing and potential solutions were outlined{,} providing insights into practical obstacles as well as future trends and prospects."}

@InProceedings{Wang2025,
author="Wang, Xiao
and Borras, Hendrik
and Klein, Bernhard
and Fr{\"o}ning, Holger",
editor="Ribeiro, Rita P.
and Pfahringer, Bernhard
and Japkowicz, Nathalie
and Larra{\~{n}}aga, Pedro
and Jorge, Al{\'i}pio M.
and Soares, Carlos
and Abreu, Pedro H.
and Gama, Jo{\~a}o",
title="Variance-Aware Noisy Training: Hardening {DNN}s Against Unstable Analog Computations",
booktitle="Proc. Mach. Learn. Knowl. Discovery Databases. Res. Track",
year="2025",
publisher="Springer Nature Switzerland",
address="Cham",
pages="147--163",
isbn="978-3-032-06109-6",
doi={10.1007/978-3-032-06109-6_9},
}

@misc{rolandi2026,
      title={Energy-Time-Accuracy Tradeoffs in Thermodynamic Computing}, 
      author={Alberto Rolandi and Paolo Abiuso and Patryk Lipka-Bartosik and Maxwell Aifer and Patrick J. Coles and Martí Perarnau-Llobet},
      year={2026},
      eprint={2601.04358},
      archivePrefix={arXiv},
      primaryClass={cond-mat.stat-mech},
      doi = {10.48550/arXiv.2601.04358},
      %url={https://arxiv.org/abs/2601.04358}, 
}

@ARTICLE{Sun2025,
  title    = "Reusable Noncomplementary {DNA}-Based Neural Network",
  author   = "Sun, Chengjie and Liu, Xiaoyang and Zhong, Jiafeng and Zhou, Qin
              and Cheng, Jianjun",
  journal  = "J Am Chem Soc",
  volume   =  147,
  number   =  38,
  pages    = "34339--34349",
  month    =  aug,
  year     =  2025,
  address  = "United States",
  language = "en",
  doi= {10.1021/jacs.5c04886},
}

@article{song2025heat,
  author    = {Tianqi Song and Lulu Qian},
  title     = {Heat-rechargeable computation in {DNA} logic circuits and neural networks},
  journal   = {Nature},
  volume    = {646},
  pages     = {315--322},
  year      = {2025},
  doi       = {10.1038/s41586-025-09570-2}
}

@ARTICLE{Chen2023,
  title    = "{DNA} strand displacement based computational systems and their
              applications",
  author   = "Chen, Congzhou and Wen, Jinda and Wen, Zhibin and Song, Sijie and
              Shi, Xiaolong",
  journal  = "Front Genet",
  volume   =  14,
  pages    = "1120791",
  month    =  feb,
  year     =  2023,
  address  = "Switzerland",
  language = "en",
  doi={10.3389/fgene.2023.1120791}
}

@article{202601.0088,
	doi = {10.20944/preprints202601.0088.v1},
	url = {https://doi.org/10.20944/preprints202601.0088.v1},
	year = 2026,
	month = jan,
	publisher = {Preprints},
	author = {Artur A. Zagitov and Egor S. Korenkov and Nail R. Bashirov and Roman A. Maksimov and Alexander M. Vinogradov and Aleksander N. Beznosikov and Maxim P. Nikitin},
	title = {Backpropagation in Molecular Neural Networks: Teaching {DNA} to Solve Machine Learning Tasks},
	journal = {Preprints}
}

@article{zhan2023recent,
  author    = {Pengfei Zhan and Andreas Peil and Qiao Jiang and Dongfang Wang and Shikufa Mousavi and Qiancheng Xiong and Qi Shen and Yingxu Shang and Baoquan Ding and Chenxiang Lin and Yonggang Ke and Na Liu},
  title     = {Recent Advances in {DNA} Origami-Engineered Nanomaterials and Applications},
  journal   = {Chemical Reviews},
  volume    = {123},
  number    = {7},
  pages     = {3976--4050},
  year      = {2023},
  doi       = {10.1021/acs.chemrev.3c00028}
}

@article{rothemund2006folding,
  author  = {Paul W. K. Rothemund},
  title   = {Folding {DNA} to create nanoscale shapes and patterns},
  journal = {Nature},
  volume  = {440},
  number  = {7082},
  pages   = {297--302},
  year    = {2006},
  doi     = {10.1038/nature04586}
}

@ARTICLE{Huang2025,
  author={Huang, Chun and Shao, Jiaying and Peng, Baolei and Guo, Qingshuang and Li, Panlong and Sun, Junwei and Wang, Yanfeng},
  journal={IEEE Transactions on Computers}, 
  title={Design of a Universal Decoder Model Based on {DNA} Winner-Takes-All Neural Networks}, 
  year={2025},
  volume={74},
  number={4},
  pages={1267-1277},
  doi={10.1109/TC.2024.3521230}
}

@ARTICLE{11045176,
  author={Sun, Junwei and Wang, Haojie and Yue, Yi and Ling, Dan and Wang, Yanfeng},
  journal={IEEE Transactions on Neural Networks and Learning Systems}, 
  title={Design of {H}opfield Neural Network Based on {DNA} Strand Displacement Circuits and Its Application in Sudoku Conjecture}, 
  year={2025},
  volume={36},
  number={10},
  pages={18889-18899},
  doi={10.1109/TNNLS.2025.3576888}
}

@ARTICLE{Liu2025,
  title    = "Deep convolutional and fully-connected {DNA} neural networks",
  author   = "Liu, Xiao and Zheng, Ziyang and Liu, Pei and Yang, Zheng and Hu,
              Hao and Wu, Meng and Lou, Xiaoding and Xia, Fan and Tao, Kaixiong
              and Li, Longjie and Dai, Jun and Xiao, Xianjin",
  journal  = "Nat Commun",
  volume   =  16,
  number   =  1,
  pages    = "10629",
  month    =  nov,
  year     =  2025,
  address  = "England",
  language = "en",
  doi={10.1038/s41467-025-65618-x},
}

@ARTICLE{Rizik2022,
  title    = "Synthetic neuromorphic computing in living cells",
  author   = "Rizik, Luna and Danial, Loai and Habib, Mouna and Weiss, Ron and
              Daniel, Ramez",
  journal  = "Nature Communications",
  volume   =  13,
  number   =  1,
  pages    = "5602",
  month    =  sep,
  year     =  2022,
  doi = {10.1038/s41467-022-33288-8},
}

@article {Leon2025,
	author = {Leon, Claudia Montufar and Wang, Yao and Bisso, Frank Britto and Mehta, Arya and Samaniego, Christian Cuba},
	title = {{CRISPR}-based neuromorphic computing for solving regression and classification},
	elocation-id = {2025.12.03.692209},
	year = {2025},
	doi = {10.64898/2025.12.03.692209},
	publisher = {Cold Spring Harbor Laboratory},
	URL = {https://www.biorxiv.org/content/early/2025/12/07/2025.12.03.692209},
	eprint = {https://www.biorxiv.org/content/early/2025/12/07/2025.12.03.692209.full.pdf},
	journal = {bioRxiv}
}

@article{cherry2025supervised,
  author    = {Kevin M. Cherry and Lulu Qian},
  title     = {Supervised learning in {DNA} neural networks},
  journal   = {Nature},
  year      = {2025},
  doi       = {10.1038/s41586-025-09479-w}
}

@inproceedings{agostini2025towards,
  author    = {Sandro Agostini and Victoria Zarth and others},
  title     = {Towards on-chip chemical computing: The oscillating {B}elousov-{Z}habotinsky reaction in silicon microchips},
  booktitle = {SCS Fall Meeting 2025 Abstract Catalogue},
  year      = {2025},
  organization = {SCS},
  note      = {Poster PC-101, IBM Research},
  url       = {https://research.ibm.com/publications/towards-on-chip-chemical-computing-the-oscillating-belousov-zhabotinsky-reaction-in-silicon-microchips}
}

@ARTICLE{Baltussen2024,
  title    = "Chemical reservoir computation in a self-organizing reaction
              network",
  author   = "Baltussen, Mathieu G and de Jong, Thijs J and Duez, Quentin and
              Robinson, William E and Huck, Wilhelm T S",
  journal  = "Nature",
  volume   =  631,
  number   =  8021,
  pages    = "549--555",
  month    =  jul,
  year     =  2024,
  doi = {10.1038/s41586-024-07567-x},
}

@article{raissi2019pinn,
  author  = {Maziar Raissi and Paris Perdikaris and George E. Karniadakis},
  title   = {Physics-informed neural networks: A deep learning framework for solving forward and inverse problems involving nonlinear partial differential equations},
  journal = {Journal of Computational Physics},
  volume  = {378},
  pages   = {686--707},
  year    = {2019},
  doi     = {10.1016/j.jcp.2018.10.045}
}

@inproceedings{peng2025spinode,
  author    = {Peng, Wenqing and Liu, Zhi-Song and Boy, Michael},
  title     = {{SPIN-ODE}: Stiff Physics-Informed Neural {ODE} for Chemical Reaction Rate Estimation},
  booktitle = {ECAI 2025 -- 28th European Conference on Artificial Intelligence, including 14th Conference on Prestigious Applications of Intelligent Systems (PAIS 2025) -- Proceedings},
  editor    = {Lynce, Ines and Murano, Nello and Vallati, Mauro and Villata, Serena and Chesani, Federico and Milano, Michela and Omicini, Andrea and Dastani, Mehdi},
  series    = {Frontiers in Artificial Intelligence and Applications},
  volume    = {413},
  pages     = {2033--2040},
  publisher = {IOS Press},
  year      = {2025},
  doi       = {10.3233/FAIA251040},
  url       = {https://doi.org/10.3233/FAIA251040}
}

@misc{gaimann2025,
      title={Robustly optimal dynamics for active matter reservoir computing}, 
      author={Mario U. Gaimann and Miriam Klopotek},
      year={2025},
      eprint={2505.05420},
      archivePrefix={arXiv},
      primaryClass={nlin.AO},
      url={https://arxiv.org/abs/2505.05420}, 
      doi = {10.48550/arXiv.2505.05420},
}

@article{li2025anti_interference_neuronal,
  author    = {Xiang Li and Mai Lu},
  title     = {Study on the anti-interference characteristics of neuronal networks: a comparative study of chemical synapses and electrical synapse},
  journal   = {Frontiers in Neuroscience},
  volume    = {19},
  pages     = {1581347},
  year      = {2025},
  doi       = {10.3389/fnins.2025.1581347},
  url       = {https://www.frontiersin.org/journals/neuroscience/articles/10.3389/fnins.2025.1581347/full}
}

@ARTICLE{Zhu2025,
  title    = "Autonomous Soft Robots: Self-regulation, Self-sustained, and
              Recovery Strategies",
  author   = "Zhu, Chen and Liu, Bo-Yu and Zhang, Li-Zhi and Xu, Lin",
  journal  = "Chinese Journal of Polymer Science",
  volume   =  43,
  number   =  4,
  pages    = "535--547",
  month    =  apr,
  year     =  2025,
  doi = {10.1007/s10118-025-3284-z},
}

@ARTICLE{Hopfield1985,
  title    = "``Neural'' computation of decisions in optimization problems",
  author   = "Hopfield, J J and Tank, D W",
  journal  = "Biological Cybernetics",
  volume   =  52,
  number   =  3,
  pages    = "141--152",
  month    =  jul,
  year     =  1985,
  doi      = {10.1007/BF00339943},
}

@article{QianWinfree2011,
  title = {Neural network computation with {DNA} strand displacement cascades},
  author = {Qian, Lulu and Winfree, Erik and Bruck, Jehoshua},
  journal = {Nature},
  volume = {475},
  number = {7356},
  pages = {368--372},
  year = {2011},
  doi = {10.1038/nature10262}
}

@article{QianWinfree2014,
  title = {Scaling up digital circuit computation with {DNA} strand displacement cascades},
  author = {Qian, Lulu and Winfree, Erik},
  journal = {Science},
  volume = {332},
  number = {6034},
  pages = {1196--1201},
  year = {2011},
  doi = {10.1126/science.1200520}
}

@article{Daniel2013,
  title = {Synthetic analog computation in living cells},
  author = {Daniel, R. and Rubens, J. and Sarpeshkar, R. and Lu, T. K.},
  journal = {Nature},
  volume = {497},
  pages = {619--623},
  year = {2013},
  doi = {10.1038/nature12148}
}

@article{TothShowalter1995,
  title = {Logic gates in excitable media},
  author = {Toth, Agnes and Showalter, Kenneth},
  journal = {Journal of Chemical Physics},
  volume = {103},
  pages = {2058--2066},
  year = {1995},
  doi = {10.1063/1.469732}
}

@article{Torrejon2017,
  title = {Neuromorphic computing with nanoscale spintronic oscillators},
  author = {Torrejon, Jacob and others},
  journal = {Nature},
  volume = {547},
  pages = {428--431},
  year = {2017},
  doi = {10.1038/nature23011}
}

@article{Cucchi2021,
author = {Matteo Cucchi  and Christopher Gruener  and Lautaro Petrauskas  and Peter Steiner  and Hsin Tseng  and Axel Fischer  and Bogdan Penkovsky  and Christian Matthus  and Peter Birkholz  and Hans Kleemann  and Karl Leo },
title = {Reservoir computing with biocompatible organic electrochemical networks for brain-inspired biosignal classification},
journal = {Science Advances},
volume = {7},
number = {34},
pages = {eabh0693},
year = {2021},
doi = {10.1126/sciadv.abh0693},
URL = {https://www.science.org/doi/abs/10.1126/sciadv.abh0693},
eprint = {https://www.science.org/doi/pdf/10.1126/sciadv.abh0693}
}

@article{Lagzi2010,
  title     = "Maze Solving by Chemotactic Droplets",
  author    = "Lagzi, Istv{\'a}n and Soh, Siowling and Wesson, Paul J and
               Browne, Kevin P and Grzybowski, Bartosz A",
  journal   = "J. Am. Chem. Soc.",
  publisher = "American Chemical Society",
  volume    =  132,
  number    =  4,
  pages     = "1198--1199",
  month     =  feb,
  year      =  2010,
  doi = {10.1021/ja9076793}
}

@article{Prezioso2015,
  title   = {Training and operation of an integrated neuromorphic network based on metal-oxide memristors},
  author  = {Prezioso, Mirko and Merrikh-Bayat, Farnood and Hoskins, Brian D. and Adam, Gina C. and Likharev, Konstantin K. and Strukov, Dmitri B.},
  journal = {Nature},
  volume  = {521},
  number  = {7550},
  pages   = {61--64},
  year    = {2015},
  doi     = {10.1038/nature14441}
}

@inproceedings{Burr2014,
  title     = {Experimental demonstration and tolerancing of a large-scale neural network (165,000 synapses), using phase-change memory as the synaptic weight element},
  author    = {Burr, Geoffrey W. and Shelby, Robert M. and Sidler, Scott and di Nolfo, Carmine and Jang, Jihyun and Boybat, Irem and Shenoy, Ravi S. and Narayanan, Pritish and Virwani, Kapil and Giacometti, Eugenio U. and Kurdi, Boudewijn N. and Hwang, H.-S.},
  booktitle = {2014 IEEE International Electron Devices Meeting (IEDM)},
  year      = {2014},
  pages     = {29.5.1--29.5.4},
  doi       = {10.1109/IEDM.2014.7047135}
}

@article{LeGallo2018,
  title   = {Mixed-precision in-memory computing},
  author  = {Le Gallo, Manuel and Sebastian, Abu and Mathis, Roland and Manica, Matteo and Giefers, Heiner and Tuma, Tomas and Bekas, Costas and Curioni, Alessandro and Eleftheriou, Evangelos},
  journal = {Nature Electronics},
  volume  = {1},
  number  = {4},
  pages   = {246--253},
  year    = {2018},
  doi     = {10.1038/s41928-018-0054-8}
}

@article{Boybat2018,
  title   = {Neuromorphic computing with multi-memristive synapses},
  author  = {Boybat, Irem and Le Gallo, Manuel and Nandakumar, S. R. and Moraitis, Timoleon and Parnell, Thomas and Tuma, Tomas and Rajendran, Bipin and Leblebici, Yusuf and Sebastian, Abu and Eleftheriou, Evangelos},
  journal = {Nature Communications},
  volume  = {9},
  number  = {1},
  pages   = {2514},
  year    = {2018},
  doi     = {10.1038/s41467-018-04933-y}
}

@article{Joshi2020,
  title   = {Accurate deep neural network inference using computational phase-change memory},
  author  = {Joshi, Vinay and Le Gallo, Manuel and Haefeli, Simon and Boybat, Irem and Nandakumar, S. R. and Piveteau, Christophe and Dazzi, Martino and Rajendran, Bipin and Sebastian, Abu and Eleftheriou, Evangelos},
  journal = {Nature Communications},
  volume  = {11},
  pages   = {2473},
  year    = {2020},
  doi     = {10.1038/s41467-020-16108-9}
}

@article{Sebastian2020,
  title   = {Memory devices and applications for in-memory computing},
  author  = {Sebastian, Abu and Le Gallo, Manuel and Khaddam-Aljameh, Riduan and Eleftheriou, Evangelos},
  journal = {Nature Nanotechnology},
  volume  = {15},
  number  = {7},
  pages   = {529--544},
  year    = {2020},
  doi     = {10.1038/s41565-020-0655-z}
}

@article{Miller2013,
  title   = {Self-configuring universal linear optical component},
  author  = {Miller, David A. B.},
  journal = {Photonics Research},
  volume  = {1},
  number  = {1},
  pages   = {1--15},
  year    = {2013},
  doi     = {10.1364/PRJ.1.000001}
}

@article{Shen2017,
  title   = {Deep learning with coherent nanophotonic circuits},
  author  = {Shen, Yichen and Harris, Nicholas C. and Skirlo, Scott and Prabhu, Mihika and Baehr-Jones, Tom and Hochberg, Michael and Sun, Xin and Zhao, Shijie and Larochelle, Hugo and Englund, Dirk and Solja{\v c}i{\'c}, Marin},
  journal = {Nature Photonics},
  volume  = {11},
  pages   = {441--446},
  year    = {2017},
  doi     = {10.1038/nphoton.2017.93}
}

@article{Tait2017,
  title   = {Neuromorphic photonic networks using silicon photonic weight banks},
  author  = {Tait, Alexander N. and de Lima, Tiago F. and Zhou, Enzhu and Wu, Axel X. and Nahmias, Mark A. and Shastri, Bhavin J. and Prucnal, Paul R.},
  journal = {Scientific Reports},
  volume  = {7},
  pages   = {7430},
  year    = {2017},
  doi     = {10.1038/s41598-017-07754-z}
}

@article{Prucnal2016,
  title   = {Recent progress in semiconductor excitable lasers for photonic spike processing},
  author  = {Prucnal, Paul R. and Shastri, Bhavin J.},
  journal = {Advances in Optics and Photonics},
  volume  = {8},
  number  = {2},
  pages   = {228--299},
  year    = {2016},
  doi     = {10.1364/AOP.8.000228}
}

@article{Appeltant2011,
  title   = {Information processing using a single dynamical node as complex system},
  author  = {Appeltant, Lennert and Soriano, Miguel C. and Van der Sande, Guy and Danckaert, Jan and Massar, Serge and Dambre, Joni and Schrauwen, Benjamin and Mirasso, Claudio R. and Fischer, Ingo},
  journal = {Nature Communications},
  volume  = {2},
  pages   = {468},
  year    = {2011},
  doi     = {10.1038/ncomms1476}
}

@article{Paquot2012,
  title   = {Optoelectronic reservoir computing},
  author  = {Paquot, Yanne and Duport, Fran{\c c}ois and Smerieri, Andrea and Haelterman, Marc and Massar, Serge},
  journal = {Scientific Reports},
  volume  = {2},
  pages   = {287},
  year    = {2012},
  doi     = {10.1038/srep00287}
}

@article{Hughes2018,
  title   = {Training of photonic neural networks through in situ backpropagation and gradient measurement},
  author  = {Hughes, Tyler W. and Minkov, Momchil and Shi, Yu and Fan, Shanhui},
  journal = {Optica},
  volume  = {5},
  number  = {7},
  pages   = {864--871},
  year    = {2018},
  doi     = {10.1364/OPTICA.5.000864}
}

@book{Adamatzky2010,
  title={Physarum Machines},
  author={Adamatzky, Andrew},
  publisher={World Scientific},
  year={2010},
  doi = {10.1142/7968},
}

@article{Tamsir2011,
  title={Robust multicellular computing using genetically encoded {NOR} gates and chemical wires},
  author={Tamsir, Anselm and Tabor, Jeffrey J. and Voigt, Christopher A.},
  journal={Nature},
  volume={469},
  pages={212--215},
  year={2011},
  doi={10.1038/nature09565}
}

@article{Brenner2008,
  title={Engineering microbial consortia: a new frontier in synthetic biology},
  author={Brenner, Kate and You, Lingchong and Arnold, Frances},
  journal={Trends in Biotechnology},
  volume={26},
  pages={483--489},
  year={2008},
  doi = {10.1016/j.tibtech.2008.05.004},
}

@article{Elowitz2000,
  title={A synthetic oscillatory network of transcriptional regulators},
  author={Elowitz, Michael and Leibler, Stanislas},
  journal={Nature},
  volume={403},
  pages={335--338},
  year={2000},
  doi = {10.1038/35002125},
}

@article{Danino2010,
  title={A synchronized quorum of genetic clocks},
  author={Danino, Tal and Mondrag{\'o}n-Palomino, Oswaldo and Tsimring, Lev and Hasty, Jeff},
  journal={Nature},
  volume={463},
  pages={326--330},
  year={2010},
  doi = {10.1038/nature08753},
}

@article{Bi1998,
  title={Synaptic modifications in cultured hippocampal neurons: dependence on spike timing, synaptic strength, and postsynaptic cell type},
  author={Bi, Guo-Qiang and Poo, Mu-Ming},
  journal={Journal of Neuroscience},
  volume={18},
  pages={10464--10472},
  year={1998},
  doi = {https://www.jneurosci.org/content/18/24/10464},
}

@article{Obien2015,
  title={Revealing neuronal function through microelectrode array recordings},
  author={Obien, Markus and Deligkaris, Kosmas and Bullmann, Till and Bakkum, Douglas and Frey, Urs},
  journal={Frontiers in Neuroscience},
  volume={9},
  pages={423},
  year={2015},
  doi = {10.3389/fnins.2014.00423},
}

@article{Drotman2021,
  title   = {Electronics-free pneumatic circuits for controlling soft-legged robots},
  author  = {Drotman, Daniel and Jadhav, Snehal and Karimi, Mohammad and de Zonia, Paul and Tolley, Michael T.},
  journal = {Science Robotics},
  volume  = {6},
  number  = {56},
  year    = {2021},
  pages   = {eaay2627},
  doi     = {10.1126/scirobotics.aay2627}
}

@book{Pfeifer2007,
  title={How the Body Shapes the Way We Think},
  author={Pfeifer, Rolf and Bongard, Josh},
  year={2007},
  publisher={MIT Press},
  doi = {10.7551/mitpress/3585.001.0001},
}

@article{Kim2013,
  title={Soft robotics: A bioinspired evolution in robotics},
  author={Kim, Sangbae and Laschi, Cecilia and Trimmer, Barry},
  journal={Trends in Biotechnology},
  volume={31},
  number={5},
  pages={287--294},
  year={2013},
  doi = {10.1016/j.tibtech.2013.03.002},
}

@article{Coulais2017,
  title={Combinatorial design of textured mechanical metamaterials},
  author={Coulais, Corentin and Teomy, Elie and de Reus, Kees and Shokef, Yair and van Hecke, Martin},
  journal={Nature},
  volume={535},
  pages={529--532},
  year={2016},
  doi = {10.1038/nature18960},
}

@article{Kane2014,
  title={Topological boundary modes in isostatic lattices},
  author={Kane, C. L. and Lubensky, T. C.},
  journal={Nature Physics},
  volume={10},
  pages={39--45},
  year={2014},
  doi = {10.1038/nphys2835},
}

@article{Waitukaitis2015,
  title={Origami multistability: From single vertices to metasheets},
  author={Waitukaitis, Scott and Menaut, Renaud and Chen, Bryan G. and van Hecke, Martin},
  journal={Physical Review Letters},
  volume={114},
  pages={055503},
  year={2015},
  doi={10.1103/PhysRevLett.114.055503},
}

@article{Zhai2021OrigamiKirigamiReview,
  title   = {Mechanical metamaterials based on origami and kirigami},
  author  = {Zhai, Zirui and Wu, Lingling and Jiang, Hanqing},
  journal = {Applied Physics Reviews},
  volume  = {8},
  number  = {4},
  pages   = {041319},
  year    = {2021},
  doi     = {10.1063/5.0051088}
}

@article{Nakajima2014,
  title    = "Exploiting short-term memory in soft body dynamics as a
              computational resource",
  author   = "Nakajima, K and Li, T and Hauser, H and Pfeifer, R",
  journal  = "J R Soc Interface",
  volume   =  11,
  number   =  100,
  pages    = "20140437",
  month    =  nov,
  year     =  2014,
  address  = "England",
  language = "en",
  doi = {10.1098/rsif.2014.0437},
}

@article{Sigmund2013,
  title={Topology optimization approaches},
  author={Sigmund, Ole and Maute, Kurt},
  journal={Structural and Multidisciplinary Optimization},
  volume={48},
  pages={1031--1055},
  year={2013},
  doi = {10.1007/s00158-013-0978-6},
}

@article{Squires2005,
  title={Microfluidics: Fluid physics at the nanoliter scale},
  author={Squires, Todd M. and Quake, Stephen R.},
  journal={Reviews of Modern Physics},
  volume={77},
  pages={977--1026},
  year={2005},
  doi = {10.1103/RevModPhys.77.977},
}

@article{Oh2012PressureDriven,
  title={Design of pressure-driven microfluidic networks using electric circuit analogy},
  author={Oh, Kwang W. and Lee, Kangsun and Ahn, Byungwook and Furlani, Edward P.},
  journal={Lab on a Chip},
  volume={12},
  number={3},
  pages={515--545},
  year={2012},
  publisher={Royal Society of Chemistry},
  doi = {10.1039/C2LC20799K},
}

@book{Bruus2008,
  title={Theoretical Microfluidics},
  author={Bruus, Henrik},
  series={Oxford Master Series in Condensed Matter Physics},
  volume={18},
  publisher={Oxford University Press},
  year={2008},
  address={Oxford},
  doi = {10.1093/oso/9780199235087.001.0001},
}

@article{Weaver2010,
  title={Static control logic for microfluidic devices using pressure-gain valves},
  author={Weaver, James A. and Melin, Jonah and Stark, Daniel and Quake, Stephen R. and Horowitz, Mark A.},
  journal={Nature Physics},
  volume={6},
  number={3},
  pages={218--223},
  year={2010},
  doi = {10.1038/nphys1513},
}

@misc{Holl2020,
      title={Learning to Control {PDE}s with Differentiable Physics}, 
      author={Philipp Holl and Vladlen Koltun and Nils Thuerey},
      year={2020},
      eprint={2001.07457},
      archivePrefix={arXiv},
      primaryClass={cs.LG},
      url={https://arxiv.org/abs/2001.07457},
      doi ={10.48550/arXiv.2001.07457},
}

@Article{D2LC00254J,
author ="McIntyre, David and Lashkaripour, Ali and Fordyce, Polly and Densmore, Douglas",
title  ="Machine learning for microfluidic design and control",
journal  ="Lab Chip",
year  ="2022",
volume  ="22",
issue  ="16",
pages  ="2925-2937",
publisher  ="The Royal Society of Chemistry",
doi  ="10.1039/D2LC00254J",
url  ="http://dx.doi.org/10.1039/D2LC00254J"}

@article{Sackmann2014,
  title={The present and future role of microfluidics in biomedical research},
  author={Sackmann, Eric K. and Fulton, Anna L. and Beebe, David J.},
  journal={Nature},
  volume={507},
  pages={181--189},
  year={2014},
  doi = {10.1038/nature13118},
}

@article{Dertinger2001,
  title={Generation of gradients having complex shapes using microfluidic networks},
  author={Dertinger, Stephan K. W. and Chiu, Daniel T. and Jeon, Noo Li and Whitesides, George M.},
  journal={Analytical Chemistry},
  volume={73},
  number={6},
  pages={1240--1246},
  year={2001},
  doi = {10.1021/ac001132d},
}

@article{Shastri2021,
  title={Photonics for artificial intelligence and neuromorphic computing},
  author={Shastri, Bhavin J. and Tait, Alexander N. and Ferreira de Lima, Thomas and Pernice, Wolfram H. P. and Bhaskaran, Harish and Wright, C. David and Prucnal, Paul R.},
  journal={Nature Photonics},
  volume={15},
  number={2},
  pages={102--114},
  year={2021},
  publisher={Nature Publishing Group},
  doi={10.1038/s41566-020-00754-y}
}

@article{Wetzstein2020,
  title={Inference in artificial intelligence with deep optics and photonics},
  author={Wetzstein, Gordon and Ozcan, Aydogan and Gigan, Sylvain and Fan, Shanhui and Englund, Dirk and Solja{\v{c}}i{\'c}, Marin and Denz, Cornelia and Miller, David A. B. and Psaltis, Demetri},
  journal={Nature},
  volume={588},
  number={7836},
  pages={39--47},
  year={2020},
  publisher={Nature Publishing Group},
  doi={10.1038/s41586-020-2973-6}
}

@article{Clements2016,
  title={Optimal design for universal multiport interferometers},
  author={Clements, William R. and Humphreys, Peter C. and Metcalf, Benjamin J. and Kolthammer, W. Steven and Walmsley, Ian A.},
  journal={Optica},
  volume={3},
  number={12},
  pages={1460--1465},
  year={2016},
  publisher={Optica Publishing Group},
  doi={10.1364/OPTICA.3.001460}
}

@article{Reck1994,
  title={Experimental realization of any discrete unitary operator},
  author={Reck, Michael and Zeilinger, Anton and Bernstein, Herbert J. and Bertani, Philip},
  journal={Physical Review Letters},
  volume={73},
  number={1},
  pages={58},
  year={1994},
  publisher={APS},
  doi={10.1103/PhysRevLett.73.58}
}

@article{Williamson2019,
  title={Reprogrammable electro-optic nonlinear activation functions for optical neural networks},
  author={Williamson, Ian A. D. and Hughes, Tyler W. and Minkov, Momchil and Bartlett, Ben and Pai, Sunil and Fan, Shanhui},
  journal={IEEE Journal of Selected Topics in Quantum Electronics},
  volume={26},
  number={1},
  pages={1--12},
  year={2019},
  publisher={IEEE},
  doi={10.1109/JSTQE.2019.2930455}
}

@article{Stern2020,
  title={Supervised learning through physical changes in a mechanical system},
  author={Stern, Menachem and Arinze, Chukwunonso and Perez, Leron and Palmer, Stephanie E. and Murugan, Arvind},
  journal={Proceedings of the National Academy of Sciences},
  volume={117},
  number={26},
  pages={14843--14850},
  year={2020},
  publisher={National Academy of Sciences},
  doi={10.1073/pnas.2000807117}
}

@inproceedings{Ion2017,
  title={Digital mechanical metamaterials},
  author={Ion, Alexandra and Wall, Johannes and Kovacs, Robert and Baudisch, Patrick},
  booktitle={Proceedings of the 2017 CHI Conference on Human Factors in Computing Systems},
  pages={977--988},
  year={2017},
  publisher={ACM},
  doi={10.1145/3025453.3025624}
}

@article{Bocquet2020,
  title={Nanofluidics coming of age},
  author={Bocquet, Lyd{\'e}ric},
  journal={Nature Materials},
  volume={19},
  number={3},
  pages={254--256},
  year={2020},
  publisher={Nature Publishing Group},
  doi = {10.1038/s41563-020-0625-8},
}

@article{Kavokine2022,
  title={Fluctuation-induced quantum friction in nanoscale water flows},
  author={Kavokine, Nikita and Bocquet, Marie-Laure and Bocquet, Lyd{\'e}ric},
  journal={Nature},
  volume={602},
  number={7895},
  pages={84--90},
  year={2022},
  publisher={Nature Publishing Group},
  doi={10.1038/s41586-021-04284-7}
}

@article{Tan2024_Vision,
  author    = {Tan, Hongwei and van Dijken, Sebastiaan},
  title     = {A universal neuromorphic vision processing system},
  journal   = {Nature Electronics},
  year      = {2024},
  volume    = {7},
  pages     = {946--956},
  publisher = {Nature Publishing Group},
  doi = {10.1038/s41928-024-01288-9},
}

@ARTICLE{Li2024-aw,
  title    = "High-performance ferroelectric field-effect transistors with
              ultra-thin indium tin oxide channels for flexible and transparent
              electronics",
  author   = "Li, Qingxuan and Wang, Siwei and Li, Zhenhai and Hu, Xuemeng and
              Liu, Yongkai and Yu, Jiajie and Yang, Yafen and Wang, Tianyu and
              Meng, Jialin and Sun, Qingqing and Zhang, David Wei and Chen, Lin",
  journal  = "Nature Communications",
  volume   =  15,
  number   =  1,
  pages    = "2686",
  month    =  mar,
  year     =  2024,
  doi = {10.1038/s41467-024-46878-5},
}

@ARTICLE{Khan2020-sh,
  title    = "The future of ferroelectric field-effect transistor technology",
  author   = "Khan, Asif Islam and Keshavarzi, Ali and Datta, Suman",
  journal  = "Nature Electronics",
  volume   =  3,
  number   =  10,
  pages    = "588--597",
  month    =  oct,
  year     =  2020,
  doi = {10.1038/s41928-020-00492-7},
}

@article{EverschorSitte2024,
  author    = {Everschor-Sitte, Karin and Majumdar, Atreya and Wolk, K. and Meier, Dennis},
  title     = {Topological magnetic and ferroelectric systems for reservoir computing},
  journal   = {Nature Reviews Physics},
  year      = {2024},
  month     = jun,
  volume    = {6},
  number    = {7},
  pages     = {455--462},
  publisher = {Nature Publishing Group},
  doi = {10.1038/s42254-024-00729-w},
}

@article{Markovic2020,
  author    = {Markovi{\'c}, Danijela and Grollier, Julie},
  title     = {Quantum neuromorphic computing},
  journal   = {Applied Physics Letters},
  year      = {2020},
  volume    = {117},
  number    = {15},
  pages     = {150501},
  publisher = {AIP Publishing},
doi ={10.1063/5.0020014},
}

@article{Pinna2020,
  author = {Pinna, D. and Bourianoff, G. and Everschor-Sitte, K.},
  title  = {Reservoir computing with random skyrmion textures},
  journal= {Phys. Rev. Applied},
  year   = {2020},
  volume = {14},
  pages  = {054020},
  doi ={10.1103/PhysRevApplied.14.054020},
}

@article{Yokouchi2022,
author = {Tomoyuki Yokouchi  and Satoshi Sugimoto  and Bivas Rana  and Shinichiro Seki  and Naoki Ogawa  and Yuki Shiomi  and Shinya Kasai  and Yoshichika Otani },
title = {Pattern recognition with neuromorphic computing using magnetic field–induced dynamics of skyrmions},
journal = {Science Advances},
volume = {8},
number = {39},
pages = {eabq5652},
year = {2022},
doi = {10.1126/sciadv.abq5652},
URL = {https://www.science.org/doi/abs/10.1126/sciadv.abq5652},
eprint = {https://www.science.org/doi/pdf/10.1126/sciadv.abq5652}
}

@article{Marrows2024,
  author  = {Marrows, Christopher H. and Barker, Joseph and Moore, Thomas A. and Moorsom, Timothy},
  title   = {Neuromorphic computing with spintronics},
  journal = {npj Spintronics},
  year    = {2024},
  volume  = {2},
  pages   = {12},
  doi     = {10.1038/s44306-024-00019-2}
}

@article{Schneider2020_SFQSciRep,
  author  = {Schneider, Michael L. and Donnelly, Christine A. and Haygood, Ian W. and Wynn, A. and Russek, Stephen E. and Pufall, M. R. and Dresselhaus, P. D. and Rippard, W. H.},
  title   = {Synaptic weighting in single flux quantum neuromorphic computing},
  journal = {Scientific Reports},
  year    = {2020},
  volume  = {10},
  pages   = {934},
  doi ={10.1038/s41598-020-57892-0},
}

@INPROCEEDINGS{Russek2016_SFQ,
  author={Russek, Stephen E. and Donnelly, Christine A. and Schneider, Michael L. and Baek, Burm and Pufall, Mathew R. and Rippard, William H. and Hopkins, Peter F. and Dresselhaus, Paul D. and Benz, Samuel P.},
  booktitle={2016 IEEE International Conference on Rebooting Computing (ICRC)}, 
  title={Stochastic single flux quantum neuromorphic computing using magnetically tunable Josephson junctions}, 
  year={2016},
  volume={},
  number={},
  pages={1-5},
  doi={10.1109/ICRC.2016.7738712}}

@article{Schneider2025_SelfTraining,
  author  = {Schneider, Michael L. and Jué, E. M. and Pufall, M. R. and Segall, K. and Anderson, C. W.},
  title   = {A self-training spiking superconducting neuromorphic architecture},
  journal = {npj Unconventional Computing},
  year    = {2025},
  volume  = {2},
  number  = {1},
  pages   = {Article No. 5},
  doi     = {10.1038/s44335-025-00021-9}
}

@article{Grollier2020,
  author    = {Grollier, Julie and Querlioz, Damien and Camsari, Kerem Y. and Everschor-Sitte, Karin and Fukami, Shunsuke and Stiles, Mark D.},
  title     = {Neuromorphic spintronics},
  journal   = {Nature Electronics},
  year      = {2020},
  volume    = {3},
  number    = {7},
  pages     = {360--370},
  publisher = {Nature Publishing Group},
  doi ={10.1038/s41928-019-0360-9},
}

@article{Niu2025_PNAS,
  author  = {Niu, Chang and Zhang, Huanyu and Xu, Chuanlong and Hu, Wenjie and Wu, Yunzhuo and Wu, Yu and Wang, Yadi and Wu, Yi and Zhu, Yinyan and Wang, Wenbin and Wu, Yizheng and Yin, Lifeng and Xiao, Jiang and Yu, Weichao and Guo, Hangwen and Shen, Jian},
  title   = {A self-learning magnetic {H}opfield neural network with intrinsic gradient descent adaption},
  journal = {Proceedings of the National Academy of Sciences},
  year    = {2024},
  month   = dec,
  volume  = {121},
  number  = {51},
  pages   = {e2416294121},
  doi     = {10.1073/pnas.2416294121}
}

@ARTICLE{Raab2022_Brownian,
  title    = "Brownian reservoir computing realized using geometrically
              confined skyrmion dynamics",
  author   = "Raab, Klaus and Brems, Maarten A and Beneke, Grischa and Dohi,
              Takaaki and Roth{\"o}rl, Jan and Kammerbauer, Fabian and Mentink,
              Johan H and Kl{\"a}ui, Mathias",
  journal  = "Nature Communications",
  volume   =  13,
  number   =  1,
  pages    = "6982",
  month    =  nov,
  year     =  2022,
  doi = {10.1038/s41467-022-34309-2},
}

@article{Holmes2013_PowerBudgets,
  author  = {Holmes, D. S. and Ripple, A. L. and Manheimer, M. A.},
  title   = {Energy-Efficient Superconducting Computing---Power Budgets and Requirements},
  journal = {IEEE Transactions on Applied Superconductivity},
  year    = {2013},
  volume  = {23},
  number  = {3},
  pages   = {1701610},
  doi ={10.1109/TASC.2013.2244634},
}

@ARTICLE{Jordan2024-jz,
  title    = "Open and remotely accessible Neuroplatform for research in
              wetware computing",
  author   = "Jordan, Fred D and Kutter, Martin and Comby, Jean-Marc and
              Brozzi, Flora and Kurtys, Ewelina",
  journal  = "Front Artif Intell",
  volume   =  7,
  pages    = "1376042",
  month    =  may,
  year     =  2024,
  address  = "Switzerland",
  language = "en",
  doi = {10.3389/frai.2024.1376042},
}

@article{Shalf2020,
    author = {Shalf, John},
    title = {The future of computing beyond Moore’s Law},
    journal = {Philosophical Transactions of the Royal Society A: Mathematical, Physical and Engineering Sciences},
    volume = {378},
    number = {2166},
    pages = {20190061},
    year = {2020},
    month = {01},
    issn = {1364-503X},
    doi = {10.1098/rsta.2019.0061}
}

@ARTICLE{Simmel2023-eq,
  title    = "Nucleic acid strand displacement - from {DNA} nanotechnology to
              translational regulation",
  author   = "Simmel, Friedrich C",
  journal  = "RNA Biol",
  volume   =  20,
  number   =  1,
  pages    = "154--163",
  month    =  jan,
  year     =  2023,
  address  = "United States",
  language = "en",
  doi={10.1080/15476286.2023.2204565}
}

@article{Sanjabi2019,
author = {Sanjabi, Mercedeh and Jahanian, Ali},
title = {Multi-threshold and multi-input {DNA} logic design style for profiling the microRNA biomarkers of real cancers},
journal = {IET Nanobiotechnology},
volume = {13},
number = {7},
pages = {665-673},
doi = {https://doi.org/10.1049/iet-nbt.2018.5275},
url = {https://ietresearch.onlinelibrary.wiley.com/doi/abs/10.1049/iet-nbt.2018.5275},
eprint = {https://ietresearch.onlinelibrary.wiley.com/doi/pdf/10.1049/iet-nbt.2018.5275},
year = {2019}
}

@ARTICLE{Okumura2022-jx,
  title    = "Nonlinear decision-making with enzymatic neural networks",
  author   = "Okumura, S and Gines, G and Lobato-Dauzier, N and Baccouche, A
              and Deteix, R and Fujii, T and Rondelez, Y and Genot, A J",
  journal  = "Nature",
  volume   =  610,
  number   =  7932,
  pages    = "496--501",
  month    =  oct,
  year     =  2022,
  doi      = {10.1038/s41586-022-05218-7},
}

@Article{Takiguchi2025,
author ="Takiguchi, Sotaro and Takeuchi, Nanami and Shenshin, Vasily and Gines, Guillaume and Genot, Anthony J. and Nivala, Jeff and Rondelez, Yannick and Kawano, Ryuji",
title  ="Harnessing {DNA} computing and nanopore decoding for practical applications: from informatics to {microRNA}-targeting diagnostics",
journal  ="Chem. Soc. Rev.",
year  ="2025",
volume  ="54",
issue  ="1",
pages  ="8-32",
publisher  ="The Royal Society of Chemistry",
doi  ="10.1039/D3CS00396E",
url  ="http://dx.doi.org/10.1039/D3CS00396E"}

@ARTICLE{Villar2013-gb,
  title    = "A tissue-like printed material",
  author   = "Villar, Gabriel and Graham, Alexander D and Bayley, Hagan",
  journal  = "Science",
  volume   =  340,
  number   =  6128,
  pages    = "48--52",
  month    =  apr,
  year     =  2013,
  address  = "United States",
  language = "en",
  doi = {10.1126/science.1229495},
}

@ARTICLE{Parrilla-Gutierrez2020-oo,
  title     = "A programmable chemical computer with memory and pattern
               recognition",
  author    = "Parrilla-Gutierrez, Juan Manuel and Sharma, Abhishek and Tsuda,
               Soichiro and Cooper, Geoffrey J T and Aragon-Camarasa, Gerardo
               and Donkers, Kevin and Cronin, Leroy",
  journal   = "Nat. Commun.",
  publisher = "Springer Science and Business Media LLC",
  volume    =  11,
  number    =  1,
  pages     = "1442",
  month     =  mar,
  year      =  2020,
  copyright = "https://creativecommons.org/licenses/by/4.0",
  language  = "en",
  doi = {10.1038/s41467-020-15190-3}
}

@ARTICLE{Cai2023-lr,
  title    = "Brain organoid reservoir computing for artificial intelligence",
  author   = "Cai, Hongwei and Ao, Zheng and Tian, Chunhui and Wu, Zhuhao and
              Liu, Hongcheng and Tchieu, Jason and Gu, Mingxia and Mackie, Ken
              and Guo, Feng",
  journal  = "Nature Electronics",
  volume   =  6,
  number   =  12,
  pages    = "1032--1039",
  month    =  dec,
  year     =  2023,
  doi = {10.1038/s41928-023-01069-w}
}

@article{Dykstra2019,
  title   = {Viscoelastic Snapping Metamaterials},
  author  = {Dykstra, David M. J. and Busink, Joris and Ennis, Bernard and Coulais, Corentin},
  journal = {Journal of Applied Mechanics},
  volume  = {86},
  number  = {11},
  pages   = {111012},
  year    = {2019},
  doi     = {10.1115/1.4044036}
}

@article{Keim2020,
  title   = {Global memory from local hysteresis in an amorphous solid},
  author  = {Keim, Nathan C. and Hass, Jacob and Kroger, Brian and Wieker, Devin},
  journal = {Physical Review Research},
  volume  = {2},
  number  = {1},
  pages   = {012004},
  year    = {2020},
  doi     = {10.1103/PhysRevResearch.2.012004}
}

@article{Pashine2019,
  title   = {Directed aging, memory, and nature's greed},
  author  = {Pashine, Nidhi and Hexner, Daniel and Liu, Andrea J. and Nagel, Sidney R.},
  journal = {Science Advances},
  volume  = {5},
  number  = {12},
  pages   = {eaax4215},
  year    = {2019},
  doi     = {10.1126/sciadv.aax4215}
}

@article{Du2022DiffPD,
  author  = {Du, Tao and Wu, Kui and Ma, Pingchuan and Wah, Sebastien and Spielberg, Andrew and Rus, Daniela and Matusik, Wojciech},
  title   = {DiffPD: Differentiable Projective Dynamics},
  journal = {ACM Transactions on Graphics},
  volume  = {41},
  number  = {2},
  articleno = {13},
  pages   = {1--21},
  year    = {2022},
  doi     = {10.1145/3490168}
}

@article{Dold2023GraphLattices,
  author  = {Dold, Dominik and Aranguren van Egmond, Derek},
  title   = {Differentiable graph-structured models for inverse design of lattice materials},
  journal = {Cell Reports Physical Science},
  volume  = {4},
  pages   = {101586},
  year    = {2023},
  doi     = {10.1016/j.xcrp.2023.101586}
}

@article{Miskin2013ArtificialEvolution,
  author  = {Miskin, Marc Z. and Jaeger, Heinrich M.},
  title   = {Adapting granular materials through artificial evolution},
  journal = {Nature Materials},
  volume  = {12},
  number  = {4},
  pages   = {326--331},
  year    = {2013},
  doi     = {10.1038/nmat3543}
}

@article{Dong2022VoxelEA,
  author  = {Dong, L. and Wang, D.},
  title   = {Optimal Design of Three-Dimensional Voxel Printed Multimaterial Lattice Metamaterials via Machine Learning and Evolutionary Algorithm},
  journal = {Physical Review Applied},
  volume  = {18},
  pages   = {054050},
  year    = {2022},
  doi     = {10.1103/PhysRevApplied.18.054050}
}

@article{Li2024AllMechanical,
  author  = {Li, Shuaifeng and Mao, Xiangcheng},
  title   = {Training all-mechanical neural networks for task learning through in situ backpropagation},
  journal = {Nature Communications},
  volume  = {15},
  pages   = {10528},
  year    = {2024},
  doi     = {10.1038/s41467-024-54849-z}
}

@article{Kamsma2024,
  author  = {Kamsma, Tim M. and Kim, Jaehyun and Kim, Kyungjun and Boon, Willem Q. and Spitoni, Cristian and Park, Jungyul and van Roij, Ren{\'e}},
  title   = {Brain-inspired computing with fluidic iontronic nanochannels},
  journal = {Proceedings of the National Academy of Sciences of the United States of America},
  year    = {2024},
  volume  = {121},
  number  = {18},
  pages   = {e2320242121},
  doi     = {10.1073/pnas.2320242121}
}

@article{Ismail2025,
  author  = {Ismail, Abdulghani and Nam, Gwang-Hyeon and Lokhandwala, Aziz and Pandey, Siddhi Vinayak and Saurav, Kalluvadi Veetil and You, Yi and Jyothilal, Hiran and Goutham, Solleti and Sajja, Ravalika and Keerthi, Ashok and Radha, Boya and others},
  title   = {Programmable memristors with two-dimensional nanofluidic channels},
  journal = {Nature Communications},
  year    = {2025},
  volume  = {16},
  pages   = {7008},
  doi     = {10.1038/s41467-025-61649-6}
}

@article{Luo2025Iontronics,
  title   = {Iontronic Devices from Biological Nanopores to Artificial Systems: Emerging Applications and Future Perspectives},
  author  = {Luo, Jiabei and Remy, Antoine and Zhang, Yujia},
  journal = {Chemical Reviews},
  year    = {2025},
  doi     = {10.1021/acs.chemrev.5c00579},
  url     = {https://pubs.acs.org/doi/10.1021/acs.chemrev.5c00579}
}

@article{Xu2024,
author = {Yi-Tong Xu  and Si-Yuan Yu  and Zheng Li  and Bo-Han Kou  and Jian-Xiang Pang  and Wei-Wei Zhao  and Hong-Yuan Chen  and Jing-Juan Xu },
title = {A nanofluidic spiking synapse},
journal = {Proceedings of the National Academy of Sciences},
volume = {121},
number = {28},
pages = {e2403143121},
year = {2024},
doi = {10.1073/pnas.2403143121},
URL = {https://www.pnas.org/doi/abs/10.1073/pnas.2403143121},
eprint = {https://www.pnas.org/doi/pdf/10.1073/pnas.2403143121}}

@ARTICLE{Momeni2025,
  title     = "Training of physical neural networks",
  author    = "Momeni, Ali and Rahmani, Babak and Scellier, Benjamin and
               Wright, Logan G and McMahon, Peter L and Wanjura, Clara C and
               Li, Yuhang and Skalli, Anas and Berloff, Natalia G and Onodera,
               Tatsuhiro and Oguz, Ilker and Morichetti, Francesco and Del
               Hougne, Philipp and Le Gallo, Manuel and Sebastian, Abu and
               Mirhoseini, Azalia and Zhang, Cheng and Markovi{\'c}, Danijela
               and Brunner, Daniel and Moser, Christophe and Gigan, Sylvain and
               Marquardt, Florian and Ozcan, Aydogan and Grollier, Julie and
               Liu, Andrea J and Psaltis, Demetri and Al{\`u}, Andrea and
               Fleury, Romain",
  journal   = "Nature",
  publisher = "Springer Science and Business Media LLC",
  volume    =  645,
  number    =  8079,
  pages     = "53--61",
  month     =  sep,
  year      =  2025,
  doi = {10.1038/s41586-025-09384-2} 
}

@article{Du2017MemristorRC,
  title        = {Reservoir computing using dynamic memristors for temporal information processing},
  author       = {Du, Chao and Cai, Fuxi and Zidan, Mohammed A. and Ma, Wen and Lee, Seung Hwan and Lu, Wei D.},
  journal      = {Nature Communications},
  year         = {2017},
  volume       = {8},
  number       = {1},
  pages        = {1--10},
  doi          = {10.1038/s41467-017-02337-y}
}

@article{Pistolesi2025,
    author = {Pistolesi, Veronica and Ceni, Andrea and Milano, Gianluca and Ricciardi, Carlo and Gallicchio, Claudio},
    title = {A memristive computational neural network model for time-series processing},
    journal = {APL Machine Learning},
    volume = {3},
    number = {1},
    pages = {016117},
    year = {2025},
    month = {03},
    issn = {2770-9019},
    doi = {10.1063/5.0255168}
}

@misc{horuz2026mars,
      title={Scalable Memristive-Friendly Reservoir Computing for Time Series Classification}, 
      author={Coşku Can Horuz and Andrea Ceni and Claudio Gallicchio and Sebastian Otte},
      year={2026},
      eprint={2604.19343},
      archivePrefix={arXiv},
      primaryClass={cs.NE},
      url={https://arxiv.org/abs/2604.19343}, 
}

@ARTICLE{OHagan2023,
  title     = "Photocleavable {Ortho-Nitrobenzyl-Protected} {DNA} Architectures
               and Their Applications",
  author    = "O'Hagan, Michael P and Duan, Zhijuan and Huang, Fujian and Laps,
               Shay and Dong, Jiantong and Xia, Fan and Willner, Itamar",
  journal   = "Chem. Rev.",
  publisher = "American Chemical Society",
  volume    =  123,
  number    =  10,
  pages     = "6839--6887",
  month     =  may,
  year      =  2023,
  doi = {10.1021/acs.chemrev.3c00016},
}

@article{Chen2018NeuralODE_arXiv,
  title   = {Neural Ordinary Differential Equations},
  author  = {Chen, Ricky T. Q. and Rubanova, Yulia and Bettencourt, Jesse and Duvenaud, David},
  journal = {arXiv preprint arXiv:1806.07366},
  year    = {2018},
  doi     = {10.48550/arXiv.1806.07366},
  url     = {https://arxiv.org/abs/1806.07366}
}
}
\begin{IEEEbiography}[{\includegraphics[width=1in,height=1.25in,clip,keepaspectratio]{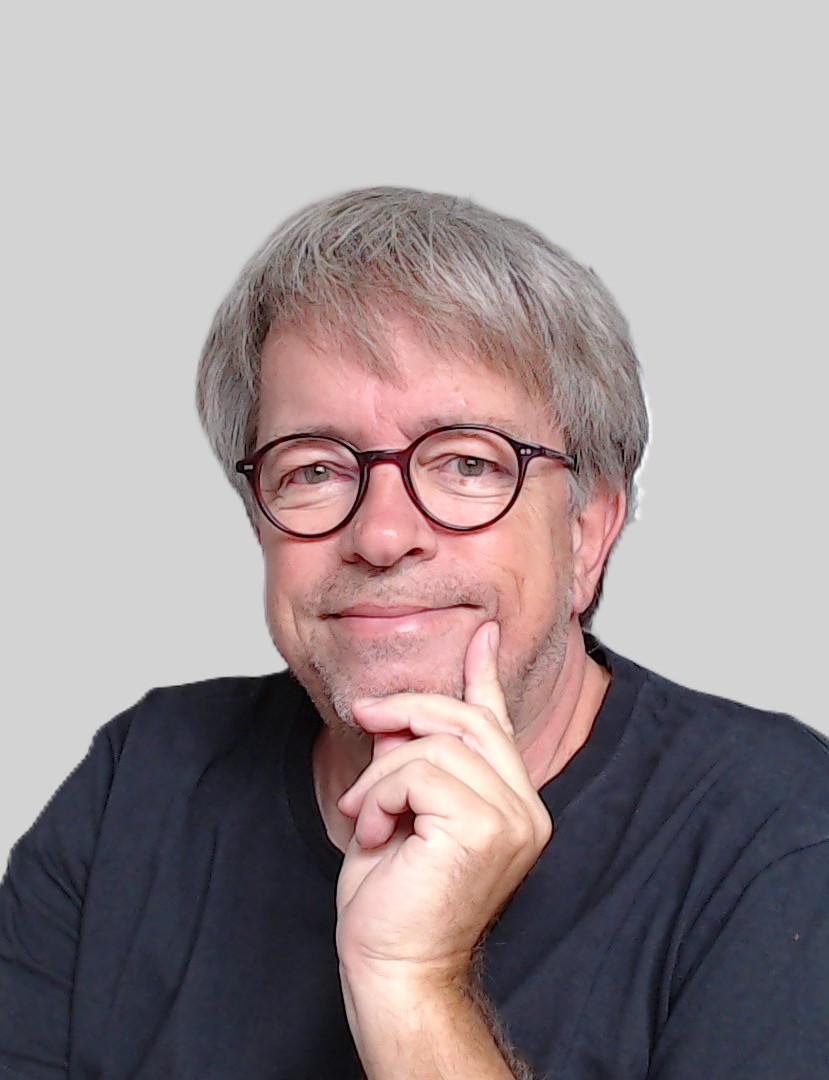}}]{Stefan Fischer} is a full professor in Computer Science at the University of Lübeck, Germany, and the director of the Institute for Telematics. He got his diploma degree in Information Systems and his doctoral degree in Computer Science from the University of Mannheim, Germany, in 1992 and 1996, respectively. After a postdoctoral year at the University of Montreal, Canada, he held positions at the International University in Germany as an assistant professor and at the Technical University of Braunschweig as an associate professor, until he joined Lübeck University in 2004. His research interest is currently focused on network and distributed system structures such as ad-hoc and sensor networks, Internet of Things, Smart Cities, and especially molecular communications on the nano level, based on DNA structures. He has (co-)authored more than 220 scientific books and articles. 
\end{IEEEbiography}
\begin{IEEEbiography}[{\includegraphics[width=1in,height=1.25in,clip,keepaspectratio]{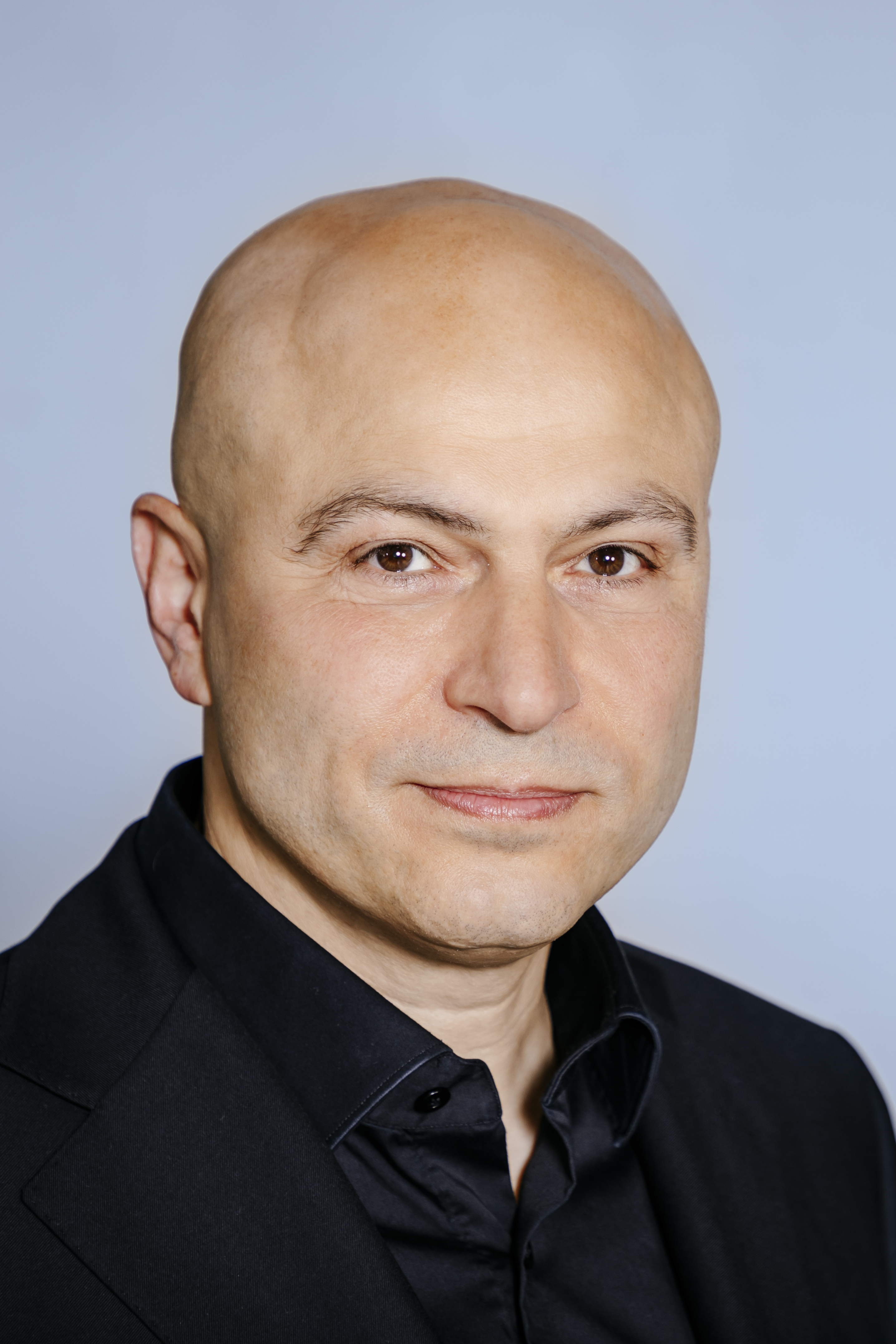}}]{Nihat Ay} is a full professor at the Hamburg University of Technology, Germany, where he heads the Institute for Data Science Foundations. He earned his PhD in Mathematics at Leipzig University in 2001. In 2003 and 2004, he was a postdoctoral fellow at the Santa Fe Institute and at the Redwood Neuroscience Institute (now the Redwood Center for Theoretical Neuroscience at UC Berkeley).  
From 2005 to 2021, he worked as a Max Planck Research Group Leader at the Max Planck Institute for Mathematics in the Sciences in Leipzig, where he was heading the group Information Theory of Cognitive Systems.
He is an external faculty member of the Santa Fe Institute, New Mexico, USA, and the Complexity Science Hub in Vienna, Austria. 
Nihat Ay works on information geometry and its applications to 
embodied intelligence and machine learning. 
He serves as the Editor-in-Chief of the Springer journal Information Geometry.
\end{IEEEbiography}
\begin{IEEEbiography}[{\includegraphics[width=1in,height=1.25in,clip,keepaspectratio]{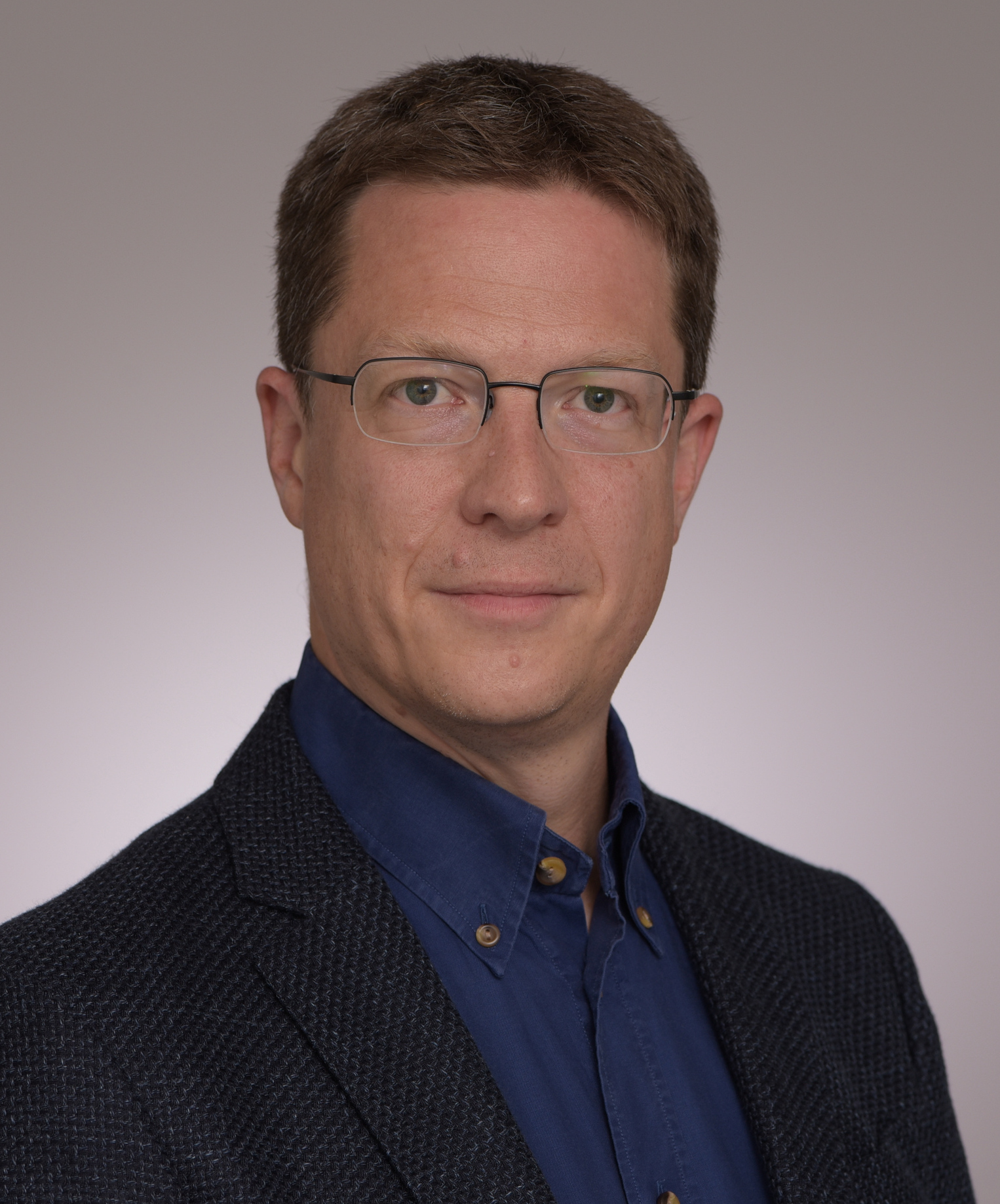}}]{Olaf Landsiedel} is a full professor of Computer Science at the Hamburg University of Technology (TUHH), Germany, where he heads the Institute for Networked Cyber-Physical Systems (NCPS). He earned his PhD at RWTH Aachen University in 2010, followed by a postdoctoral stay at KTH Royal Institute of Technology in Stockholm, Sweden. He was a Fulbright Scholar in the US from 2002 to 2004. Before joining TUHH, he held faculty positions at Chalmers University of Technology in Gothenburg, Sweden, and was a full professor leading the Distributed Systems Group at Kiel University, Germany. His research focuses on distributed and networked systems, with particular emphasis on the Internet of Things (IoT), cyber-physical systems, TinyML, Edge AI, and edge \& fog computing. 
\end{IEEEbiography}
\vspace{-0.4cm}
\begin{IEEEbiography}[{\includegraphics[width=1in,height=1.25in,clip,keepaspectratio]{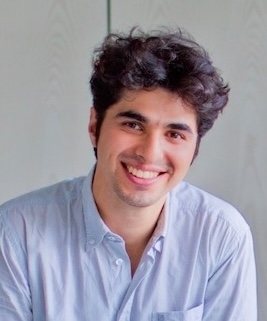}}]{Esfandiar Mohammadi} is a professor at the University of Lübeck (UzL), Germany, where he heads the Privacy \& Security group. He received his PhD at Saarland University in 2015, worked at ETH Zurich as a ZISC fellow from 2016 to 2019, and joined UzL in 2019. His research focuses on privacy-preserving machine learning, distributed learning, data privacy, backdoor injection in machine learning, and anonymous communication networks. Since 2018, he has served in the technical PCs of various international conferences, such as ACM CCS, USENIX Security, and PoPETS. Since 2022, he has served as the director of the AnoMed competence center, which focuses on data privacy and secure computation for medical applications and is one of the nation-wide three competence centers on data privacy. 
\end{IEEEbiography}
\begin{IEEEbiography}[{\includegraphics[width=1in,height=1.25in,clip,keepaspectratio]{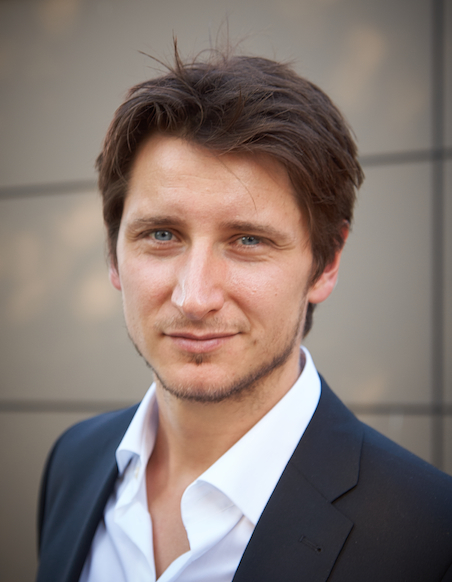}}]{Sebastian Otte} is a professor at the University of Lübeck, Germany, where he heads the Research Group for Adaptive Artificial Intelligence. His research focuses on flexible, continuously learning AI systems that operate efficiently under constraints such as limited energy, memory, and computational resources. The work combines artificial intelligence, cognitive science, and robotics in an interdisciplinary framework. He received his PhD in AI and computer science from the University of Tübingen in 2017 and subsequently conducted postdoctoral research there in the Neuro-Cognitive Modeling Group (2017–2023). In 2020, he served as acting professor for the Chair of Distributed Intelligence. As a Feodor Lynen Fellow, he was a visiting scientist at CWI in Amsterdam (2022–2023). He assumed his current professorship in Lübeck in September 2023.
\end{IEEEbiography}

%\newpage

\begin{IEEEbiography}[{\includegraphics[width=1in,height=1.25in,clip,keepaspectratio]{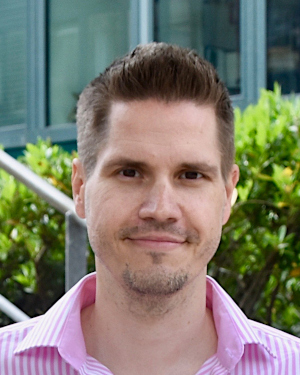}}]{Bernd-Christian Renner} is a full professor and head of the Institute for Autonomous Cyber-Physical Systems~(ACPS) at the Hamburg University of Technology ~(TUHH). From 2020 to 2022, he was an associate professor at the University of Koblenz-Landau, and from 2016 to 2020, he was an assistant professor at TUHH.
He has published articles in several international journals and conferences, and reviewed for and served on several TPCs. He has served the ENSsys workshop at ACM SenSys in several roles since 2014 and is a member of its steering committee. His research interests include applications of networked low-power sensing and cyber-physical systems running on energy harvested from the environment. He is also active in networking protocols for networks of embedded sensors, active and passive backscatter acoustic communication, localization, and navigation in mobile, low-power underwater networks.
\end{IEEEbiography}

%\newpage

\begin{IEEEbiography}[{\includegraphics[width=1in,height=1.25in,clip,keepaspectratio]{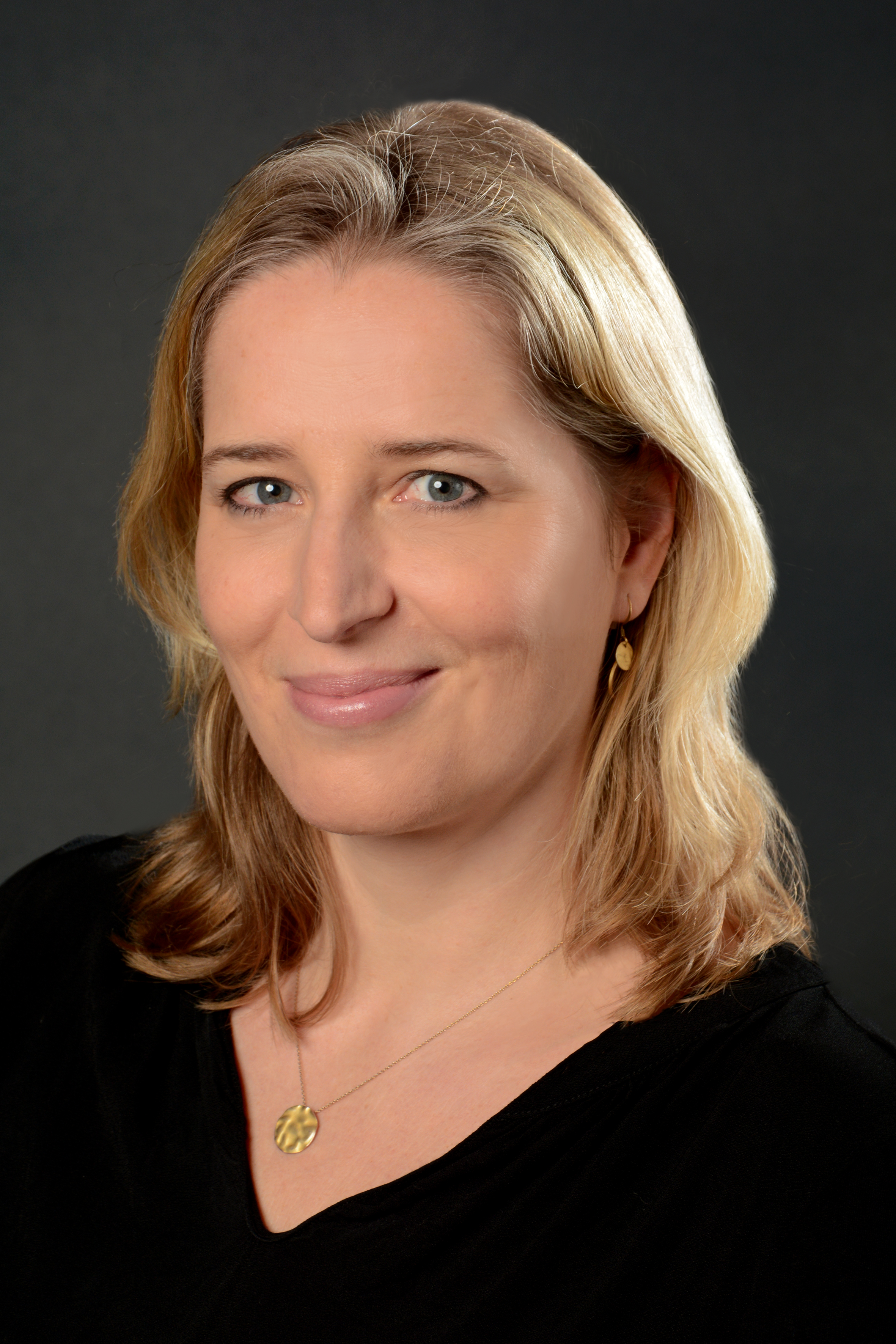}}]{Nele Russwinkel} Nele Russwinkel is a full professor at the University of Lübeck, Germany, for Human Aware AI and the head of the Institute for Information Systems. Her research focuses on cognitive Systems and intelligent agents interacting in dynamic environments. She is also working on cognitive Robotics and the questions how to provide a sense of Control for such systems in dynamic and partly unknown environments. She received her PhD at the DFG-founded Research Training Group Prometei (Prospective human machine interaction) at the TU Berlin. She was holding a Juniorprofessorship at the TU Berlin and got the call to Lübeck in 2022.
She is the president of the International Society for Cognitive Modelling, the vice president of the German Cognitive Science Community (GKev.).
\end{IEEEbiography}

\vfill

\end{document}